\documentclass[lettersize,journal]{IEEEtran}

\usepackage[numbers]{natbib}
\usepackage{amsmath,amsfonts,amssymb,amsthm,bm}
\usepackage{algorithmic}
\usepackage{algorithm}
\usepackage{array}
\usepackage[caption=false,font=normalsize,labelfont=sf,textfont=sf]{subfig}
\usepackage{textcomp}
\usepackage{stfloats}
\usepackage{url}
\usepackage{verbatim}
\usepackage{graphicx}
\usepackage{booktabs}
\usepackage{multirow}
\usepackage[table,dvipsnames]{xcolor} 
\usepackage{colortbl}   
\usepackage{tikz}
\usepackage{forest}
\usepackage{makecell} 
\usepackage{setspace}

\definecolor{myLinkColor}{RGB}{0,0,128} 
\definecolor{myCiteColor}{RGB}{0,100,100} 
\usepackage[colorlinks=true,linkcolor=myLinkColor,citecolor=myCiteColor,urlcolor=myLinkColor]{hyperref}

\usepackage[margin=1.5cm]{geometry}
\usepackage{cleveref}
\crefname{section}{Sec.}{Sec.}
\crefname{subsection}{Sec.}{Sec.}
\crefname{subsubsection}{Sec.}{Sec.}
\crefname{figure}{Fig.}{Fig.}
\crefname{table}{Tab.}{Tab.}
\crefname{equation}{Eq.}{Eq.}

\definecolor{lan}{rgb}{0.353,0.714,0.722}
\definecolor{mul}{rgb}{0.635,0.698,0.922}
\definecolor{uni}{rgb}{0.718,0.427,0.435}

\usepackage{pifont}  

\definecolor{MatteLight}{HTML}{D6EAF8}
\definecolor{MatteDark} {HTML}{2471A3}
\definecolor{BaseLight} {HTML}{D5F5E3}
\definecolor{BaseDark}  {HTML}{1E8449}
\definecolor{TrainLight}{HTML}{E8DAEF}
\definecolor{TrainDark} {HTML}{7D3C98}
\definecolor{InferLight}{HTML}{FDEBD0}
\definecolor{InferDark} {HTML}{D68910}
\definecolor{AppLight}  {HTML}{EBDEF0}
\definecolor{AppDark}   {HTML}{7D3C98}
\definecolor{FutLight}  {HTML}{FADBD8}
\definecolor{FutDark}   {HTML}{A93226}
\definecolor{CiteBoxBG} {RGB}{245,245,245}

\usepackage{fontawesome5}

\usepackage{titlesec}
\titlelabel{\thetitle\ }

\renewcommand\thesubsection{\thesection.\Alph{subsection}}

\renewcommand\thesubsubsection{\thesection.\Alph{subsection}.\arabic{subsubsection}}

\titleclass{\subsubsubsection}{straight}[\subsubsection]

\newcounter{subsubsubsection}[subsubsection]
\renewcommand\thesubsubsubsection{\thesection.\Alph{subsection}.\arabic{subsubsection}.(\alph{subsubsubsection})}

\titleformat{\subsubsubsection}[block]
  {\normalfont\normalsize\itshape}         
  {\thesubsubsubsection}           
  {4pt}                               
  {}                               
  
\titlespacing*{\subsubsubsection}
  {0pt}    
  {1ex plus 0.0ex minus 0.2ex} 
  {0ex plus 0ex}              

\begin{document}
\title{Discrete Diffusion in Large Language and Multimodal Models: A Survey}

\author{
Runpeng Yu\textsuperscript{*}, \, Qi Li\textsuperscript{*}, \, Xinchao Wang\textsuperscript{\dag}\\
National Univerisity of Singapore\\
\{r.yu, liqi\}@u.nus.edu, xinchao@nus.edu.sg
}



\maketitle

\begin{abstract}
In this work, we provide a systematic survey of Discrete Diffusion Language Models (dLLMs) and Discrete Diffusion Multimodal Language Models (dMLLMs). Unlike autoregressive (AR) models, dLLMs and dMLLMs adopt a multi-token, parallel decoding paradigm using full attention and a denoising-based generation strategy. This paradigm naturally enables parallel generation, fine-grained output control, and dynamic perception. These capabilities are previously difficult to achieve with AR models. A growing number of industrial-scale proprietary d(M)LLMs, as well as a large number of open-source academic d(M)LLMs, have demonstrated performance comparable to their autoregressive counterparts, while achieving up to \textit{10$\times$} acceleration in inference speed. These developments position discrete diffusion models as a promising alternative to intelligence based on the traditional autoregressive approach.
In this work, we present a comprehensive overview of the research in the dLLM and dMLLM domains. We trace the historical development of dLLMs and dMLLMs, formalize the underlying mathematical frameworks, list commonly-used modeling methods, and categorize representative models. We further analyze key techniques for training, inference, quantization. We also discuss the trustworthy issues and summarize emerging applications across language, vision-language, and biological domains and \textit{etc.}. We conclude by discussing future directions for research and deployment.
\end{abstract}

\href{https://github.com/LiQiiiii/Awesome-Discrete-Diffusion-LLM_MLLM}{\faGithub\ Relative papers are collected in this repo.} 
\vspace{0.5em}

\begin{IEEEkeywords}
Discrete Diffusion, Large Language Model, Multimodal Large Language Model, Diffusion Large Language Model, Diffusion Multimodal Large Language Model, Language Model, Unified Model
\end{IEEEkeywords}

\begingroup
\renewcommand\thefootnote{} 
\footnotetext{------------------------------------------------------------}
\footnotetext{* Equal contribution, random order.}
\footnotetext{\dag\ Corresponding author.}
\endgroup

\section{Introduction}
\label{intro}

In recent years, Large Language Models (LLMs) and Multimodal Large Language Models (MLLMs) have demonstrated remarkable advances, exhibiting capabilities that increasingly resemble, or even surpass, human-level performance in domains traditionally associated with intelligence. Modern LLMs and MLLMs achieve superior scores on standard benchmarks designed for general knowledge, comprehension, and reasoning, suggesting that these systems are no longer merely text completion engines but competent general-purpose agents.

To date, the predominant paradigm for both LLMs and MLLMs has been autoregressive (AR)~\cite{openai2024gpt4ocard,openai2024gpt4technicalreport,deepseekai2025deepseekr1incentivizingreasoningcapability,geminiteam2024gemini15unlockingmultimodal,geminiteam2025geminifamilyhighlycapable}. 
Despite their successes, autoregressive (AR) models face intrinsic limitations. Their left-to-right decoding hinders parallel inference, reducing efficiency. They also struggle with precise structural control (e.g., length or format), making natural language inefficient for fine-grained orchestration of tools and agentic tasks. Moreover, causal attention forces one-pass static perception of inputs, limiting dynamic task-aware and dynamic perception without costly chain-of-thought or multi-round processing.

Discrete Diffusion Large Language Models (dLLMs) and Discrete Diffusion Multimodal Large Language Models (dMLLMs) ~\cite{nie2025large,dream2025,yu2025dimple,you2025llada,li2025lavida,yang2025mmada} have recently emerged as a promising direction. In tasks such as code generation~\cite{deepmind_gemini_diffusion,inceptionlabs_mercury}, planning~\cite{dream2025}, and Sudoku~\cite{dream2025}, dLLMs have been widely shown to achieve better performance than autoregressive models. Moreover, \cite{prabhudesai2025diffusionbeatsautoregressivedataconstrained} demonstrates that in a data-constrained setting, increasing the number of training FLOPs for dLLMs enables it to outperform AR.

In contrast to AR generation, discrete diffusion models treat generation as an iterative denoising process over discrete token sequences. This paradigm eliminates the left-to-right constraint and enables generation that is parallel and structurally controllable with bidirectional attention mechanism. Here are some unique properties of discrete diffusion.
\begin{itemize}
\item \textbf{Parallel Decoding:} Unlike AR models that decode one token at a time, discrete diffusion models generate multiple tokens simultaneously in each denoising step. This parallel decoding significantly accelerates inference speed.
\item \textbf{Better Controllability:} Discrete diffusion treats generation as a denoising or infilling task rather than unbounded left-to-right generation. This allows for precise control over output properties such as response length, format, and even the reasoning structure—by conditioning the generation on predefined templates.
\item \textbf{Dynamic Perception:} Enabled by bidirectional attention, discrete diffusion models continuously revise their perception of visual and linguistic context throughout the generation process. This facilitates adaptive comprehension that evolves with the response, overcoming the static, one-pass limitations of AR models.
\end{itemize}

Early efforts in discrete diffusion established the foundational mathematical formulations of discrete diffusion, introducing a token corruption schemes specifically designed for categorical data~\cite{austin2021structured,hoogeboom2021argmax}. These models demonstrated the feasibility of diffusion-based generation on various types of discrete data, including natural language~\cite{austin2021structured,zheng2023reparameterized}, images~\cite{austin2021structured}, and biological sequences such as DNA~\cite{sahoo2024simple}.
In this early stage, experiments were limited to models with around or fewer than 1 billion parameters. Through  simplifications and reparameterizations~\cite{zheng2023reparameterized,sahoo2024simple,shi2024simplified}, along with practical engineering efforts, the absorbing-state discrete diffusion formulation has gradually become the predominant mathematical framework adopted by open-source models and is termed as the masked diffusion model.

\begin{figure*}[t]
\centering
    \includegraphics[width=\linewidth]{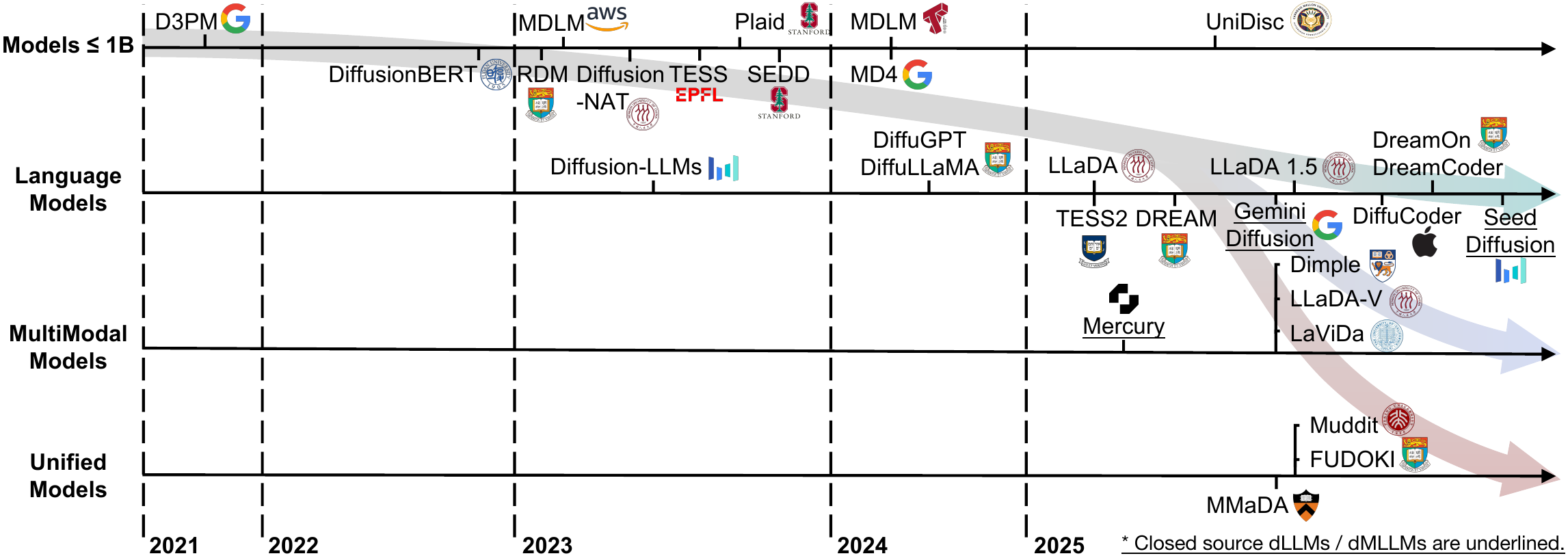}
    \caption{A timeline of existing dLLMs and dMLLMs in recent years. The timeline is established mainly according to the release date (e.g., the submission date to arXiv) of the technical paper for a model. The affiliation marked in the figure is based on the first affiliation listed in each paper.}
\label{fig_vein}
\vspace{-1em}
\end{figure*}

With the masked diffusion formulation, recent advances have significantly improved the scalability and effectiveness of discrete diffusion models~\cite{gong2024scaling,nie2025scalingmaskeddiffusionmodels}. A major breakthrough on the industrial front came with the presentation of discrete diffusion-based large language models by Inception Labs and Google, namely \textit{Mercury}~\cite{inceptionlabs_mercury} and \textit{Gemini Diffusion}~\cite{deepmind_gemini_diffusion}. These models reports comparable performance on code and mathematics benchmarks with their AR counterpart, while also achieving \textit{10$\times$} speedups in decoding, with about 1000 tokens per second.

In parallel, the research community has developed and open-sourced an increasing number of discrete diffusion-based language and multimodal models. The development began with dLLM models trained on large-scale text corpora, such as \textit{LLaDA}~\cite{nie2025large} and \textit{Dream}~\cite{dream2025}. 
Later, using the public available dLLMs as the backbones, dMLLMs, such as \textit{Dimple}~\cite{yu2025dimple}, \textit{LaViDa}~\cite{li2025lavida}, and \textit{LLaDA-V}~\cite{you2025llada}, are developed through multimodal alignment, instruction tuning, preference learning, and then reasoning enhancement. 

To provide a comprehensive framework for understanding discrete diffusion large language models (dLLMs) and discrete diffusion multimodal large language models (dMLLMs), this survey systematically explores recent advances in modeling, training, generation and applications of discrete diffusion techniques.

\tikzstyle{my-box}=[
 rectangle,
 rounded corners,
 text opacity=1,
 minimum height=1em,
 minimum width=5em,
 inner sep=2pt,
 align=center,
 fill opacity=.8,
 draw
]
\tikzstyle{leaf}=[
 my-box,
 minimum height=1.5em,
 text=black,
 align=left,
 font=\footnotesize,
 inner xsep=2pt,
 inner ysep=4pt,
]

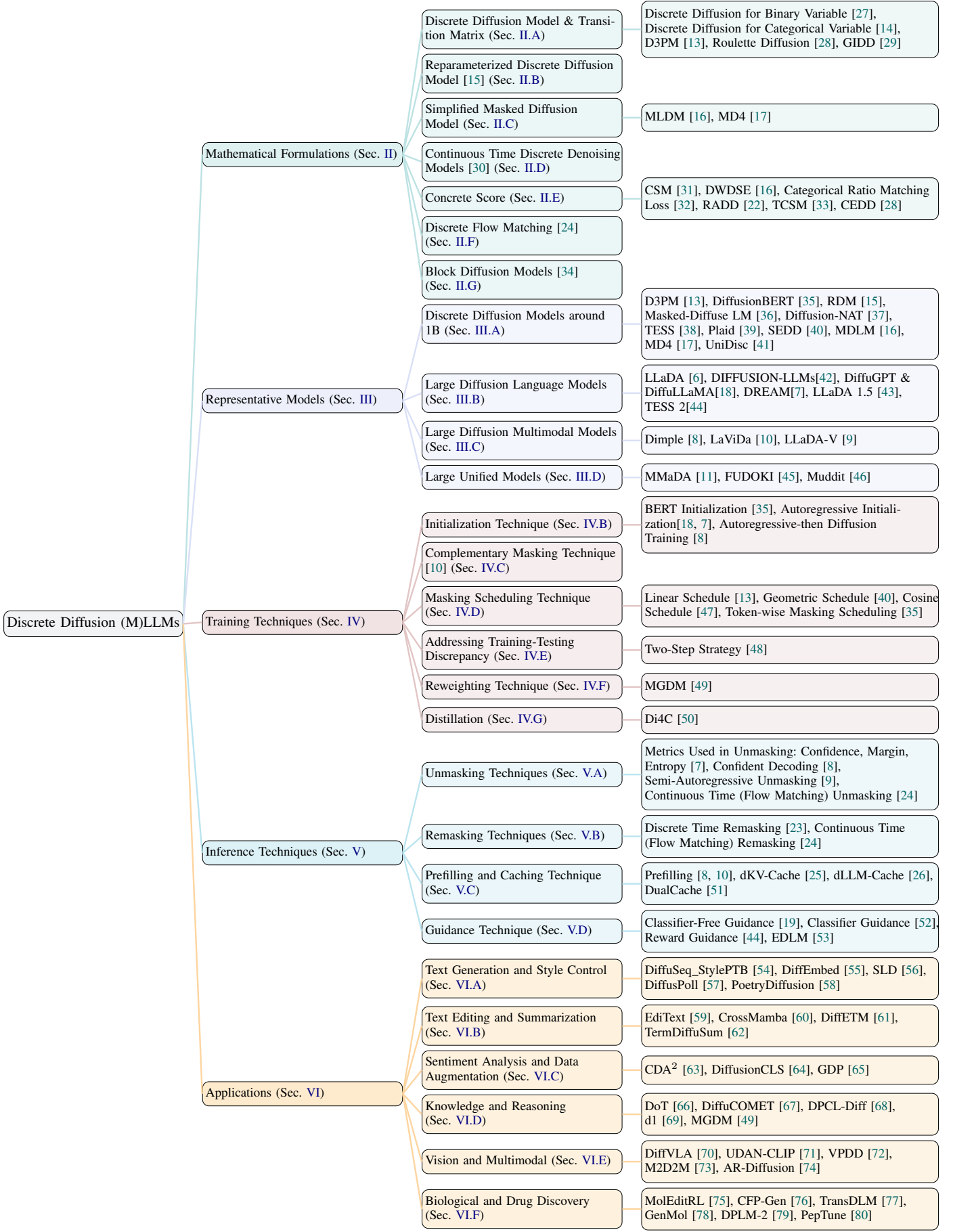
\begin{figure*}[p]
  \centering
  \resizebox{0.9\textwidth}{!}{
    \begin{forest}
        for tree={
            grow=east,
            reversed=true,
            anchor=base west,
            parent anchor=east,
            child anchor=west,
            base=left,
            font=\small,
            rectangle,
            draw=black,
            rounded corners,
            align=left,
            minimum width=4em,
            s sep=3pt,
            inner xsep=2pt,
            inner ysep=3pt,
            ver/.style={rotate=90, child anchor=north, parent anchor=south, anchor=center},
            edge={-, opacity=1.0, line width=1.pt},
        },
        where level=1{text width=13em,font=\footnotesize,edge={line width=1.pt}}{},
        where level=2{text width=12em,font=\footnotesize,edge={line width=1.pt}}{},
        where level=3{text width=20em,font=\footnotesize,edge={line width=1.pt}}{},
      [Discrete Diffusion (M)LLMs, fill=Gray!10, edge={Black!60}
        [Mathematical Formulations (\cref{math}), fill=lan!19, edge={lan!40}
          [Discrete Diffusion Model \& Transi-\\tion Matrix (\cref{math_1}), fill=lan!12, edge={lan!40}
            [Discrete Diffusion for Binary~\cite{sohl2015deep} and Categorical~\cite{hoogeboom2021argmax} Vari-\\able{,} D3PM~\cite{austin2021structured}{,} Roulette Diffusion~\cite{haxholli2025efficient}{,} GIDD~\cite{rutte2025generalized}, leaf, fill=lan!12, edge={lan!40}]
          ]
          [Simplified Masked Diffusion Model (\cref{math_smd}), fill=lan!12, edge={lan!40}, text width=17em
            [MLDM~\cite{sahoo2024simple}{,} MD4~\cite{shi2024simplified}, leaf, fill=lan!12, edge={lan!40}, text width=8em]
          ]
          [Continuous Time Discrete Denoising Models~\cite{continuoustimedlm} (\cref{math_3}), fill=lan!12, edge={lan!40}, text width=21em
          ]
          [Concrete Score (\cref{math_4}), fill=lan!12, edge={lan!40}, text width=10em
            [CSM~\cite{concrete_score}{,} DWDSE~\cite{sahoo2024simple}{,} Categorical Ratio Matching Loss~\cite{sun2023scorebased}{,} \\RADD~\cite{ou2025absorbingdiscretediffusionsecretly}{,} TCSM~\cite{zhang2025target}{,} CEDD~\cite{haxholli2025efficient}, leaf, fill=lan!12, edge={lan!40}, text width=22em]
          ]
          [Discrete Flow Matching~\cite{dfm} (\cref{math_5}), fill=lan!12, edge={lan!40}, text width=14em
          ]
          [Reparameterized Discrete Diffusion Model~\cite{zheng2023reparameterized} (Sec. Apdx.A.I), fill=lan!12, edge={lan!40}, text width=22em
          ]
        ]
        [Modeling Language Diffusion (\cref{modeling}), fill=red!10, edge={red!30}
          [Block Diffusion Models~\cite{arriola2025block}(\cref{modeling_1}){,} Flexible-Length Masked Diffusion~\cite{kim2025anyorderflexiblelengthmasked}(\cref{modeling_2}){,} \\Partial Masking~\cite{chao2025maskedunmaskeddiscretediffusion}(\cref{modeling_3}){,} DDOT~\cite{zhang2025flexiblelengthtextinfillingdiscrete}(\cref{modeling_4}), fill=red!10, edge={red!30}, text width=33.6em
          ]
        ]
        [Representative Models (\cref{models}), fill=mul!19, edge={mul!40}
          [Discrete Diffusion Models around\\1B (\cref{small_models}), fill=mul!12, edge={mul!40}
            [
            D3PM~\cite{austin2021structured}{,} DiffusionBERT~\cite{he2022diffusionbert}{,} RDM~\cite{zheng2023reparameterized}{,}Masked-Diffuse\\LM~\cite{chen2023cheaper}{,} Diffusion-NAT \cite{zhou2023diffusion}{,} TESS~\cite{mahabadi2023tess}{,} Plaid \cite{gulrajani2023likelihood}{,} SEDD\\\cite{lou2023discrete}{,} MDLM \cite{sahoo2024simple}{,} MD4 \cite{shi2024simplified}{,} UniDisc \cite{swerdlow2025unified}
            , leaf, fill=mul!12, edge={mul!40}]
          ]
          [Large Diffusion Language Models\\(\cref{llm}), fill=mul!12, edge={mul!40}
            [
            LLaDA \cite{nie2025large}{,}DIFFUSION-LLMs\cite{ye2023diffusion}{,}DiffuGPT\&DiffuLLaMA\\\cite{gong2024scaling}{,} DREAM\cite{dream2025}{,} LLaDA 1.5 \cite{zhu2025llada}{,} TESS 2\cite{tae2025tess}{,} DreamOn\\\cite{Dreamon2025}{,} DreamCoder \cite{dreamcoder2025}{,} DiffuCoder \cite{gong2025diffucoder}{,} Seed Diffusion \cite{song2025seed}
            , leaf, fill=mul!12, edge={mul!40}]
          ]
          [Large Diffusion Multimodal Models\\(\cref{mllm}), fill=mul!12, edge={mul!40}
            [
            Dimple \cite{yu2025dimple}{,} LaViDa \cite{li2025lavida}{,} LLaDA-V \cite{you2025llada}
            , leaf, fill=mul!12, edge={mul!40}]
          ]
          [Large Unified Models (\cref{unified_models}), fill=mul!12, edge={mul!40}
            [
            MMaDA \cite{yang2025mmada}{,} FUDOKI \cite{wang2025fudoki}{,} Muddit \cite{shi2025muddit}
            , leaf, fill=mul!12, edge={mul!40}]
          ]
        ]
        [Training Techniques (\cref{training}), fill=uni!19, edge={uni!40}
          [Initialization Technique (\cref{initialization}), fill=uni!12, edge={uni!40}
            [
            BERT Initialization~\cite{he2022diffusionbert}{,} Autoregressive Initialization\cite{gong2024scaling,dream2025}{,} \\Autoregressive-then Diffusion Training~\cite{yu2025dimple}
            , leaf, fill=uni!12, edge={uni!40}]
          ]
          [Complementary Masking\cite{li2025lavida}(\cref{comp_masking}), fill=uni!12, edge={uni!40}, text width=13em
          ]
          [Masking Scheduling Technique\\(\cref{masking_scheduling}), fill=uni!12, edge={uni!40}
            [
            Linear~\cite{austin2021structured}{,}Geometric~\cite{lou2023discrete,shi2024simplified}{,}Cosine~\cite{chang2022maskgit}{,}Polynomial~\cite{shi2024simplified}{,} \\Token-wise Masking Scheduling~\cite{he2022diffusionbert}{,} Blockwise SFT~\cite{sun2025blockwisesftdiffusionlanguage}
            , leaf, fill=uni!12, edge={uni!40}]
          ]
          [Reweighting Technique (\cref{training-rw}), fill=uni!12, edge={uni!40}
            [
            MGDM~\cite{ye2025beyond}
            , leaf, fill=uni!12, edge={uni!40}]
          ]
            [Distillation (\cref{distillation}), fill=uni!12, edge={uni!40}
            [
            Di4C~\cite{hayakawa2024distillation}
            , leaf, fill=uni!12, edge={uni!40}]
          ]
            [Reinforcement Learning (\cref{rl}), fill=uni!12, edge={uni!40}
            [
            UniGRPO~\cite{yang2025mmada}{,}VRPO~\cite{zhu2025llada}{,}SDPO~\cite{han2025discretediffusiontrajectoryalignment}{,}wd1~\cite{tang2025wd1weightedpolicyoptimization}{,}DCoLT~\cite{huang2025reinforcingdiffusionchainlateral}
            , leaf, fill=uni!12, edge={uni!40}]
          ]
          [Training-Testing Input Discrepancy (Appendix Sec.C.III), fill=uni!12, edge={uni!40}, text width=21em
            [
            Two-Step Strategy~\cite{asada-miwa-2025-addressing}
            , leaf, fill=uni!12, edge={uni!40}, text width=11em]
          ]
        ]
        [Inference Techniques (\cref{inference}), fill=SkyBlue!19, edge={SkyBlue!40}
          [Unmasking Techniques (\cref{unmask}), fill=SkyBlue!12, edge={SkyBlue!40}
            [
            Metrics Used in Unmasking: Confidence{,} Margin{,} Entropy~\\\cite{dream2025,kim2025train}{,} EB-Sampler~\cite{benhamu2025acceleratedsamplingmaskeddiffusion}{,} PC-Sampler.~\cite{huang2025pcsamplerpositionawarecalibrationdecoding}{;} Confident \\Decoding~\cite{yu2025dimple}{,} Block-wise~\cite{you2025llada}{,} DUS~\cite{luxembourg2025planspeeddilatedscheduling}{,} Continuous Time \\(Flow Matching) Unmasking~\cite{dfm}
            , leaf, fill=SkyBlue!12, edge={SkyBlue!40}]
          ]
          [Remasking Techniques (\cref{remask}), fill=SkyBlue!12, edge={SkyBlue!40}
            [
            Discrete Time Remasking~\cite{wang2025remasking}{,}Wide\ In Narrow\ Out~\cite{hong2025wideinnarrowoutrevokabledecoding}{} \\Continuous Time (Flow Matching) Remasking~\cite{dfm}
            , leaf, fill=SkyBlue!12, edge={SkyBlue!40}]
          ]
          [Prefilling and Caching Technique \\(\cref{prefilling}), fill=SkyBlue!12, edge={SkyBlue!40}
            [
            Prefilling~\cite{yu2025dimple,li2025lavida}{,} dKV-Cache~\cite{ma2025dkv}{,} dLLM-Cache~\cite{liu2025dllm}{,} \\DualCache~\cite{wu2025fastdllm}
            , leaf, fill=SkyBlue!12, edge={SkyBlue!40}]
          ]
          [Guidance Technique (\cref{guidance}), fill=SkyBlue!12, edge={SkyBlue!40}
            [
            Classifier-Free Guidance~\cite{nie2025scalingmaskeddiffusionmodels}{,} Classifier Guidance~\cite{huang2025ctrldiffboostinglargediffusion}{,} \\Reward Guidance~\cite{tae2025tess}{,} EDLM~\cite{xu2025energybased}
            , leaf, fill=SkyBlue!12, edge={SkyBlue!40}]
          ]
          [Sampling Technique (\cref{sampling}), fill=SkyBlue!12, edge={SkyBlue!40}
            [
            Early Stopping~\cite{li2025diffusionlanguagemodelsknow,jin2025thinkinginsidemaskinplace}{,} Particle Gibbs Sampling~\cite{dang2025inferencetimescalingdiffusionlanguage}{,} \\Temporal Self-Consistency~\cite{wang2025timefeatureexploitingtemporal}, leaf, fill=SkyBlue!12, edge={SkyBlue!40}]
          ]
          [
          Context Length Extension (\cref{context}), fill=SkyBlue!12, edge={SkyBlue!40}, text width=13em
            [
            LongLLaDA~\cite{liu2025longlladaunlockinglongcontext}, leaf, fill=SkyBlue!12, edge={SkyBlue!40}, text width=19em
            ]
          ]
          [Sparse Computation (\cref{sparse}), fill=SkyBlue!12, edge={SkyBlue!40}
            [
            Sparse-dLLM~\cite{song2025sparsedllmacceleratingdiffusionllms}{,} DPad~\cite{chen2025dpadefficientdiffusionlanguage}{,}Pipelined Parallel~\cite{wang2025diffusion}, leaf, fill=SkyBlue!12, edge={SkyBlue!40}]
          ]
          [Response Length Control (\cref{length}), fill=SkyBlue!12, edge={SkyBlue!40}, text width=13em
            [
            DAEDAL~\cite{li2025fixedtrainingfreevariablelengthdenoising}, leaf, fill=SkyBlue!12, edge={SkyBlue!40}, text width=19em]
          ]
        ]
        [Quantization (\cref{sec:quantization}), fill=SpringGreen!19, edge={SpringGreen!40}
          [Quantization Meets dLLMs\cite{lin2025quantization}{,} DLLMQuant \cite{xu2025dllmquant}, fill=SpringGreen!12, edge={SpringGreen!40}, text width=17em
          ]
        ]
        [Privacy and Safety (\cref{sec:safety}), fill=Goldenrod!19, edge={Goldenrod!40}
          [            DIJA \cite{wen2025devilmaskemergentsafety}{,} PAD \cite{zhang2025jailbreaking}{,} MOSA \cite{xie2025start}, fill=Goldenrod!12, edge={Goldenrod!40}
          ]
        ]
        [Applications (\cref{app}), fill=YellowOrange!19, edge={YellowOrange!40}
          [Language Related Applications \\(\cref{app_1}), fill=YellowOrange!12, edge={YellowOrange!40}
            [DiffuSeq\_StylePTB \cite{lyu2023fine}{,} DiffEmbed  \cite{zhang2025diffusion}{,} SLD \cite{zhu2024segment}{,} \\DiffusPoll \cite{cheng2024diffuspoll}{,} PoetryDiffusion \cite{hu2024poetrydiffusion}{,} SVDD \cite{padole2025improving}{,} \\EdiText \cite{lee2025editext}{,} CrossMamba \cite{do2025discrete}{,} DiffETM \cite{shao2025diffetm}{,} \\TermDiffuSum \cite{dong2025termdiffusum}{,} $\textrm{CDA}^2$ \cite{xin2025cdaˆ2}{,} DiffusionCLS \cite{chen2024effective}{,} GDP \\\cite{zhu2024pinpointing}{,} Layout-Corrector \cite{iwai2024layout}
            , leaf, fill=YellowOrange!12, edge={YellowOrange!40}]
          ]
          [Knowledge and Reasoning \\(\cref{app_2}), fill=YellowOrange!12, edge={YellowOrange!40}
            [
            DoT \cite{ye2024diffusion}{,} DiffuCOMET \cite{gao2024diffucomet}{,} DPCL-Diff \cite{cao2025dpcl}{,} \\d1 \cite{zhao2025d1}{,} MGDM~\cite{ye2025beyond}{,} NeSyDM~\cite{van2025neurosymbolic}, leaf, fill=YellowOrange!12, edge={YellowOrange!40}]
          ]
          [Vision and Multimodal (\cref{app_3}), fill=YellowOrange!12, edge={YellowOrange!40}
            [
            UDAN-CLIP \cite{shaahid2025underwater}{,} M2D2M \cite{chi2024m2d2m}{,} AR-Diffusion \cite{sun2025ar}
            , leaf, fill=YellowOrange!12, edge={YellowOrange!40}]
          ]
          [Robotics and Autonomous Driving \\(\cref{app_4}), fill=YellowOrange!12, edge={YellowOrange!40}
            [
            DiffVLA \cite{jiang2025diffvla}{,} DDVLA \cite{liang2025discrete}{,} ViLaD \cite{cui2025vilad}{,} VPDD \cite{he2024learning}{,} \\DGD \cite{liang2025discreteGuided}
            , leaf, fill=YellowOrange!12, edge={YellowOrange!40}]
          ]
          [Graph and Structured Predictions \\(\cref{app_5}), fill=YellowOrange!12, edge={YellowOrange!40}
            [
            LO-ARM \cite{wang2025learning}{,} ReDiSC \cite{li2025redisc}{,} Scaffold Diffusion \cite{jung2025scaffolddiffusionsparsemulticategory}
            , leaf, fill=YellowOrange!12, edge={YellowOrange!40}]
          ]
          [Biological and Drug Discovery \\(\cref{app_8}), fill=YellowOrange!12, edge={YellowOrange!40}
            [
            MolEditRL \cite{zhuang2025moleditrl}{,} CFP-Gen \cite{yin2025cfp}{,} TransDLM \cite{xiong2024text}{,} \\GenMol \cite{lee2025genmol}{,} DPLM-2 \cite{wang2024dplm}{,} PepTune \cite{tang2025peptune}{,} \\CMCM-DLM \cite{zhang2025cross}
            , leaf, fill=YellowOrange!12, edge={YellowOrange!40}]
          ]
        ]
      ]
    \end{forest}
  }
\caption{Overview of our survey. We begin by introducing the mathematical foundations (\cref{math}) and modeling methods (\cref{modeling}) of discrete diffusion language models . Next, we present a high-level overview of representative base models (\cref{models}), followed by discussions on training strategies (\cref{training}), inference techniques (\cref{inference}) and quantization (\cref{sec:quantization}). Furthermore, we discuss the privacy and safety studies of discrete diffusion language models (\cref{sec:safety}). In addition, we also review a wide range of applications (\cref{app}) that adopt discrete diffusion language models as their core model.}
\label{fig:tree}
\end{figure*}

In the rest of this paper, \cref{math} presents the mathematical foundations of discrete diffusion models. \cref{modeling} lists several modeling techniques for the discrete diffusion task. These techniques build upon mathematical formulations, primarily aiming to enhance model flexibility or introduce additional capabilities.
\cref{models} surveys representative discrete diffusion language models across varying scales. This includes early-stage models, scaled dLLMs, dMLLMs, and unified models. \cref{training} discusses the key training strategies used in dLLMs and dMLLMs. \cref{inference} lists various inference techniques used in dLLMs and dMLLMs. \cref{sec:safety} discusses trustworthy issues in dLLMs. \cref{sec:quantization} discusses the quantization techniques of dLLMs.  \cref{app} reviews the broad range of applications powered by discrete diffusion models. Finally, Appendix Sec.F summarize potential directions for future research. The organization of this survey is illustrated in \cref{fig:tree}.

\section{Mathematical Formulations}
\label{math}
In this section, we discuss the mathematical formulation of the discrete diffusion.
\subsection{Discrete Diffusion Model and Transition Matrix}
\label{math_1}
Diffusion models with discrete state spaces were initially introduced in \cite{sohl2015deep} for binary variables, and later extended to categorical variables in \cite{hoogeboom2021argmax}. Based on the previous works, Discrete Denoising Diffusion Probabilistic Models (D3PMs) \cite{austin2021structured} provides a general and flexible framework.

Let $x_0 \sim q(x_0)$ denote the data distribution over sequences composed of $K$ categorical values. The D3PM framework defines a forward Markov process $q(x_{1:T} \mid x_0)$ that gradually corrupts the data into noise, and a parameterized reverse process $p_\theta(x_{0:T})$ that learns to denoise:
\begin{align}
    q(x_{1:T} \mid x_0) &= \prod_{t=1}^{T} q(x_t \mid x_{t-1}), \\
    p_\theta(x_{0:T}) &= p(x_T) \prod_{t=1}^{T} p_\theta(x_{t-1} \mid x_t).
\end{align}

Each $x_t$ is a discrete random variable and $q(x_t \mid x_{t-1})$ is defined via a time-dependent transition matrix $Q_t$, with categorical transition probabilities:
\begin{equation}
    q(x_t \mid x_{t-1}) = \text{Cat}(x_t; p = x_{t-1} Q_t),
\end{equation}
where $x_{t-1}$ is a one-hot vector and $Q_t \in \mathbb{R}^{K \times K}$ is a row-stochastic matrix. The marginal distribution $q(x_t \mid x_0)$ and the posterior $q(x_{t-1} \mid x_t, x_0)$ are:
\begin{align}
    q(x_t \mid x_0) &= \text{Cat}(x_t; p = x_0 Q_{1:t}),\\ \quad Q_{1:t} &= Q_1 Q_2 \dots Q_t, \\
    q(x_{t-1} \mid x_t, x_0) &= \text{Cat}\left(x_{t-1}; p = \frac{x_t Q_t^\top \circ x_0 Q_{1:t-1}}{x_0 Q_{1:t} x_t^\top} \right).
\end{align}

 D3PM framework support various types of transition matrices $Q_t$, each inducing a different corruption behavior. Here, we present the two most commonly used transition matrices: uniform and absorbing. Additional types, including hybrid, discretized Gaussian, band-diagonal, and embedding-based, are described in Appendix Section A.I.
\begin{itemize}
    \item \textbf{Uniform}: $Q_t = (1 - \beta_t) I + \frac{\beta_t}{K} \mathbf{1} \mathbf{1}^\top$ yields a uniform stationary distribution. The uniform transition matrix looks like
    \begin{equation}
        Q_t^{\text{uniform}} = \begin{bmatrix}
            1 - \frac{K-1}{K}\beta_t & \frac{\beta_t}{K} & \cdots & \frac{\beta_t}{K} \\
            \frac{\beta_t}{K} & 1 - \frac{K-1}{K}\beta_t & \cdots & \frac{\beta_t}{K} \\
            \vdots & \vdots & \ddots & \vdots \\
            \frac{\beta_t}{K} & \frac{\beta_t}{K} & \cdots & 1 - \frac{K-1}{K}\beta_t
            \end{bmatrix}.
    \end{equation}
    \item \textbf{Absorbing}: $Q_t = (1 - \beta_t) I + \beta_t \mathbf{1} e_m^\top$,
    where \( e_m \) is a vector with a one on the absorbing state and zeros elsewhere.
    Tokens either remain unchanged or are mapped to a special [MASK] token with probability $\beta_t$.
    The absorbing transition matrix looks like
        \begin{equation}
            Q_t^{\text{absorb}} = \begin{bmatrix}
            1-\beta_t & 0 & \cdots & 0 & \beta_t \\
            0 & 1-\beta_t & \cdots & 0 & \beta_t \\
            \vdots & \vdots & \ddots & \vdots & \vdots \\
            0 & 0 & \cdots & 1-\beta_t & \beta_t \\
            0 & 0 & \cdots & 0 & 1
            \end{bmatrix}.
        \end{equation}
\end{itemize}

Following the $x_0$-parameterization, the model predicts $p_\theta(x_{t-1} \mid x_t)$ using:
\begin{equation}
    p_\theta(x_{t-1} \mid x_t) \propto \sum_{\tilde{x}_0} q(x_{t-1}, x_t \mid \tilde{x}_0) \tilde{p}_\theta(\tilde{x}_0 \mid x_t),
\end{equation}
where $\tilde{p}_\theta(\tilde{x}_0 \mid x_t)$ is a network predicting logits over $x_0$. This parameterization ensures the reverse distribution respects the sparsity pattern of $Q_t$.

The loss function combines the variational lower bound $\mathcal{L}_{\mathrm{vb}}$ with an auxiliary cross-entropy denoising term $\mathcal{L}_{\mathrm{ce}}$:
\begin{align}
    \mathcal{L}_\lambda &= \mathcal{L}_{\mathrm{vb}} + \lambda \, \mathcal{L}_{\mathrm{ce}},\\
    \mathcal{L}_{\mathrm{vb}} &= \mathbb{E}_{q(x_0)} \Big[
\underbrace{\mathrm{KL}(q(x_T \mid x_0) \, \| \, p(x_T))}_{\mathcal{L}_T} \nonumber\\ 
&\quad + \underbrace{\sum_{t=2}^{T} \mathbb{E}_{q(x_t \mid x_0)} \left[
\mathrm{KL}(q(x_{t-1} \mid x_t, x_0) \, \| \, p_\theta(x_{t-1} \mid x_t))\right]}_{\mathcal{L}_{t-1}} \nonumber\\
&\qquad - \underbrace{\mathbb{E}_{q(x_1 \mid x_0)} \left[ \log p_\theta(x_0 \mid x_1) \right]}_{\mathcal{L}_0}\Big],
\\
    \mathcal{L}_{\mathrm{ce}} &= \mathbb{E}_{q(x_0)} \mathbb{E}_{q(x_t \mid x_0)}[-\log \tilde{p}_\theta(x_0 \mid x_t)].
\end{align}
In $\mathcal{L}_{\mathrm{vb}}$, 
\begin{itemize}
    \item $\mathcal{L}_T$ is the KL divergence between the forward terminal distribution $q(x_T \mid x_0)$ and the prior $p(x_T)$,
    \item $\mathcal{L}_{t-1}$ is the KL divergence between the forward posterior $q(x_{t-1} \mid x_t, x_0)$ and the learned reverse model $p_\theta(x_{t-1} \mid x_t)$ at each intermediate step,
    \item $\mathcal{L}_0$ is the cross-entropy loss for reconstructing $x_0$ from $x_1$ using the reverse model.
\end{itemize}
Such decomposition enables the model to be trained efficiently by sampling time steps $t$ uniformly and estimating each term using stochastic gradient descent. The forward posterior $q(x_{t-1} \mid x_t, x_0)$ has a closed-form expression under categorical diffusion, and the model is typically parameterized to predict $p_\theta(x_0 \mid x_t)$, from which $p_\theta(x_{t-1} \mid x_t)$ is derived analytically.
The auxiliary $\mathcal{L}_{\mathrm{ce}}$ is added to encourage accurate prediction of the original data $x_0$ from corrupted samples.

Besides a flexible representation of discrete diffusion, D3PM also unifies various paradigms such as BERT, autoregressive models, and masked language models within the discrete diffusion framework.

\subsection{Simplified Masked Diffusion Model}
\label{math_smd}

A widely adopted class of discrete diffusion models is based on the absorbing state and is often referred to as the \textit{Masked Diffusion Model}. Both \cite{shi2024simplified,sahoo2024simple} introduced simplifications to the diffusion process and the corresponding training objective for masked diffusion, yielding improved performance and computational efficiency. For the simplification of general discrete diffusion, we discuss Reparameterized Discrete Diffusion Models (RDMs)~\cite{zheng2023reparameterized} in Appendix Sec.A.II.

In masked diffusion, the forward process progressively replaces input tokens with a special \texttt{[MASK]} token. Once a token is masked, it remains in that state throughout the remainder of the process, making the \texttt{[MASK]} token an absorbing state. At each time step $t$, the forward transition for a token $x$ is defined as:
\begin{equation}
q(x_t \mid x_0) = \text{Cat}(x_t; \alpha_t x_0 + (1 - \alpha_t)m),
\end{equation}
where $m$ is the one-hot vector corresponding to the \texttt{[MASK]} token, and $\alpha_t \in [0,1]$ is the monotonically decreasing schedule such that $\alpha_0 \approx 1$ and $\alpha_T = 0$.

The reverse process aims to denoise the masked sequence by substituting \texttt{[MASK]} tokens with predicted tokens. Importantly, unmasked tokens are carried out unchanged throughout the denoising process.
The reverse posterior at a previous time $s$ conditioned on $x_t$ and $x_0$ is given by:
\begin{align}
&q(x_s \mid x_t, x_0) = \nonumber\\
&\begin{cases}
\text{Cat}(x_s; x_t), & \text{if } x_t \neq m, \\
\text{Cat} \left( x_s; \dfrac{(1 - \alpha_s)m + (\alpha_s - \alpha_t) x_0}{1 - \alpha_t} \right), & \text{if } x_t = m.
\end{cases}
\end{align}
This formulation reflects two key properties of the masking process:
(1) If the current token $x_t$ is not masked, the posterior is deterministic: $x_s = x_t$.
(2) If the token is masked, then the posterior is a linear interpolation between the mask vector $m$ and the clean token $x_0$, scaled by the noise schedule parameters $\alpha_s$ and $\alpha_t$.

Let $f_\theta(x_t)$ be the neural network output predicting the original token $x_0$ from the noisy input $x_t$. The above reverse transition is rewritten as:
\begin{equation}
p_\theta(x_s \mid x_t) =
\begin{cases}
\text{Cat}(x_s; x_t), & \text{if } x_t \neq m, \\
\text{Cat}\left(x_s; \frac{(1 - \alpha_s)m + (\alpha_s - \alpha_t) f_\theta(x_t)}{1 - \alpha_t} \right), & \text{if } x_t = m.
\end{cases}
\end{equation}

The variational lower bound for discrete diffusion can be simplified using the above formulation, leading to an final loss in the form of 
\begin{align}
\mathcal{L} &= \sum_{t=2}^T \mathbb{E}_{x_0, x_{1:T}} \Big[ - \frac{\alpha_{t-1} - \alpha_t}{1 - \alpha_t} \sum_{n=1}^N \delta_m(x_{t,n}) x_{0,n} \log [f_{\theta}(x_{t})]_n
\Big],\label{eq:simp_loss}
\end{align}
where $\delta_m(x_{t,n})$ denotes the indicator function, $\delta_m(x_{t,n})=1$, if the n-th token of $x_t$ is a masked token, otherwise, $\delta_m(x_{t,n})=0$. 

By defining the discrete time series as $0, \frac{1}{T}, \ldots, 1 - \frac{1}{T}, 1$ and letting $T \to \infty$, both works~\cite{sahoo2024simple,shi2024simplified} extend the above loss \cref{eq:simp_loss} to the continuous-time setting:
\begin{align}
\mathcal{L} &=  \int_0^1 \mathbb{E}_{x_{0:1}}\Big[ \frac{\alpha_t'}{1 - \alpha_t}\sum_{n=1}^N \delta_m(x_{t,n}) x_{0,n} \log [f_{\theta}(x_{t})]_n
\Big]dt.\label{eq:cont_simp_loss}
\end{align}

This loss corresponds to a reweighted cross-entropy loss evaluated only over masked tokens.
Such loss formulation is significantly simpler than the original variational bound and has become the standard training objective for subsequent large discrete diffusion models. 

\subsection{Continuous Time Discrete Denoising Models}
\label{math_3}
D3PM operates in discrete time, \textit{i.e.}, with time steps $t = 0, 1, 2, \ldots, T$. \cite{continuoustimedlm} describes a continuous-time framework for discrete denoising models, formulated as a Continuous-Time Markov Chain (CTMC), where $t \in [0, T]$. This approach generalizes discrete diffusion models by allowing transitions at arbitrary time points. The infinitesimal transition probabilities are given by:
\begin{equation}
q_{t|t-\Delta t}(x' \mid x) = \delta_{x,x'} + R_t(x, x') \Delta t + o(\Delta t).
\end{equation}

This process converges to a tractable reference distribution as $t \to T$. The time-reversed generative process is another CTMC with reverse rate matrix $\hat{R}_t$, expressed as:
\begin{equation}
\hat{R}_t(x, x') = R_t(x', x) \sum_{x_0} \frac{q_{t|0}(x' \mid x_0)}{q_{t|0}(x \mid x_0)} p_\theta(x_0 \mid x),
\end{equation}
where $p_\theta(x_0 \mid x)$ is a learnable denoising model approximating $q_{0|t}(x_0 \mid x)$.

The training of $p_\theta(x_0 \mid x)$ is guided by a continuous-time version of the variational lower bound. Let $Z_t(x) = \sum_{x' \ne x} R_t(x, x')$ be the total outgoing rate and $r_t(x' \mid x) = R_t(x, x') / Z_t(x)$ the normalized jump probability. The continuous-time variational lower bound is:
\begin{align}
\mathcal{L}_{\mathrm{vb}}(\theta) =&
  T \, \mathbb{E}_{t \sim \mathcal{U}(0, T)}
      \mathbb{E}_{x \sim q_t}
      \mathbb{E}_{x' \sim r_t(\cdot \mid x)}
      \Bigl[\nonumber\\
      &\sum_{x'' \ne x} \hat{R}_t^\theta(x, x'')  
  -  Z_t(x)\,\log \hat{R}_t^\theta(x', x)
      \Bigr]
  + C,
\end{align}
where $C$ is constant with respect to $\theta$. This objective can be efficiently optimized using stochastic gradient descent by sampling $(x, x')$ pairs according to the forward process.

During inference, however, the exact simulation of the reverse CTMC can be computationally prohibitive. Instead, the tau-leaping algorithm [reference] approximates the reverse process by applying multiple transitions within a time interval $\tau$ simultaneously. For a current state $x_t$, the number of transitions to each $x'$ during $[t - \tau, t]$ is modeled as:
\begin{equation}
P_{x'} \sim \mathrm{Poisson}(\tau \cdot \hat{R}_t^\theta(x_t, x')).
\end{equation}

The next state $x_{t-\tau}$ is obtained by aggregating the sampled transitions. This method supports parallel decoding by allowing simultaneous updates across multiple dimensions.

To further improve sample fidelity, predictor-corrector steps is used. After a tau-leaping step, corrector transitions with rate matrix $R_c = R_t + \hat{R}_t^\theta$ are applied to refine the sample distribution toward the target marginal $q_t(x)$. This approach is analogous to Langevin correctors in continuous diffusion models.

\subsection{Concrete Score}
\label{math_4}
In the continuous-time discrete diffusion framework, as formulated in previous \cite{continuoustimedlm}, the reverse process can also be analytically expressed in terms of the forward transition rate matrix and a function known as the concrete score~\cite{concrete_score}. This construction enables training via score matching, analogous to score-based models in continuous diffusion model.

Let $R_t(x, x')$ be the forward transition rate matrix of a continuous-time Markov chain (CTMC) over a discrete state space $\mathcal{X}$. The reverse-time transition rate $\hat{R}_t(x, x')$ can be formulated as:
\begin{equation}
\hat{R}_t(x, x') =
\begin{cases}
\displaystyle \frac{p_t(x')}{p_t(x)} R_t(x', x), & x' \ne x, \\
\displaystyle - \sum_{k \ne x} \hat{R}_t(x, k), & x' = x.
\end{cases}
\end{equation}
Here, the scalar ratio $\frac{p_t(x')}{p_t(x)}$ is referred to as the concrete score. It quantifies the relative likelihood of two discrete states at time $t$ and modulates the reverse transition rate accordingly. 
Thus, instead of learning the full reverse transition distribution, one can train a model $s_\theta(x, t)$ to estimate the concrete score:
\begin{equation}
s_\theta(x, t) \approx \left[ \frac{p_t(x')}{p_t(x)} \right]_{x' \in \mathcal{X}}.
\end{equation}

In the Appendix Sec.A.III, we discuss the commonly-used training loss under concrete score formulation, the connection of concrete score with traditional cross-entropy loss and the time independency simplification of concrete score.

\subsection{Discrete Flow Matching (DFM)}
\label{math_5}
Build upon the continuous-time Markov chain (CTMC) framework in Continuous Time Discrete Denoising Models~\cite{continuoustimedlm}, Discrete Flow Matching (DFM)~\cite{dfm} extends the Flow Matching paradigm to categorical sequence data. The model defines a probability path $p_t$ interpolating between a source distribution $p$ (e.g., all-mask sequences) and a target distribution $q$ (e.g., the data distribution), such that $p_0 = p$ and $p_1 = q$.

Given a coupling distribution $\pi(x_0, x_1)$ between source and target sequences, the marginal probability at time $t$ is defined as:
\begin{equation}
p_t(x) = \sum_{x_0, x_1 \in D} p_t(x \mid x_0, x_1)\pi(x_0, x_1),
\end{equation}
where the conditional path $p_t(x \mid x_0, x_1)$ factorizes over positions:
\begin{equation}
p_t(x \mid x_0, x_1) = \prod_{i=1}^N p_t(x^i \mid x_0, x_1),
\end{equation}
with token-level conditional paths defined as a convex combination of basis distributions:
\begin{equation}
p_t(x^i \mid x_0, x_1) = \sum_{j=1}^m \kappa^i_{t,j} w_j(x^i \mid x_0, x_1),
\end{equation}
where $\kappa^i_{t,j} \geq 0$, $\sum_j \kappa^i_{t,j} = 1$ form a scheduler controlling the path dynamics.

The generative process is defined via probability velocity fields $\{u^i_t\}_{i=1}^N$ guiding transitions between states. The update rule for sampling is:
\begin{equation}
x^i_{t+h} \sim \delta_{x^i_t}(\cdot) + h u^i_t(\cdot, x_t),
\end{equation}
where $u_t$ is said to generate $p_t$ if the process satisfies:
\begin{equation}
p_{t+h}(x) = p_t(x) - h\, \mathrm{div}_x(p_t u_t) + o(h),
\end{equation}
with the discrete divergence operator:
\begin{equation}
\mathrm{div}_x(v) = \sum_{z \in D} \left[v(z, x) - v(x, z)\right].
\end{equation}

For the convex interpolation path:
\begin{equation}
p_t(x^i \mid x_0, x_1) = (1 - \kappa_t) \delta_{x_0}(x^i) + \kappa_t \delta_{x_1}(x^i),
\end{equation}
the corresponding generating velocity takes the closed form:
\begin{equation}
u^i_t(x^i, z) = \frac{\dot{\kappa}_t}{1 - \kappa_t} \left[p_{1|t}(x^i \mid z) - \delta_z(x^i)\right],
\end{equation}
where $p_{1|t}(x^i \mid z)$ is the probability denoiser: the conditional probability of the target token $x^i_1$ given the current state $z$.

To estimate the posteriors required in the generative process, the model minimizes the cross-entropy loss:
\begin{equation}
\mathcal{L}(\theta) = -\sum_{i=1}^N \mathbb{E}_{t, (x_0, x_1), x_t} \left[\log p_{1|t}(x^i_1 \mid x_t; \theta)\right],
\end{equation}
where $x_t \sim p_t(\cdot \mid x_0, x_1)$.

\section{Modeling Language Diffusion}
\label{modeling}
Beyond the mathematical formulation introduced in the previous section, in this section, we present several techniques used when modeling the language diffusion task, such as specialized neural network designs, and extra sub-tasks. These techniques are introduced to enhance the controllability and flexibility.

\subsection{Block Diffusion Models}
\label{modeling_1}
Block Diffusion models (BD3-LMs)~\cite{arriola2025block} provide a hybrid framework that interpolates between autoregressive language models and fully parallel diffusion models. Instead of denoising all tokens simultaneously, BD3-LMs segment the sequence into blocks and perform discrete denoising diffusion within each block, while conditioning autoregressively on all preceding blocks. The mathematical formualtion of BD3-LMs are provided in Appendix Sec.B.I.

\subsection{Flexible-Length Masked Diffusion}
\label{modeling_2}
Flexible-Length Masked Diffusion (FlexMDM)~\cite{kim2025anyorderflexiblelengthmasked} extends masked diffusion models to handle variable-length sequences by introducing a third token state, called empty, in addition to the usual masked and ground truth states. This allows the model not only to recover missing tokens (unmasking) but also to insert new tokens into the sequence. During inference, the process alternates between two steps: first, the model predicts how many new mask tokens should be inserted before each existing token, effectively expanding the sequence length; then, it replaces all mask tokens with actual content tokens. By repeating this insertion–unmasking cycle, FlexMDM can gradually generate longer sequences, enabling it to produce variable-length outputs while maintaining the flexibility of generating tokens in any order. In Appendix Sec. B.II, we provide the mathematical formulation of FlexMDM along with a more detailed description of its generation process.

\subsection{Partial Masking}
\label{modeling_3}
In masked diffusion model, each token has only two states. Partial Masking~\cite{chao2025maskedunmaskeddiscretediffusion} introduces an additional \emph{intermediate state} by decomposing each token into a sequence of sub-tokens.

Each original token $x^i_0 \in \mathcal{X}$, where $\mathcal{X} = \{0, \ldots, C-1\}$, is mapped into a sub-token sequence $y^i_0 = [y^{i,1}_0, \ldots, y^{i,\ell}_0] \in \mathcal{Y}^\ell$ through an invertible base-$b$ encoding function $f$: $
f: \mathcal{X} \to \mathcal{Y}^\ell$, $y^i_0 = f(x^i_0)$,
where $\mathcal{Y} = \{0, \ldots, b-1\}$ and $b = \lceil \sqrt[\ell]{C} \rceil$.  
The inverse mapping $f^{-1}$ reconstructs tokens from sub-tokens, ensuring lossless transformation:
$ x^i_0 = f^{-1}(y^i_0)$.
Consequently, the forward pass, backward pass, training, and inference of diffusion with partial masking all occur at the sub-token level.

\subsection{Diffusion with Optimal Transport Position Coupling}
\label{modeling_4}
Standard discrete diffusion models relies on fixed token positions during generation, which prevents them from handling flexible-length or flexible-position text infilling. Discrete Diffusion with Optimal Transport Position Coupling (DDOT)~\cite{zhang2025flexiblelengthtextinfillingdiscrete} jointly denoises both token values and token positions, enabling dynamic sequence restructuring while preserving relative ordering.
During sampling, DDOT alternates between token denoising and position refinement:
\begin{enumerate}
    \item Predict token distributions $s_\theta(x_t,t)$ and replace masks accordingly.
    \item Predict position velocities $v_\theta(z_t,t)$ and update the position variable $z_t$ with Euler steps.
\end{enumerate}
In Appendix Section B.III, we present additional mathematical formulations of DDOT.

\begin{figure*}[t]
\centering
    \includegraphics[width=\linewidth]{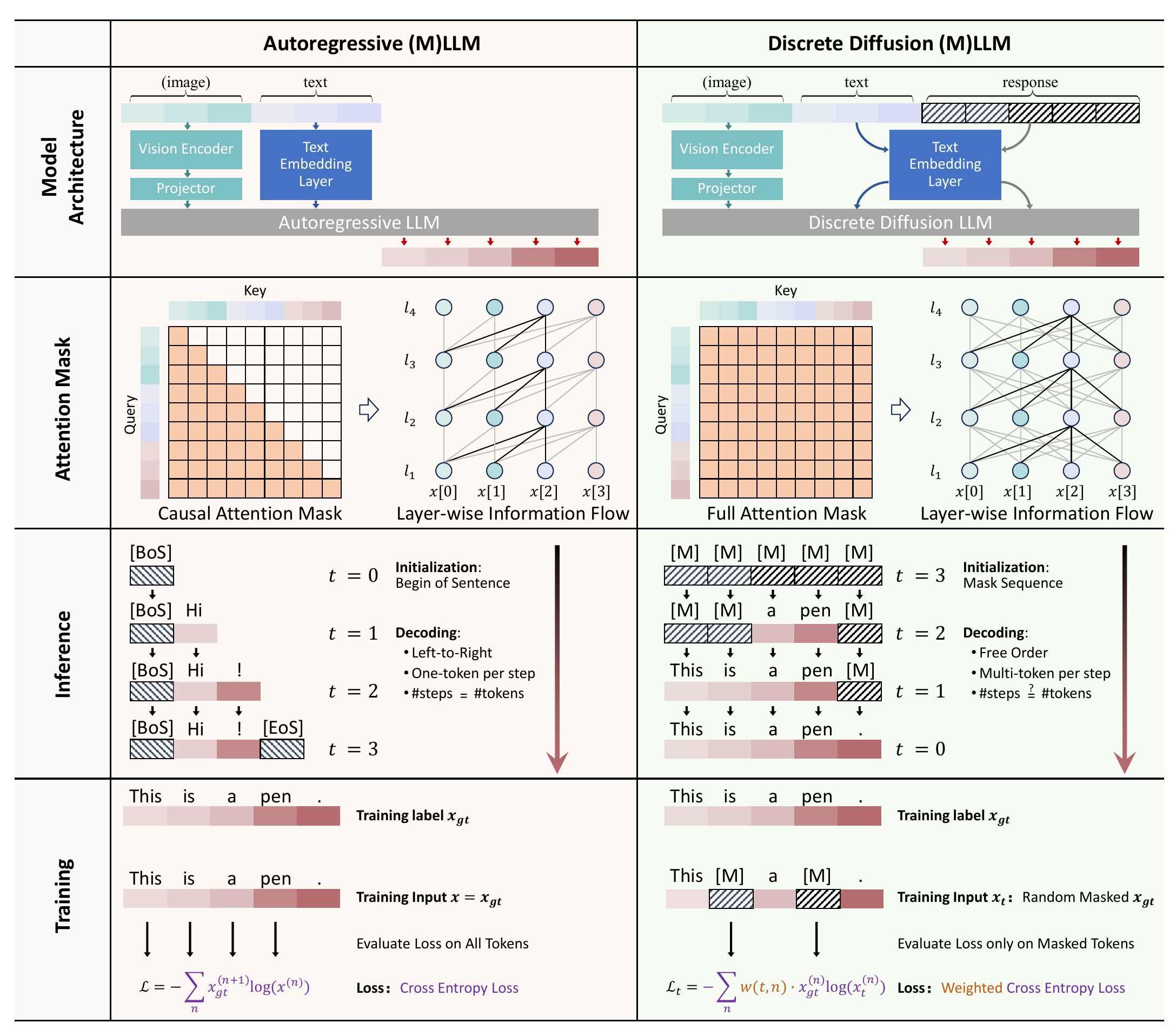}
    \caption{This figure compares autoregressive models and discrete diffusion models from four perspectives. First, regarding model architecture, both models share the same network structure; the key difference lies in their generation mechanisms. In addition to the LLM, both MLLM and dMLLM require an additional vision encoder. In terms of the attention mask, the autoregressive model uses a causal attention mask, whereas the discrete diffusion model adopts a full (bidirectional) attention mask. During inference, the autoregressive model starts from a BoS token and generates tokens one by one from left to right. In contrast, the discrete diffusion model begins with a sequence of mask tokens and denoises all tokens in parallel. At each step, a subset of tokens is selected and replaced with non-mask tokens, continuing until no mask tokens remain. For training, the autoregressive model directly takes the input sequence and applies next-token prediction loss. The discrete diffusion model first randomly masks the input tokens and then computes a weighted cross-entropy loss over the masked positions.
}
\label{fig_compare}
\vspace{-1em}
\end{figure*}

\section{Representative Models}
\label{models}
In this section, we provide a high-level overview of the representative works. In the following sections, we give detailed discussions on the training paradigms and inference-time decoding strategies of the models scaled to sizes comparable to LLMs. An evolutionary diagram of representative dLLM and dMLLM models is shown in \cref{fig_vein}.

\subsection{Discrete Diffusion Models around 1B}
\label{small_models}
\cite{sohl2015deep} first introduces a diffusion process over binary variables. This idea is generalized to categorical variables by \cite{hoogeboom2021argmax}, who demonstrates its effectiveness on image generation. Building on these foundations, D3PM \cite{austin2021structured} proposes a more flexible family of noising schedules that extends discrete diffusion to a broader class of discrete spaces (see \cref{math_1}). DiffusionBERT \cite{he2022diffusionbert} explores training BERT \cite{devlin2019bert} to reverse a discrete diffusion process with an absorbing state, introducing a token-aware noise schedule for the forward pass and methods to embed time-step information into BERT. 

\cite{zheng2023reparameterized} simplifies the formulation of discrete diffusion process in D3PM and introduces a new model family named Reparameterized Discrete Diffusion Models (RDMs). It reformulate the backward process of discrete diffusion in D3PM \cite{austin2021structured} into a two-stage sampling procedure and yield a greatly simplified training objective and enable more flexible decoding algorithms. MLDM \cite{sahoo2024simple} and MD4 \cite{shi2024simplified} further simplified the discrete diffusion model specifically for cases with an absorbing state, also referred to as masked diffusion models.
Masked-Diffuse LM (MDLM) \cite{chen2023cheaper} propose to leverage the inherit linguistic features of texts to encourage the model to recover the text following an easy-first-generation nature, and directly predict the discrete token with cross-entropy loss to stabilize the intermediate diffusion steps. Diffusion-NAT \cite{zhou2023diffusion} uses the pretrained BART \cite{lewis2019bart} as the language backbone and unifies the inference process of pretrained language models and the denoising process of discrete diffusion models in a non-autoregressive manner, thus the BART model plays the role of the parameterized denoiser in discrete diffusion models. Furthermore, TESS \cite{mahabadi2023tess} leverages a new form of self-conditioning and applies diffusion on the logit simplex instead of the learned embedding space. Plaid \cite{gulrajani2023likelihood} takes the first step towards closing the likelihood gap between autoregressive and diffusion-based language models through 
SEDD \cite{lou2023discrete} generalize the idea of score matching into the discrete spaces by proposing a novel loss named score entropy, which can be integrated seamlessly to build discrete diffusion models and can significantly boost model performance.

For unified models, UniDisc~\cite{swerdlow2025unified} is a prior work on unified discrete diffusion model for text and image modalities. Conceptually, UniDisc treats an image and a caption as two token sequences (from discrete codebooks) and denoises them together with a decoder-only Transformer and bidirectional attention.

\subsection{Large Diffusion Language Models}
\label{llm}

\subsubsection{DIFFUSION-LLMs}
DIFFUSION-LLMs~\cite{ye2023diffusion} is the first work we identified that scales discrete diffusion language models to 3B and 10B parameters, showing that their performance improves consistently with model size. Its training involves two stages. First, the model is pretrained with a masked LLM objective (similar to BERT~\cite{devlin2019bert}) to acquire world knowledge. Next, the pretrained masked LLM is reprogrammed into a dLLM using the RDM loss and training strategy on either specific downstream task datasets or instruction-following datasets. Experiments demonstrate that scaled-up discrete diffusion models, like autoregressive models, possess zero-shot generation, in-context learning, and even reasoning capabilities. Moreover, DIFFUSION-LLMs show that dLLMs can outperform autoregressive models on reasoning tasks requiring implicit planning, such as Path-Finding on Path-Star Graphs~\cite{pmlr-v235-bachmann24a}.

\subsubsection{LLaDA Series}
The LLaDA series represents the pioneering line of discrete diffusion-based alternatives to autoregressive LLMs. LLaDA \cite{nie2025large}, the first work in this line of research, is a discrete diffusion large language model. It follows the standard masked diffusion model framework, employing a Transformer with bidirectional attention. Its training objective is the variational likelihood bound (ELBO) \cite{song2020score}, rather than the exact log-likelihood.
While LLaDA demonstrates strong performance after supervised finetuning, aligning a dLLM with human preferences (akin to RLHF~\cite{kaufmann2023survey} for AR models) remains challenging. LLaDA 1.5 \cite{zhu2025llada} specifically addresses this by introducing Variance-Reduced Preference Optimization (VRPO) for dLLMs. 
The main challenge addressed by LLaDA 1.5 is the high variance that arises when estimating log-probabilities with ELBO during reinforcement learning. Based on both theoretical and empirical analysis, LLaDA 1.5 proposes three solutions: increasing the number of Monte Carlo samples, using optimal allocation of samples (many timesteps but one mask per step), and applying antithetic sampling in preference comparisons to reduce variance.

\subsubsection{DiffuGPT \& DiffuLLaMA}
DiffuGPT and DiffuLLaMA \cite{gong2024scaling} propose converting a pretrained autoregressive Transformer (such as GPT-2~\cite{radford2019language} or LLaMA~\cite{touvron2023llama}) into a dLLM, thereby avoiding the cost of training large models entirely from scratch. Crucially, \cite{gong2024scaling} establish theoretical connections between autoregressive next-token prediction and the diffusion denoising objective, enabling alignment between the two paradigms during adaptation. Leveraging the autoregressive model’s knowledge as initialization, the diffusion model can be scaled up with significantly less data (under 200 billion tokens, compared to trillions for training from scratch).
Experiments span models from 127M to 7B parameters, and the resulting series achieves performance comparable to autoregressive LLMs. Notably, due to their bidirectional nature, DiffuGPT and DiffuLLaMA can perform infilling natively without prompt engineering or token reordering, a challenge for autoregressive models. During inference, they also support a trade-off between generation speed and quality by adjusting the number of diffusion iterations, often requiring fewer refinement steps to achieve fluent outputs than other dLLMs.

\subsubsection{DiffuCoder}
DiffuCoder \cite{gong2025diffucoder} is a 7B-parameter dLLM on a large code corpus, the authors reveal how dLLMs dynamically shift between autoregressive-like and parallel generation patterns. To stabilize preference optimization under diffusion training, they propose coupled-GRPO, a reinforcement learning algorithm that reduces variance via complementary mask sampling, yielding measurable improvements in code generation benchmarks. 

\subsubsection{DREAM Series}
DREAM 7B~\cite{dream2025} is one of the most powerful open-source dLLMs to date. DREAM 7B achieves performance on par with, or exceeding, autoregressive models of similar size (e.g. it matches LLaMA3 8B~\cite{grattafiori2024llama} and Qwen 2.5-7B~\cite{qwen2025qwen25technicalreport} on many benchmarks). A key to DREAM’s success is an optimized training recipe distilled from extensive experiments at smaller scales. \cite{dream2025} carefully explores the design choices on a 1B model and identify two especially valuable components: (1) AR weight initialization as in \cite{gong2024scaling} and (2) context adaptive noise scheduling as in ~\cite{ye2025beyond}. The AR initialization of DREAM is chosen to be Qwen2.5~\cite{qwen2025qwen25technicalreport}.

The Dream series has several follow-up works. Dream-Coder 7B \cite{dreamcoder2025} introduces a diffusion-based code model, adapted from Qwen2.5-Coder~\cite{hui2024qwen25codertechnicalreport} and trained on 322B tokens. It supports multiple decoding styles, including sketch-first generation, left-to-right decoding, and interleaved reasoning. DreamOn \cite{Dreamon2025} resolves the fixed-canvas limitation of diffusion decoding by introducing special tokens ($<\!\!|\text{expand}|\!\!>$, $<\!\!|\text{delete}|\!\!>$) for variable-length generation. During inference, a masked token can be predicted as one of these operations: the $<\!\!|\text{expand}|\!\!>$ token splits the current mask into two masks, while the $<\!\!|\text{delete}|\!\!>$ token removes the current mask. This innovation enables true flexible infilling and significantly improves performance on benchmarks such as HumanEval-Infilling.

\subsubsection{TESS 2}
TESS 2~\cite{tae2025tess} is another dLLM that is not only large-scale but also instruction-following and general-purpose. The training recipe for TESS 2 is a culmination of ideas from prior works, such as AR initialization~\cite{gong2024scaling} and reward guidance~\cite{mahabadi2023tess}. TESS 2 starts by adapting a powerful AR base model via continued pretraining on the diffusion objective, and then applies thorough instruction-tuning to that adapted model. TESS 2 finds that both the adaptation procedure and the choice of base model are crucial for a good dLLM. For reward guidance, TESS 2 shows that the choice of reward model exhibits some robustness.

\subsubsection{Seed Diffusion}
Seed Diffusion (Preview) \cite{song2025seed}  is a powerful discrete diffusion model primarily designed for code generation, achieving an inference speed of 2,146 tokens per second on H20 GPUs. Seed Diffusion employs two types of data noising: 80\% masked language modeling and 20\% random deletion, insertion, or substitution, with edit distance used to measure differences between sequences. The training loss combines a mask prediction term and an overall reconstruction term. To enable better sequential generation, Seed Diffusion leverages a pretrained diffusion model to generate partial trajectory data, which is then filtered by overall log-likelihood to construct a refined trajectory dataset. To further reduce decoding steps, it introduces a reinforcement learning paradigm that maximizes the edit distance between consecutive steps, encouraging the model to decode as many tokens as possible per step while maintaining correctness.

\subsection{Large Diffusion Multimodal Models}
\label{mllm}
\subsubsection{Dimple}
Dimple \cite{yu2025dimple} is one of the first Discrete Diffusion Multimodal Large Language Models (dMLLMs). Its base architecture (vision encoder + transformer LLM) resembles existing vision-language models (e.g. Qwen-VL \cite{Qwen-VL}, LLaVA \cite{liu2023llava,liu2023improvedllava,liu2024llavanext}). One of the Dimple’s key innovations is its two-stage hybrid training. In Stage 1, with the weights of Dream 7B \cite{dream2025} as an initialization, it is fine-tuned autoregressively on vision-language instruction data (for alignment and instruction following). In Stage 2, it is then further fine-tuned with a discrete diffusion objective. This hybrid approach is devised because pure diffusion training is found to be unstable (leading to length bias and performance drop). By warming up with autoregressive training first, Dimple-7B achieves stable training and eventually surpasses the fully-autoregressive models.

During inference, Dimple introduces a \emph{confident decoding} strategy for efficiency: the model dynamically chooses how many tokens to fill in at each step based on model confidence (see \cref{unmask}). Empirically, this reduces the number of iterations to about $\frac{\text{response length}}{3}$. Dimple also re-implements an autoregressive \emph{prefilling} trick: by filling in some tokens from the existing context, it speeds up inference by about 1.5$\times$–7$\times$ with minimal impact on quality. Under the same training budget and dataset as LLaVA-NEXT \cite{liu2024llavanext}, Dimple-7B achieves higher aggregate scores on multimodal benchmarks than LLaVA-NEXT-7B. This result shows that with a proper hybrid training recipe, a discrete dMLLM can match or exceed strong autoregressive baselines on vision-language tasks.

\subsubsection{LaViDa}
LaViDa \cite{li2025lavida} is among the first models to extend discrete diffusion into the multimodal large language model (dMLLM) setting. It consists of a vision encoder (e.g. SigLIP-400M \cite{zhai2023sigmoid}) and a discrete diffusion Transformer. Its language model is a standard discrete dLLM (either LLaDA-8B or Dream-7B). 
LaViDa's key innovation is complementary masking in training: for each training sample, two distinct mask patterns are created so that each token is masked in one of the two versions. This ensures that even short or rare answer tokens (e.g. object names in vision tasks) contribute to the loss and all tokens are learned efficiently, improving alignment between the visual encoder and the language model. During inference, LaViDa employs a special Prefix-DLM \cite{li2025lavida} attention mask so that the encoded image and prompt tokens can be cached and reused. The model also uses a timestep-shifting schedule to improve sample quality. 

\subsubsection{LLaDA-V}
LLaDA-V \cite{you2025llada} is one of the pioneering efforts in developing Discrete Diffusion Multimodal Large Language Models (dMLLMs). Built on LLaDA \cite{nie2025large}, LLaDA-V undergoes three training phases. In the first stage, language–image alignment, the MLP projector is trained to align visual features with LLaDA’s word embeddings. The second stage, visual instruction tuning, fine-tunes the model on large-scale multimodal instruction data to build instruction-following abilities. Finally, the third stage, multimodal reasoning enhancement, focuses on strengthening reasoning capabilities by training on reasoning-focused multimodal QA data with detailed reasoning chains, and further balancing direct answering and explicit reasoning through mixed training with ``no\_think'' and ``think'' tags.

\subsection{Large Unified Model}
\label{unified_models}

\subsubsection{MMaDA}
MMaDA~\cite{yang2025mmada} employs a unified diffusion architecture with a shared probabilistic formulation across image and text modalities. It uses a single diffusion-based transformer for all data types (text, images, etc.), rather than separate encoders for each modality. 
During training, MMaDA is fine-tuned with a \emph{mixed long chain-of-thought} strategy. Reasoning steps from both text and vision tasks are converted into a unified CoT format so that the model learns aligned reasoning across modalities. For example, the rationale for answering a visual question is interleaved into the textual input. This CoT alignment provides a form of cold-start for the final reinforcement learning (RL) stage, allowing complex multi-step reasoning from the outset. Finally, MMaDA proposes a unified policy-gradient-based RL algorithm named \emph{UniGRPO}. By incorporating diversified reward modeling, UniGRPO unifies the post-training across both reasoning and generation tasks, improving the performance.

\subsubsection{FUDOKI}
FUDOKI~\cite{wang2025fudoki} is a unified multimodal model built on discrete flow matching~\cite{shaul2024flow}. It uses a metric-induced probability path with kinetic-optimal velocities~\cite{shaul2024flow}, which significantly improves over simple masking by enabling continuous self-correction during generation. For efficiency, FUDOKI is initialized from a pretrained AR-based multimodal LLM (Janus-1.5B~\cite{wu2025janus}) and then transferred to the discrete flow matching framework. For input modalities, text is tokenized normally, while images are handled by separate pipelines: a semantic encoder (SigLIP~\cite{zhai2023sigmoid}) extracts features for image understanding, and a pixel encoder/decoder \cite{sun2024autoregressive} converts images to/from discrete image tokens for generation. At the output stage, FUDOKI has two output heads---one predicting text tokens and one predicting image tokens---and selects the appropriate head depending on the target modality during inference.

\subsubsection{Muddit}
Muddit~\cite{shi2025muddit} is another unified model that uses purely discrete diffusion to handle text and images under one framework. 
The architecture of Muddit comprises a single multimodal diffusion transformer (MM-DiT)~\cite{shi2025muddit}, plus encoders/decoders for each modality. The MM-DiT follows a dual-/single-stream design (similar to FLUX~\cite{flux2024}) and is initialized from the pretrained Meissonic~\cite{bai2024meissonic}. Inputs are quantized into a shared token space: images are encoded by a pretrained VQ-VAE \cite{van2017neural} into discrete codebook indices, and text is encoded by a CLIP text encoder~\cite{radford2021learning}. During training and inference, the MM-DiT predicts masked tokens in this joint space. A linear head maps the predicted tokens to actual text tokens for text output, while the VQ-VAE decoder reconstructs pixels from image tokens. 
\section{Training Techniques}
\label{training}
In this section, we summarize the techniques employed during the training of diffusion language models (dLLMs) and diffusion multimodal language models (dMLLMs).

\subsection{Challenges}
First, we summarize several challenges encountered in the training of discrete diffusion models.
These challenges stem from low corpus utilization and high variance due to stochastic masking.

\subsubsection{Low Corpus Utilization} 
Unlike autoregressive training, where each token in the answer sequence contributes to the learning signal, discrete diffusion training applies supervision only to a randomly selected subset of tokens at each time step. Given an input sequence $x$ of length $L = L_{\text{prompt}} + L_{\text{answer}}$, diffusion training samples a timestep $t \in [1, T]$ and computes loss only over the masked tokens at that timestep. This leads to sparse supervision across training samples, resulting in inefficient utilization of the corpus.

\subsubsection{Random Sampling of Time Index}

In diffusion training, the time index $t$ is randomly sampled for each training instance. As a result, only a single generation step is supervised per sample, while the decoding process at inference time typically involves multiple time steps. This mismatch introduces a coverage gap between training and inference: although decoding requires  refinement over many steps, training provides gradient signals for only one of those steps per instance. 

\subsection{Initialization Techniques}
\label{initialization}
To address the inefficiencies and instabilities in training dLLMs and dMLLMs, several works adopt advanced initialization strategies that convert the full diffusion training process into a fine-tuning task. This approach accelerates convergence and enhances final model performance.

Because the diffusion generation process can be interpreted as a multi-step masked language modeling (MLM) procedure. \cite{he2022diffusionbert} initializes diffusion models from pretrained BERT.
Moreover, \cite{gong2024scaling,dream2025} explored direct adaptation from autoregressive language models by aligning the training objectives of the two paradigms. A key technique enabling this transition is the \emph{shift operation}.
In standard diffusion training, the model predicts the original token $x_0$ from its corrupted version $x_t$ at each timestep. However, this formulation differs from AR training, where each hidden state $h_{i}$ is trained to predict the next token $x_{i+1}$ in a left-to-right fashion. To bridge this gap, \cite{gong2024scaling,dream2025} propose shifting the output logits of the diffusion model by one position, such that the model's prediction at position $i$ corresponds to token $x_{i+1}$.
This allows the diffusion model to be initialized with pretrained autoregressive models.

Another approach similar to initialization is Autoregressive-then-Diffusion Training, making the Diffusion Training as posting-training of autoregressive training.
Dimple~\cite{yu2025dimple} uses an \emph{autoregressive-then-diffusion} training approach, demonstrating notable performance improvements for DMLLM.
The Dimple training pipeline is divided into two distinct phases:
\begin{itemize}
    \item \textbf{Phase I: Autoregressive Training.} In the first phase, Dimple is treated as an autoregressive model using the causal attention mask and next-token prediction loss.
    \item \textbf{Phase II: Diffusion Fine-tuning.} After autoregressive training, Dimple is treated as an diffusion model using the full attention masks and timestep-dependent masked language modeling losses.
\end{itemize}

\subsection{Complementary Masking  Technique}
\label{comp_masking}
To improve the utilization of the corpus, in \cite{li2025lavida}, to ensure that all tokens participate in training, complementary masking constructs two complementary masked versions of each input sequence: $X_t$ and $X_t^C$. $X_t$ and $X_t^C$ have non-overlapping masked spans. For example, consider the sentence:
\begin{center}
\texttt{``The answer is dog.''}
\end{center}
One masked variant might be:
\begin{center}
\texttt{``The [M] [M] dog.''}
\end{center}
and its complement:
\begin{center}
\texttt{``[M] answer is [M].''}
\end{center}
This setup ensures that all tokens are eventually masked and optimized over the course of training. 

\subsection{Masking Scheduling Technique}
\label{masking_scheduling}
\emph{Masking scheduling} governs the corruption process in the forward diffusion formulation. Specifically, the schedule defines the corruption level $\alpha_t$ at each timestep $t$, thereby determining the proportion of tokens masked during training. An effective schedule balances learning stability and generation quality by controlling the signal-to-noise ratio across timesteps.

\subsubsection{Uniform Masking Scheduling}
Masking scheduling can either apply the same scheduling function to all tokens, referred to as uniform masking scheduling, or assign different scheduling functions to individual tokens, referred to as token-wise masking scheduling. We first introduce two commonly used  uniform masking scheduling methods. 

\begin{figure}[t]
    \centering
    \subfloat[$\alpha_t$ Schedules\label{fig:alpha}]{
        \includegraphics[height=105pt]{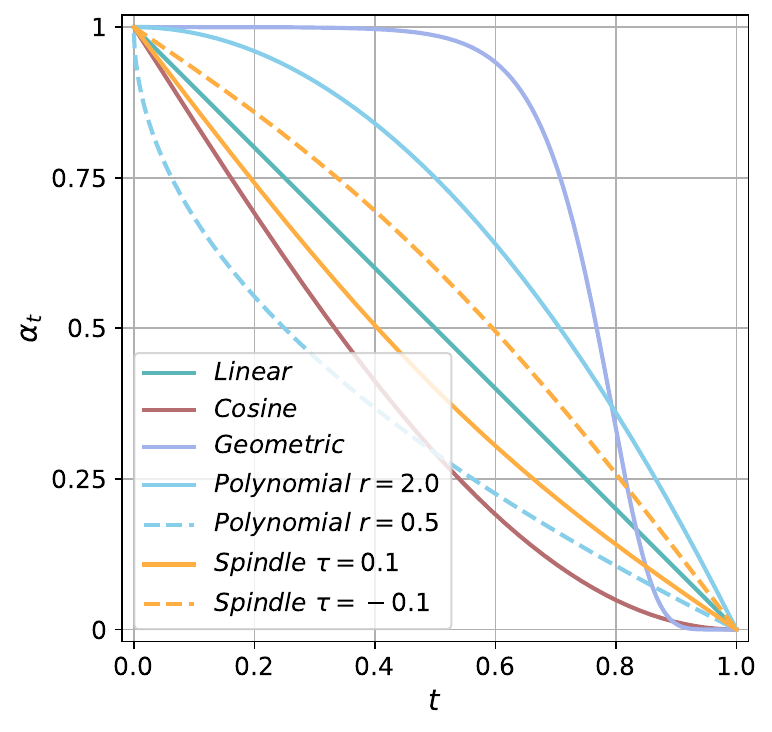}
    }
    \hfill
    \subfloat[$w_t$ Schedules\label{fig:w}]{
        \includegraphics[height=105pt]{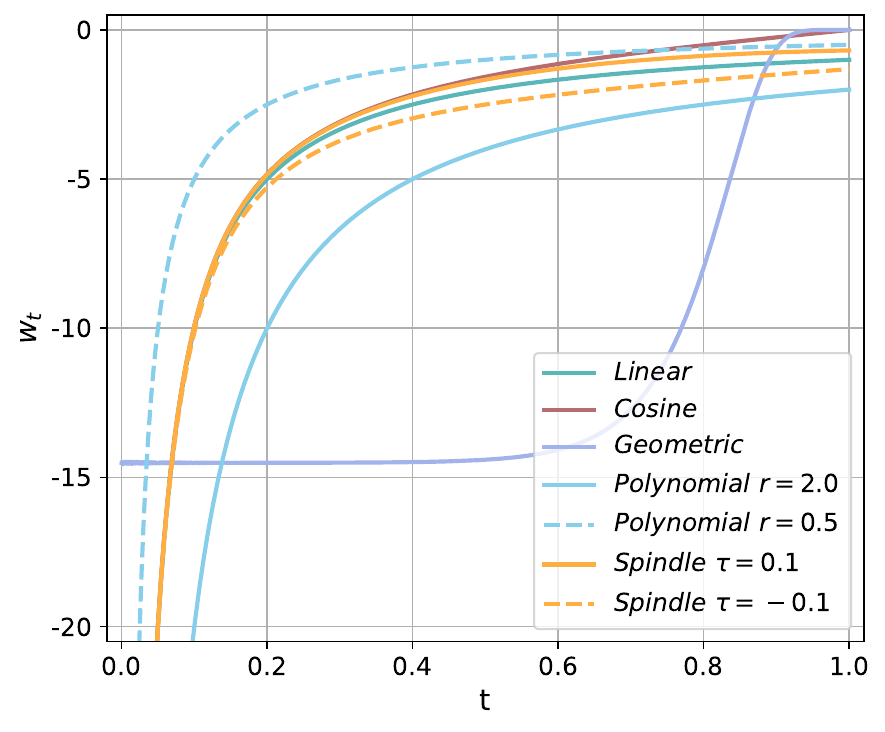}
    }
    \caption{Different schedules for $\alpha_t$ and $w_t$. To unify notation, we also transformed the spindle schedule from the discrete-time format used in the original paper to a continuous-time format. The revised formulation is as follows: $\alpha_t = 1-t-\tau\sin(\pi t)$ and $w_t=-\frac{1+\tau\pi \cos(\pi t)}{t + \tau\sin(\pi t)}$, where $\tau$ corresponds to the original $\lambda \tilde{H}$.
}
    \label{fig:schedules}
\vspace{-1em}
\end{figure}

Given a timestep $t \in [0, 1]$, the forward process defines the corruption as:
\begin{equation}
q(x_t \mid x_0) = \alpha_t x_0 + (1 - \alpha_t) m,
\end{equation}
where $m$ is the one-hot [MASK] token. The loss at each step is reweighted according to:
\begin{equation}
w_t = \frac{\alpha_t'}{1 - \alpha_t},
\end{equation}
where $\alpha_t' = \frac{d\alpha_t}{dt}$ is the derivative of $\alpha_t$ with respect to time.
Linear Schedule and Cosine Schedule are two commonly-used scheduling strategies are as follows. Their corresponding schedule functions are plotted in \cref{fig:schedules}.

\begin{itemize}
    \item \textbf{Linear Schedule~\cite{austin2021structured}:}
    \begin{align}
        \alpha_t = 1 - t, \quad
        w_t = -\frac{1}{t}.
    \end{align}
    \item \textbf{Cosine Schedule~\cite{chang2022maskgit}:}
    \begin{align}
        \alpha_t = 1 - \cos\left( \frac{\pi}{2}(1 - t) \right),  \quad
        w_t = -\frac{\pi}{2} \tan\left( \frac{\pi}{2}(1 - t) \right).
    \end{align}
\end{itemize}
The theoretical analyses of the optimal masking scheduling function remain scarce. One pioneering work \cite{zhang2025cosineschedulefisherraooptimalmasked} theoretically proves the optimality of cosine scheduling under the Fisher–Rao geometry. In Appendix Sec.C.I, we include another two uniform scheduling functions.

\subsubsection{Token-wise Masking Scheduling}
Uniform masking scheduling uses the same scheduling for all tokens, which ignores the inherent variation in informativeness among different tokens. 
\cite{he2022diffusionbert} introduces a \emph{token-wise masking schedule}, which defines a \emph{Spindle-shaped Noise Schedule}, where the entropy-aware corruption pattern resembles a spindle curve: more informative tokens are masked earlier and less informative ones later. This design ensures that low-entropy, easier-to-predict tokens are decoded first, leading to more coherent and stable sequence generation. The math of the Spindle-shaped Noise Schedule is discussed in Appendix Sec.C.I.

\subsubsection{Block-wise Masking Scheduling}
For reasoning tasks, the prevailing approach adopts a blockwise semi-autoregressive decoding strategy. However, previous random masking across the full response, does not reflect this procedure.
Let the response be partitioned into $M$ contiguous blocks $b^{(1)}, \ldots, b^{(M)}$ of size $B$. For a given active block $a$, Blockwise SFT~\cite{sun2025blockwisesftdiffusionlanguage} defines:
\begin{itemize}
    \item \emph{Prefix} $I^{(a)}_{\text{prefix}}$: tokens before block $a$, kept clean and fixed.
    \item \emph{Active block} $I^{(a)}$: tokens of block $a$, subject to stochastic masking.
    \item \emph{Suffix} $I^{(a)}_{\text{suffix}}$: tokens after block $a$, fully hidden.
\end{itemize}
In other words, the masking rule is:
\begin{equation*}
m_i = 
\begin{cases}
0, & i \in I^{(a)}_{\text{prefix}} \quad \text{(clean prefix)}, \\
\text{Bernoulli}(\pi), & i \in I^{(a)} \quad \text{(masked active block)}, \\
1, & i \in I^{(a)}_{\text{suffix}} \quad \text{(fully hidden suffix)},
\end{cases}
\end{equation*}
with $\pi \sim \text{Uniform}(10^{-3}, 1)$. Loss and gradient updates are computed only for the active block.

\subsection{Reweighting Technique}
\label{training-rw}
Multi-Granularity Diffusion Modeling (MGDM)~\cite{ye2025beyond} introduces an additional token-level reweighting factor \( v(x_{t,n}) \) in the loss function, yielding:
\begin{equation}
\mathcal{L}_{\text{MGDM}} = \sum_{n=1}^{N} \sum_{t=1}^{T} w(t) \cdot v(x_{t,n}) \cdot \ell(x_0, x_t, n; \theta),
\end{equation}
where $\ell(x_0, x_t, n; \theta)$ is the CE loss on the n-th token, and the adaptive token-level weight is defined as:
\begin{equation}
v(x_{t,n}) = \alpha (1 - \exp(-\ell(x_0, x_t, n; \theta)))^{\beta},
\end{equation}
with hyperparameters \( \alpha > 0 \), \( \beta > 0 \). This reweighting assigns larger weights to harder tokens (i.e., those with higher loss), thereby prioritizing difficult subgoals during training and accelerating convergence.

\subsection{Distillation through Dimensional Correlations}
\label{distillation}
To enable efficient few-step or even one-step generation while maintaining performance, Di4C \cite{hayakawa2024distillation} introduces a distillation strategy that compresses a multi-step dLLM into a fewer-step counterpart. It employs two loss functions: *Distillation Loss* and *Consistency Loss*, grounded in distributional matching and multi-path coherence. The distillation loss transfers probabilistic knowledge from a teacher model performing multi-step denoising to a student model designed for fewer-step generation, while the consistency loss enforces stable behavior of the student model across predictions from different intermediate noise levels. The detailed mathematical formulations of these two losses are provided in Appendix Sec. C.II.

\subsection{Reinforcement Learning}
\label{rl}
Reinforcement learning has been extensively applied to discrete diffusion language models~\cite{yang2025mmada,zhu2025llada,huang2025reinforcingdiffusionchainlateral,tang2025wd1weightedpolicyoptimization,han2025discretediffusiontrajectoryalignment,he2025mdpoovercomingtraininginferencedivide,wang2025timefeatureexploitingtemporal}. Reinforcement learning for diffusion language models largely inherits the paradigm from autoregressive models; however, it also presents several unique challenges, such as the estimation of likelihood. The following provides a brief summary of several reinforcement learning techniques, with detailed descriptions presented in Appendix Sec.~C.IV.

\emph{Diffusion-based GRPO} (UniGRPO) extends clipped policy optimization by integrating structured noising and KL-regularized surrogate rewards \cite{yang2025mmada}. \emph{Variance-Reduced Preference Optimization} (VRPO) improves stability by replacing intractable log-likelihoods in DPO with ELBO estimates and applying advanced variance reduction techniques \cite{zhu2025llada}. To address reward propagation across trajectories, \emph{Stepwise Decomposition Preference Optimization} (SDPO) reformulates alignment into tractable per-step KL-regularized objectives \cite{han2025discretediffusiontrajectoryalignment}. \emph{Weighted Policy Optimization} (wd1) recasts the objective as a weighted likelihood maximization, where weights derived from centered rewards ensure better sample efficiency \cite{tang2025wd1weightedpolicyoptimization}. Finally, \emph{Diffusion Chain of Lateral Thought} (DCoLT) introduces a reinforcement-learned Unmask Policy Module that adaptively controls the token unmasking order during generation \cite{huang2025reinforcingdiffusionchainlateral}. Together, these methods highlight complementary directions for improving stability, efficiency, and controllability in diffusion-based reinforcement learning.

\section{Inference Techniques}
\label{inference}
In this section, we summarize the techniques involved in the inference phase of dLLMs and dMLLMs. These techniques affect both performance and efficiency, typically requiring a trade-off between the two. Ideally, the goal is to improve decoding efficiency without compromising performance.

\subsection{Unmasking Techniques}
\label{unmask}
\begin{figure*}[t!]
    \centering
\subfloat[Metric-Based Unmasking. \label{fig:metric_based}]{
  \includegraphics[width=0.30\textwidth]{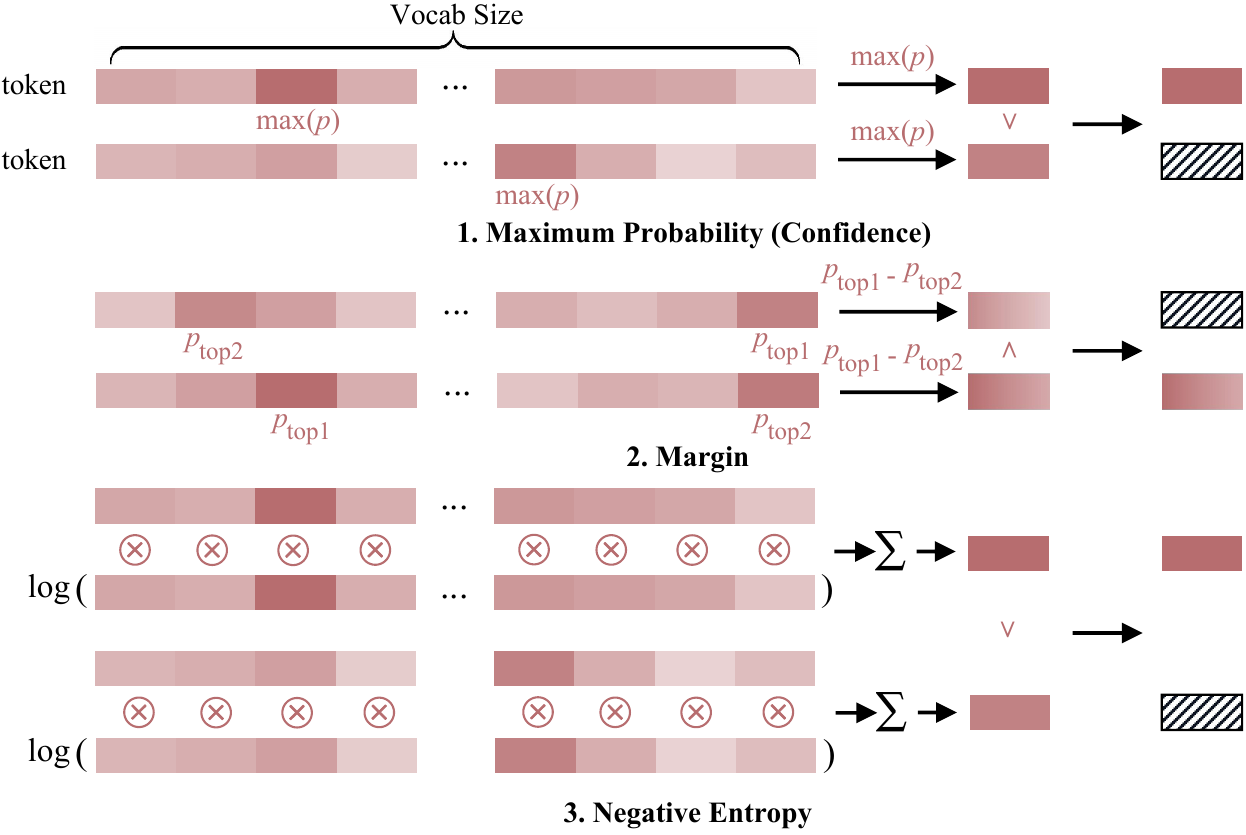}
} 
\vspace{2mm}
\subfloat[Confident Decoding. \label{fig:conf_3}]{
  \includegraphics[width=0.64\textwidth]{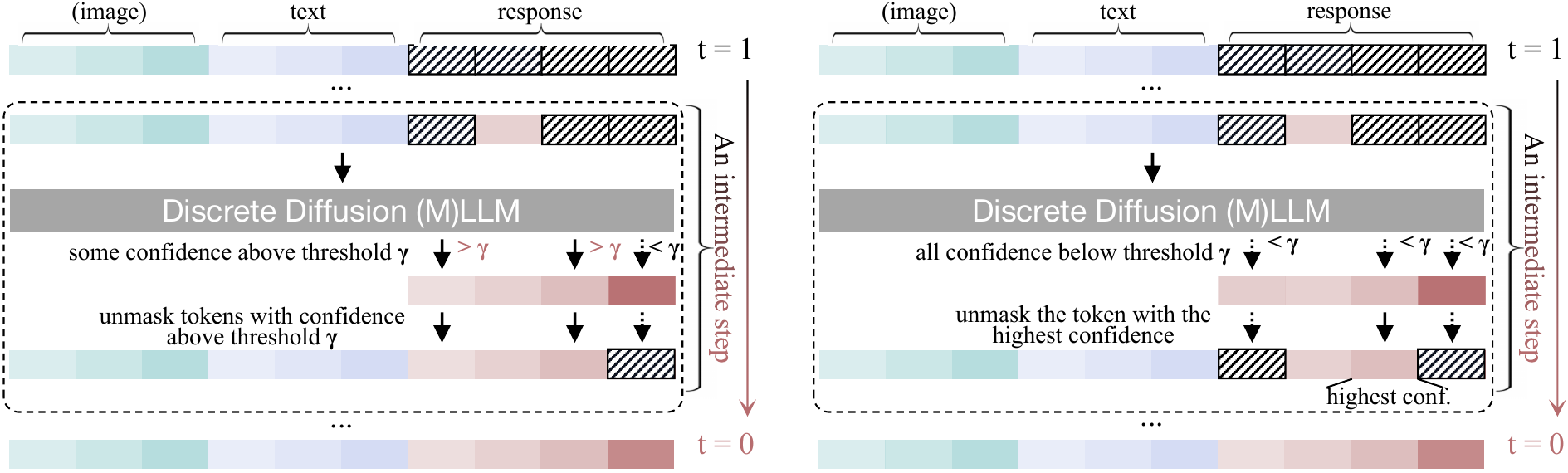}
} 
\vspace{2mm}
\subfloat[Top-$s_t$ Strategy. \label{fig:conf_2}]{
  \includegraphics[width=0.31\textwidth]{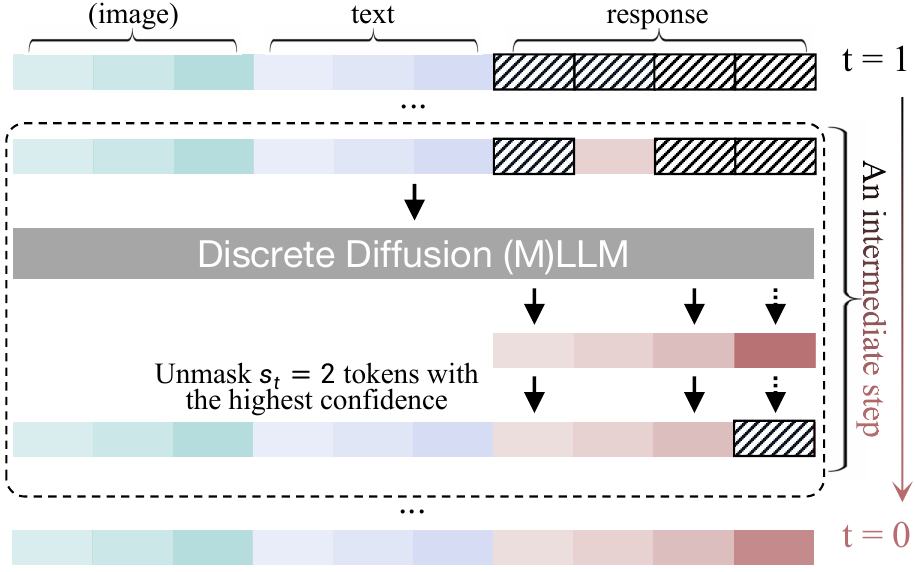}
}
\vspace{2mm}
\subfloat[Local Unmasking. \label{fig:conf_4}]{
  \includegraphics[width=0.31\textwidth]{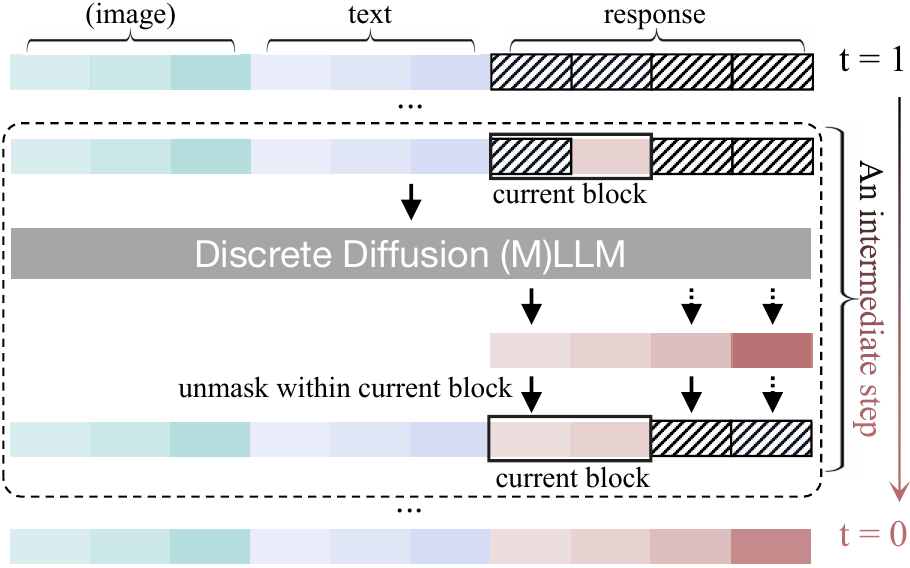}
}
\vspace{2mm}
\subfloat[Continuous-Time Unmasking. \label{fig:time_2}]{
  \includegraphics[width=0.31\textwidth]{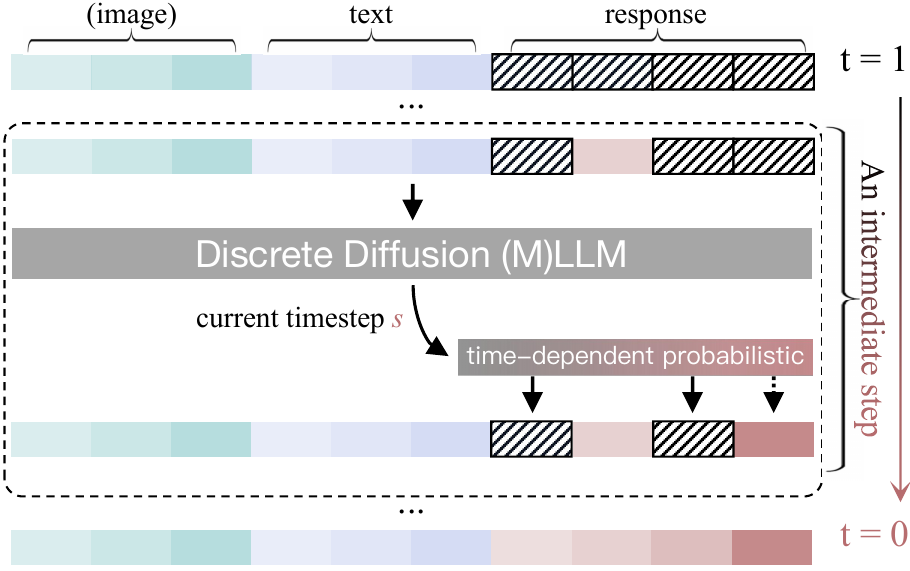}
} 
\caption{\textbf{Unmasking strategies.} We divide the unmasking strategies used in dLLMs and dMLLMs into two categories: Discrete-Time Unmasking (a,b,c,d) and Continuous-Time Unmasking (e). In discrete-time unmasking, besides random unmasking, there are other two unmasking strategies: Metric-Based Unmasking (Maximum Probability (Confidence), Margin and Negative Entropy, see (a)) and Selection Policies (Top-$s_t$ Strategy (see (c)), Confident Decoding (see (b)), and Local Unmasking (see (d))).}
\label{fig:unmask_all}
\vspace{-1em}
\end{figure*}

In dLLMs and dMLLMs, the model predicts all the response tokens at each step. However, only a subset of the masked tokens are selected to be unmasked at every iteration, while the remainder remain masked. The main challenges are deciding which and how many tokens to unmask per iteration.
Figure \ref{fig:unmask_all} provides a detailed illustration of the unmasking strategies. In this section, we discuss each category in detail and describe the specific unmasking strategies proposed in each work.

\subsubsection{Discrete-Time Unmasking}

\subsubsubsection{Random Unmasking}

The simplest strategy is to randomly select $s_t$ masked tokens to unmask at step $t$. The value of $s_t$ can be fixed across steps or controlled by a scheduling function as discussed in the training techniques, such as cosine scheduling~\cite{you2025llada}. 

\subsubsubsection{Metric-Based Unmasking}

Rather than relying on random selection, metric-based strategies assign a metric value to each token prediction and select tokens to be unmasked based on the metric value.

Let $p^{(i)} \in \mathbb{R}^K$ be the predicted probability distribution over the vocabulary for the $i$-th token, where $K$ is the vocabulary size. The following are some commonly used metrics.

\begin{itemize}
    \item \textbf{Maximum Probability (Confidence)}~\cite{dream2025,kim2025train}:
    \begin{equation}
    c^{(i)} = \max(p^{(i)}),
    \end{equation}
    indicating the model’s certainty about the most likely token. \cite{wu2025fastdllm} provides a theoretical analysis of the equivalence between the parallel decoding using confidence and sequential decoding using confidence.

    \item \textbf{Margin}~\cite{dream2025,kim2025train}:
    \begin{equation}
    c^{(i)} = p_{\text{top1}}^{(i)} - p_{\text{top2}}^{(i)},
    \end{equation}
    where $p_{\text{top1}}^{(i)}$ and $p_{\text{top2}}^{(i)}$ are the first and second highest probabilities, respectively. This measures how dominant the top prediction is.

    \item \textbf{Negative Entropy}~\cite{dream2025,kim2025train}:
    \begin{equation}
    c^{(i)} = -\sum_{k=1}^{K} p_k^{(i)} \log(p_i + \epsilon),
    \end{equation}
    with a small constant $\epsilon$ for numerical stability. This captures the peakedness of the distribution.
    
    \item In \textbf{EB-Sampler}~\cite{benhamu2025acceleratedsamplingmaskeddiffusion}, instead of performing filtering at the token level, the metric is defined over a set of tokens, aiming to directly decide on unmasking a subset $U$ of tokens.
    \begin{equation}
    c^{(U)} = \sum_{l \in U} H (p^{(l)})-  \max_{l \in U} H (p^{(l)})
    \end{equation}
    with $H (p^{(l)})$ is the entropy of $p^{(l)}$.
    
    \item In \textbf{PC-Sampler.}~\cite{huang2025pcsamplerpositionawarecalibrationdecoding}, the metric is defined as 
    \begin{equation}
    c^{(i)} = w^{(i)}\cdot p^{(i)}_{x^{(i)}}\cdot [-\log p_{\mathcal{D}}(x^{(i)})], \quad w^{(i)}=e^{-\lambda \cdot i},
    \end{equation}
    where $x^{(i)}$ is the predicted token at position $i$. In this metric, $w^{(i)}$ controls the degree of left-to-right decoding by applying an exponentially decaying weight based on token positions. The parameter $\lambda$ serves as a hyperparameter. $p^{(i)}_{x^{(i)}}$ denotes the model’s predicted probability for the current token, reflecting its confidence. $p_{\mathcal{D}}$ represents the token frequency distribution estimated from a publicly available corpus $p_{\mathcal{D}}$, used to downweight trivial words such as ``the'' and ``a''.
\end{itemize}

\subsubsubsection{Selection Policies}

After obtaining the metric value for each token, the diffusion model performs selection based on different policies.
\begin{itemize}
    \item \textbf{Top-$s_t$ Strategy~\cite{austin2021structured}:} Select the $s_t$ tokens with the highest confidence scores for unmasking. The value of $s_t$ follows the same scheduling principles as in random unmasking.

    \item \textbf{Confident Decoding:} As introduced in Dimple~\cite{yu2025dimple}, this strategy dynamically selects the number of tokens to unmask based on a fixed confidence threshold $\gamma \in (0, 1)$. The motivation of this approach is that decoding should adapt to the semantic structure of  the text: some steps may allow many tokens to be confidently predicted, while others may necessitate more caution. Thus, the number of decoded token should be adaptively adjusted at each step.
    At step $t$, the set of positions to decode is defined as:
    \begin{equation}
    \mathcal{I}_t = \{ i \mid c_t^{(i)} \geq \gamma \},
    \end{equation}
    where $c_t^{(i)}$ is the confidence score at position $i$. If $\mathcal{I}_t$ is non-empty, all tokens in $\mathcal{I}_t$ are unmasked. Otherwise, only the token with the highest confidence is selected. This approach enables:
    \begin{itemize}
        \item decoding multiple tokens in parallel when the model is confident, improving efficiency;
        \item avoiding low-confidence predictions, preserving generation quality.
    \end{itemize}

    \item \textbf{Block-wise Unmasking:} Block-wise semi-autoregressive decoding strategy~\cite{nie2025large} divides the full response into multiple blocks, similar to block diffusion. During each forward pass, predictions are generated for all blocks simultaneously. However, the unmasking of tokens follows a left-to-right, block-by-block order. Tokens in the next block are only allowed to be unmasked once all tokens in the previous block have been unmasked.
    \item \textbf{Dilated Unmasking Schedule (DUS)}~\cite{luxembourg2025planspeeddilatedscheduling}: DUS considers the relative distances between tokens during decoding and requires the  distances among decoded tokens to decrease gradually from large to small. Specifically, let $\mathcal{U}_t$ denote the selected tokens at time $t$, DUS selects equally spaced tokens as following:
    \begin{align}
    \mathcal{U}_0 &= \varnothing,  \\
    \mathcal{P}_t &= \{k \mid (k-1) \bmod s_t = 0 \},  \\
    \mathcal{U}_t &= \mathcal{U}_{t-1} \cup \mathcal{P}_t,
    \end{align}
    where $s_t$ is a dilation coefficient that decreases over time.
\end{itemize}

\subsection{Remasking Techniques}
\label{remask}

For masked discrete diffusion model, once a token is unmasked, it remains unchanged in subsequent steps. This static behavior limits the model’s capacity to revise or refine earlier predictions. To address this, the remasking technique reintroduces masked tokens at previously unmasked positions, enabling iterative refinement of generated outputs. 

\subsubsection{Remasking in General Masked Diffusion Models.}
\cite{wang2025remasking} formulates the reversal diffusion process with remasking as 
\begin{align}
&q_\sigma(\mathbf{z}_s \mid \mathbf{z}_t, \mathbf{x}) =\nonumber\\
&\ 
\begin{cases}
\mathrm{Cat}(\mathbf{z}_s; (1 - \sigma_t)\mathbf{x} + \sigma_t \mathbf{m}), & \mathbf{z}_t \ne \mathbf{m}, \\
\mathrm{Cat}\left(\mathbf{z}_s; \frac{\alpha_s - (1 - \sigma_t)\alpha_t}{1 - \alpha_t} \mathbf{x} + \frac{1 - \alpha_s - \sigma_t \alpha_t}{1 - \alpha_t} \mathbf{m} \right), & \mathbf{z}_t = \mathbf{m},
\end{cases}
\end{align}
where $\sigma_t$ is used to control the ratio of remasked tokens. 
When $\mathbf{z}_t \ne \mathbf{m}$, the token has already been decoded. The model samples the next token $\mathbf{z}_s$ from a distribution that mixes the input $\mathbf{x}$ and the mask token $\mathbf{m}$, controlled by $\sigma_t$. This enables remasking by reintroducing uncertainty into already decoded tokens.
When $\mathbf{z}_t = \mathbf{m}$, the token is still masked. The sampling distribution is a weighted combination of $\mathbf{x}$ and $\mathbf{m}$, adjusted by both $\alpha_t$ and $\sigma_t$, allowing flexible control over how much information from the input or the mask dominates.

\subsubsection{Wide-In Narrow-Out}
Under the block-wise decoding setting, \cite{hong2025wideinnarrowoutrevokabledecoding} introduces an alternative verification-based remasking approach. Specifically, A appends an additional set of shadow tokens after the current response, with the number of shadow tokens equal to the current block size and each shadow token corresponding to one token in the block. By adjusting the attention mask, the original token sequence is prevented from attending to the shadow tokens, while each shadow token can attend to all tokens except its corresponding token in the current block. Consequently, the shadow tokens verify the decoded tokens without interfering with the decoding process of the original response sequence. If some decoded tokens are found to have low confidence, they are replaced with mask tokens.

\subsection{Prefilling and Caching Techniques}
\label{prefilling}
Prefilling and Key-Value Cache (KV-Cache) are standard inference acceleration techniques widely adopted in autoregressive language models. Intuitively, Prefilling and KV-Cache avoids redundant computation by storing the key and value representations from previous decoding steps, enabling the model to reuse them instead of recalculating at each new time step. In autoregressive models, the use of causal attention masks ensures that caching is theoretically lossless. In contrast, dLLMs and dMLLMs employ full (bidirectional) attention mechanisms, wherein each token can attend to all other positions. As a result, even tokens that have already been decoded and unmasked may have their key and value vectors influenced by updates to other tokens during subsequent diffusion iterations. Thus, caching in dLLM and dMLLM are not theoretically lossless. 

For dLLMs, dKV-Cache~\cite{ma2025dkv} and dLLM-Cache~\cite{liu2025dllm} develops the naive KV-Cache techniques. Their observations are that, with small update intervals (\textit{e.g.}, 2 to 8), caching in dLLMs leads to minimal performance degradation and achieves \~10x speed-up.
For dMLLMs, Dimple~\cite{yu2025dimple} and LaViDa~\cite{li2025lavida} empirically verify that the use of prefilling incurs negligible performance degradation on the majority of vision-language benchmarks and provides a 
 2× to 7× speed-up. 

Following are some representative designs for KV-Cache.
\begin{itemize}\setlength\itemsep{4pt}
  \item \textbf{dKV-Cache} \cite{ma2025dkv}.  
The core idea of dKV-Cache is to cache the key-value pairs when the tokens are unmasked and reuse the cached key-value pairs with an interval hyperparameter to periodically update the cached key-value pairs.

\item \textbf{dLLM-Cache} \cite{liu2025dllm}. 
In addition to key-value (KV) pairs, dLLM-Cache stores attention (AttnOut) and feed-forward (FFNOut) outputs, reusing them in later steps. It applies different update intervals for prompt and response segments, with the prompt cache updated far less frequently. For responses, beyond periodic full updates, an Adaptive Partial Update strategy selectively refreshes tokens: cosine similarity between current and cached values identifies significant changes, and after each forward pass, a subset of tokens is proportionally updated.

\item \textbf{DualCache} \cite{wu2025fastdllm}. DualCache adopts a block-wise caching strategy, decoding text block by block. The block currently being decoded is not cached, while the surrounding blocks are cached. 
\end{itemize}

\begin{figure}[t]
    \centering
    \includegraphics[width=0.9\columnwidth]{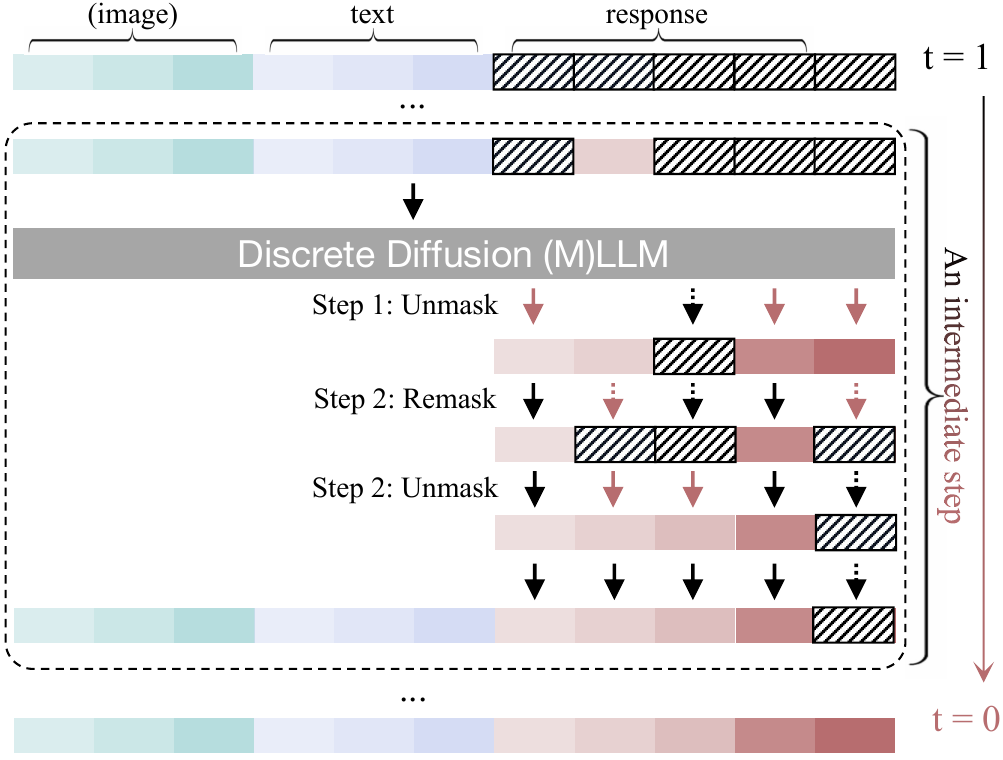}
    \caption{\textbf{Remasking} in General Discrete Diffusion Models.}
    \label{fig:remask}
\vspace{-1em}
\end{figure}

\subsection{Guidance}
\label{guidance}
\begin{figure*}[t!]
    \centering
\subfloat[Classifier-Free Guidance. \label{fig:guide_1}]{
  \includegraphics[width=0.3\textwidth]{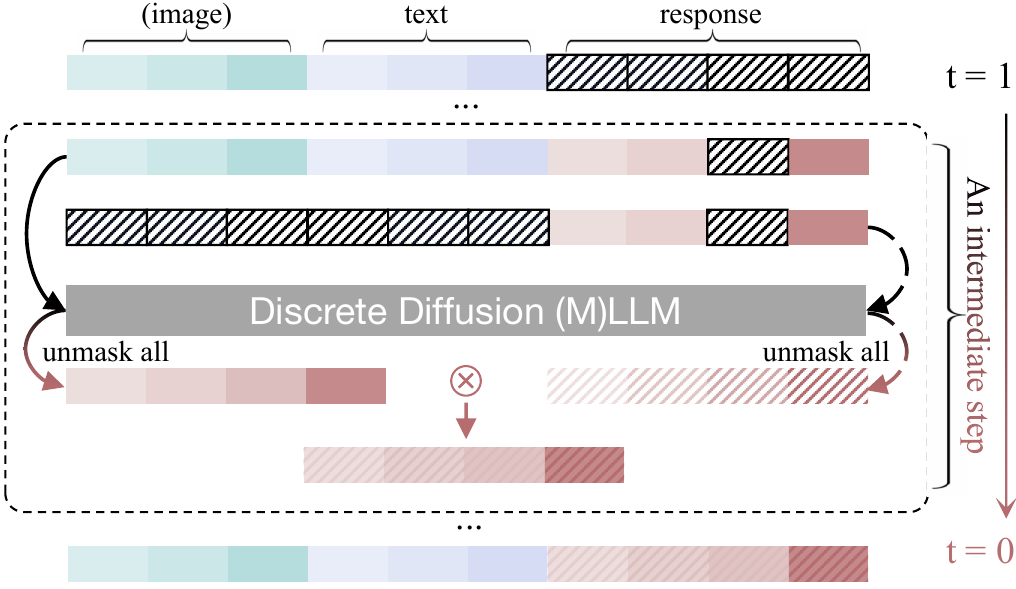}
}
\hfill
\subfloat[Classifier Guidance. \label{fig:guide_2}]{
  \includegraphics[width=0.33\textwidth]{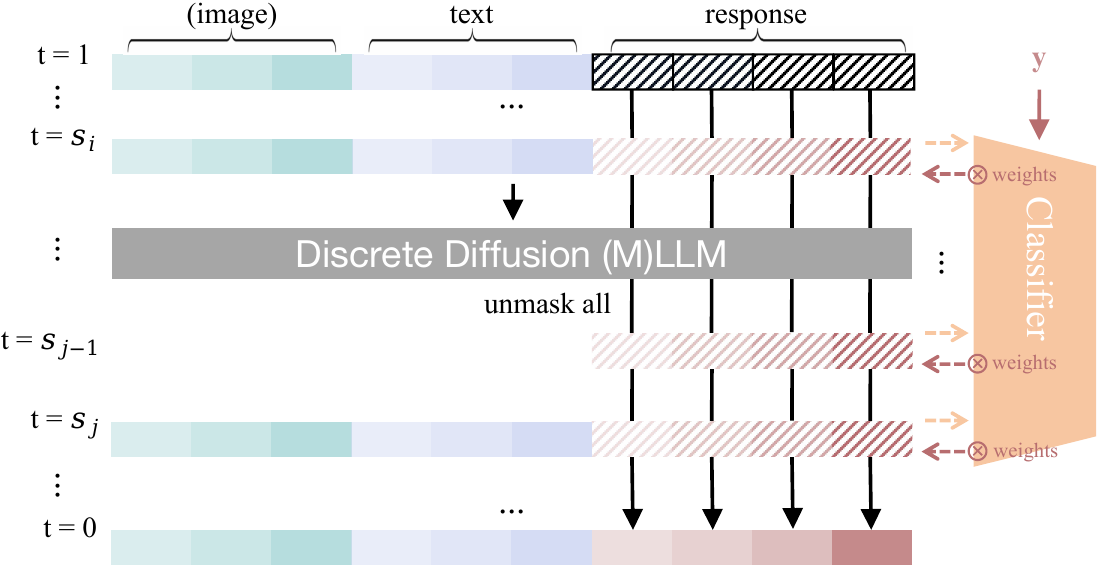}
} 
\hfill
\subfloat[Reward Guidance. \label{fig:guide_3}]{
  \includegraphics[width=0.33\textwidth]{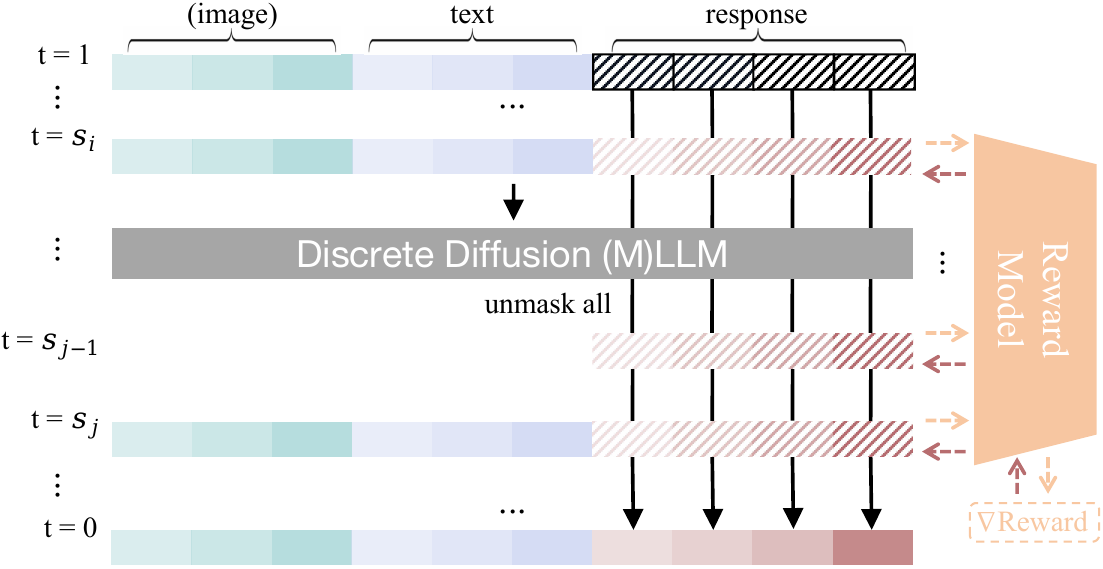}
} 
\caption{\textbf{Guidance Techniques.} We divide the guidance techniques (i.e., the post-processing on the predicted logits or sampling probabilities) into three categories: (a) Classifier-Free Guidance, (b) Classifier Guidance, (c) and Reward Guidance.}
\label{fig:guide_all}
\vspace{-1em}
\end{figure*}

In dLLMs and dMLLMs, post-processing on the predicted logits or sampling probabilities is commonly referred to as \emph{guidance}, following terminology from image diffusion models. 
Guidance methods are used to influence the generation process toward desired characteristics, such as improved diversity or controllability. 
\cref{fig:guide_all} provides an illustration of several guidance techniques. The detailed mathematical formulations are included in Appendix Sec.~D.II.
\emph{Classifier-free guidance} adjusts conditional predictions with unconditional ones to mitigate prompt-independent bias and enhance text diversity \cite{schiff2025simple,nie2025scalingmaskeddiffusionmodels}, though excessive guidance may degrade quality \cite{nisonoff2025unlocking,rojas2025theoryinformedimprovementsclassifierfreeguidance}. In contrast, \emph{classifier guidance} explicitly integrates class-conditional signals via an auxiliary classifier, enabling controllable generation across diffusion blocks \cite{schiff2025simple,huang2025ctrldiffboostinglargediffusion}. Beyond classifier-based methods, \emph{reward guidance}~\cite{tae2025tess} leverages a reward model to adjust logits at inference by gradient ascent,
thus steering outputs toward high-quality responses. Finally, \emph{energy-based diffusion} augments the denoising distribution with an energy function $E_\phi$ and reweights candidate samples via importance sampling, 
providing a theoretically grounded correction mechanism \cite{xu2025energybased}. Together, these techniques highlight the trade-offs between diversity, controllability, quality, and theoretical interpretability in guided discrete diffusion.

\subsection{Sampling Techniques}
\label{sampling}
The generation process of dLLMs relies on a multi-round iterative denoising procedure. Beyond strictly following the mathematical definition of the inverse transition in discrete diffusion, many techniques introduce additional control over the iterative denoising process to achieve better performance or efficiency, such as, majority voting based on decoding history and early stopping. In this subsection, we provide a summary of these techniques.

\subsubsection{Temporal Self-Consistency}
\cite{wang2025timefeatureexploitingtemporal} identifies a  phenomenon termed \emph{temporal oscillation}, where correct intermediate predictions emerge during the iterative decoding steps but are later overwritten by incorrect outputs in subsequent steps. 

To mitigate the performance degradation caused by temporal oscillation, rather than relying exclusively on the final denoising step, \cite{wang2025timefeatureexploitingtemporal} utilizes the predictions at all timesteps and predicts through a weighted voting scheme:
\begin{equation}
\text{answer}^\ast = \arg\max_a \sum_{t=1}^T f(t) \cdot \mathbf{1}\big(\text{semantic meaning}(x^t_0) = a\big),
\end{equation}
where $x^t_0$ denotes the prediction at time step $t$, and $\text{semantic meaning}(x^t_0)$ represents the extracted answer from $x^t_0$, $\mathbf{1}(\cdot)$ is the indicator function and $f(t)$ is a weighting function (e.g., exponential decay) assigning more importance to later steps.
\cite{wang2025timefeatureexploitingtemporal} also introduces Temporal Semantic Entropy (TSE), a measure of semantic uncertainty across intermediate predictions. The computation of TSE is based on the semantic clustering of historical answers. This metric can be incorporated into the reinforcement learning advantage function, thereby mitigating the issue of temporal oscillation in the model.

\subsubsection{Particle Gibbs Sampling for Discrete Diffusion Models}
Instead of sampling only once to obtain a single trajectory,
\cite{dang2025inferencetimescalingdiffusionlanguage} introduces a reference trajectory that is iteratively refined across multiple rounds of sampling.

The method begins by drawing an initial reference trajectory $x'_{0:T}$ using Sequential Monte Carlo (SMC) with a single particle. At each subsequent iteration, a set of $k$ particles is generated, where one particle is deterministically fixed to the reference trajectory while the remaining $k-1$ particles are defined as the candidate particles and are initialized randomly. At each step $t$, candidate particles are propagated according to the model's denoising distribution, $\bar{x}^{(i)}_{t-1} \sim p_\theta(x_{t-1} \mid c, x^{(i)}_t)$,
with the reference particle $\bar{x}^{(k)}_{t-1}$ kept fixed. The importance weight are for each candidate particles at time $t$ is defined based on a reward function $r$:
\begin{equation}
w^{(i)}_{t-1} = \exp \left( \frac{r(c,\bar{x}^{(i)}_{t-1}) - r(c,x^{(i)}_t)}{\beta} \right),
\end{equation}
and normalized before resampling. After resampling, the reference trajectory is updated by selecting one trajectory from the set with probability proportional to its importance weight. Iterating this procedure $m$ times yields a refined trajectory distribution that converges to the reward-weighted posterior.

\subsubsection{Early Stopping}
\cite{li2025diffusionlanguagemodelsknow} finds that diffusion models can often generate the correct answer before completing the entire decoding process. Therefore, discrete diffusion can leverage early stopping to improve decoding efficiency. Both \cite{li2025diffusionlanguagemodelsknow} and \cite{jin2025thinkinginsidemaskinplace} employ either predefined response templates or in-place prompts to divide the response space into two parts: the final answer region and the reasoning region. Once the confidence in the final answer region reaches a sufficient level, the subsequent decoding is terminated.

\subsection{Context Length Extension}
\label{context}
A notable phenomenon observed in dLLMs is that, with bidriectional attention, their robustness in handling extended contexts. Unlike autoregressive models that fail beyond the pretraining context window, diffusion LLMs can still  retrieve information from the most recent segment of the input, even at depths far beyond the training context length.

To further extend the context window,  LongLLaDA~\cite{liu2025longlladaunlockinglongcontext}  applies the NTK-aware scaling method to RoPE within diffusion LLMs. Following \cite{LocalLLaMA_dynamically_scaled_RoPE_2023}, the critical dimension $d_{\text{extra}}$ is:
\begin{equation}
d_{\text{extra}} = 2 \cdot \left\lceil \frac{d}{2} \cdot \log_{\beta_0} \left( \frac{T_{\text{train}}}{2\pi} \right) \right\rceil,
\end{equation}
where $d$ is the hidden dimension, $\beta_0$ is the rotary base, and $T_{\text{train}}$ is the pretraining context length. Given a target context length $t$, the NTK scaling factor $\lambda$ is chosen as:
\begin{equation}
\lambda = \beta_0^{-1} \cdot \left(\frac{t}{2\pi}\right)^{d/d_{\text{extra}}}.
\end{equation}

\subsection{Sparse Computation}
\label{sparse}
The forward computation of diffusion-based large language models exhibits inherent sparsity, allowing inference efficiency to be improved by eliminating part of the computations.

Under the block-wise decoding setup, many techniques have been explored. Sparse-dLLM~\cite{song2025sparsedllmacceleratingdiffusionllms} indicates that it is unnecessary to cache and reuse all key–value pairs of tokens outside the current block. By applying an average attention score between non-current tokens and tokens in the current block, a subset of tokens can be filtered out, thereby reducing the memory consumption of the KV cache and improving inference throughput. DPad~\cite{chen2025dpadefficientdiffusionlanguage} demonstrates that tokens in subsequent blocks do not need to attend to all tokens in the current block. Instead, a random dropout can be applied, with a higher dropout rate assigned to tokens that are farther from the current block.
Pipelined Parallel decoding~\cite{wang2025diffusion} adopts a ``soft'' block-wise decoding strategy. Instead of waiting for the current block to be fully decoded before moving to the next, it begins decoding tokens in the subsequent block once the proportion of decoded tokens in the last active block exceeds a predefined threshold. This gradual inclusion of additional blocks reduces computation on the right-side tokens and thereby improves efficiency.

\subsection{Response Length Control}
\label{length}
Adaptive control of response length is an essential capability in discrete diffusion models, and it can be achieved through either training-based or training-free approaches. In training-based methods, DreamOn~\cite{Dreamon2025} enables dynamic adjustment of token counts by introducing two special tokens, expand and delete. Similarly, FlexMDM~\cite{kim2025anyorderflexiblelengthmasked} employs an auxiliary network to predict how many tokens should be inserted before each existing token, thereby allowing the response length to grow dynamically. In contrast, the following section introduces training-free techniques for adaptive response length control.

DAEDAL~\cite{li2025fixedtrainingfreevariablelengthdenoising}  introduces a two-stage adaptive mechanism for response length control. It uses the confidence in predicting the End-of-Sequence (EOS) token as a signal of length sufficiency. 
First, it starts with a short sequence of length $L_{\text{init}}$ and repeatedly appends the response length until the EOS confidence exceeds $\tau_{\text{eos}}$ or the maximum length $L_{\max}$ is reached. Second, while  denoising, low-confidence tokens ($p_{\theta}(x_t[i]) < \tau_{\text{low}}$) trigger expansion: the least confident position is replaced by $E_{\text{factor}}$ mask tokens, allowing the sequence to grow adaptively.
\section{Quantization}
\label{sec:quantization}
\cite{lin2025quantization} presents the first systematic study of post-training quantization (PTQ) on dLLMs, benchmarking mainstream quantization methods across multiple models. Their analysis highlights severe activation outliers in dLLMs, sensitivity differences across tasks, and the relative robustness of instruction-tuned variants compared to base models. 

\cite{xu2025dllmquant} also analyzes why conventional PTQ methods degrade severely on dLLMs, identifying three key challenges: (1) quantization errors accumulate across iterations, (2) distinct token distributions across decoding steps, and (3) significant disparities in feature distributions across both token and channel dimensions. To address these, they introduce \textbf{DLLMQuant}, a framework featuring Temporal-Mask Adaptive Sampling (TMAS), Interaction-Aware Activation Quantization (IA-AQ), and Certainty-Guided Quantization (CGQ). The details of these techniques are discussed in Appendix Sec.E.I.
\section{Privacy and Safety}
\label{sec:safety}
dLLMs introduce unique safety vulnerabilities stemming from their \emph{bidirectional context modeling} and \emph{parallel decoding} mechanisms. These features, while enabling efficient infilling and interactive generation, also weaken the defenses that are effective in autoregressive (AR) LLMs. 

Diffusion-based LLMs Jailbreak Attack (DIJA)~\cite{wen2025devilmaskemergentsafety} and PArallel Decoding jailbreak (PAD)\cite{zhang2025jailbreaking}  reformulates conventional jailbreak prompts into an interleaved mask-text format, compelling the model to generate unsafe outputs while maintaining contextual consistency. Formally, let $a = (a_1, \ldots, a_R)$ denote a vanilla jailbreak prompt, and let $m = ([MASK], \ldots, [MASK])_Q$ represent $Q$ consecutive mask tokens. DIJA and PAD construct an interleaved prompt
\begin{equation}
p_i = a \oplus (m \otimes w),
\end{equation}
where $\oplus$ denotes concatenation, $\otimes$ denotes interleaving, and $w$ represents benign separator text, such as, ``Step 1'' and ''Step 2''. Importantly, the hazardous intent contained in $a$ is preserved, while critical instructions are forced into masked positions.

\cite{xie2025start} shows that dLLMs are more vulnerable to manipulation at the middle of the response than at the initial tokens and previous optimization-based jailbreak methods are effective at manipulating initial tokens but largely fail to optimize middle tokens. Thus, \cite{xie2025start} proposes the Middle tOken Safety Alignment (MOSA), a reinforcement learning alignment strategy, which requires the model’s generated middle tokens to align with a set of predefined safe tokens. 
Experimental evidence shows that optimizing alignment in the middle portion of the sequence is more effective than optimizing alignment at the beginning. 
\section{Applications}
\label{app}
\subsection{Language Related Applications}
\label{app_1}
Recent work applies diffusion language models to language tasks. \cite{lyu2023fine} introduce a diffusion-based model for fine-grained style transfer. \cite{zhang2025diffusion} leverage the bidirectional nature of diffusion for text embeddings, achieving strong reasoning performance. To enable controllable generation, \cite{zhu2024segment} propose Segment-Level Diffusion, decoding segments sequentially. In applications, \cite{cheng2024diffuspoll} use diffusion with task-specific masking and attribute tags to generate poll options; \cite{hu2024poetrydiffusion} enforce both semantics and meter in poetry; and \cite{padole2025improving} apply masked diffusion with verifier guidance for style control. EdiText \cite{lee2025editext} extends controllable generation to text editing, operating at both coarse and fine levels to achieve target attributes. 
Moving from editing to long-form generation, \cite{do2025discrete} design a discrete diffusion model for abstractive summarization. 
Complementing this direction, DiffETM \cite{shao2025diffetm} inject diffusion into the embedded topic model, enabling document–topic distributions to be sampled through a more realistic stochastic process. The $\textrm{CDA}^2$ framework~\cite{xin2025cdaˆ2} uses counterfactual diffusion augmentation to improve cross-domain sentiment adaptation. DiffusionCLS~\cite{chen2024effective} enhances low-resource classification by generating label-consistent pseudo-samples via sentiment-relevant token reconstruction. For aspect sentiment quad prediction, GDP~\cite{zhu2024pinpointing} employs a template-guided diffusion strategy. In layout generation, \cite{iwai2024layout} mitigate layout sticking in discrete diffusion models by introducing Layout-Corrector, which scores and resets misplaced tokens for improved positioning. TermDiffuSum \cite{dong2025termdiffusum} integrates term-aware attention and a re-ranking loss during diffusion, effectively emphasizing legally legally salient sentences.

\subsection{Reasoning}
\label{app_2}
Diffusion-of-Thought (DoT)~\cite{ye2024diffusion} firstly integrates chain-of-thought reasoning into dLLMs to enhance reasoning capabilities.  DiffuCOMET~\cite{gao2024diffucomet} develops a series of models that leverage diffusion to infer contextually relevant commonsense knowledge from narratives. DPCL-Diff~\cite{cao2025dpcl} combines graph node diffusion with dual-domain periodic contrastive learning for temporal knowledge graph reasoning. The d1 framework~\cite{zhao2025d1} adapts pretrained dLLMs into reasoning models via a combination of supervised fine-tuning and reinforcement learning. It introduces a novel critic-free policy-gradient algorithm called diffu-GRPO and employs masked SFT to distill reasoning knowledge directly from existing datasets. \cite{ye2025beyond} provides an insight into dLLM reasoning: the difficulty of decoding individual tokens in a response varies, and it is not necessarily the case that tokens on the left are easier to decode than those on the right. NeSyDM \cite{van2025neurosymbolic} introduces a discrete diffusion process in the symbolic space. The framework first extracts symbolic representations from perceptual inputs and then performs a multi-step denoising diffusion process over these symbols.

\subsection{Vision and Multimodal}
\label{app_3}
UDAN-CLIP \cite{shaahid2025underwater} proposes an image-to-image diffusion framework for underwater enhancement. 
Turning to motion synthesis, M2D2M \cite{chi2024m2d2m} employs a discrete diffusion model to generate continuous human-motion sequences from textual descriptions of multiple actions.
AR-Diffusion \cite{sun2025ar} introduces a novel architecture that blends autoregressive and diffusion techniques for flexible, asynchronous video generation. 
Besides the understanding tasks, discrete diffusion is also largely used in the vision generation tasks~\cite{chang2022maskgit,bai2024meissonic}.

\subsection{Robotics and Autonomous Driving}
\label{app_4}
DiffVLA \cite{jiang2025diffvla} introduces a vision-language-guided diffusion policy for autonomous-driving planning.
ViLaD~\cite{cui2025vilad} introduces a large vision–language diffusion framework for end-to-end autonomous driving, addressing the latency and unidirectional limitations of autoregressive VLM-based decision models. Meanwhile, Discrete Diffusion VLA \cite{liang2025discrete} unifies vision, language, and action decoding in a single transformer. It discretizes actions into tokens and employs iterative re-masking for adaptive, parallel action generation.
Extending diffusion policies to robotics, VPDD~\cite{he2024learning} first pre-trains on large-scale actionless human videos and then transfers the learned discrete diffusion policy to robot-control tasks. 
Discrete-Guided Diffusion (DGD)~\cite{liang2025discreteGuided} integrates discrete multi-agent path finding (MAPF) with diffusion models to address scalable and safe multi-robot motion planning. 

\subsection{Graph and Structured Predictions}
\label{app_5}
LO-ARM \cite{wang2025learning} introduces a learning-order autoregressive model that dynamically adapts generation order for molecular graphs, improving flexibility and validity. ReDiSC \cite{li2025redisc} proposes a reparameterized discrete masked diffusion model for node classification and achieves scalable and interpretable predictions on large graphs. Scaffold Diffusion \cite{jung2025scaffolddiffusionsparsemulticategory} formulates sparse multi-category voxels as discrete token sequences and applies a dLLM for 3D sparse structure generation.

\subsection{Biological and Drug Discovery}
\label{app_8}
A molecular-editing framework named MolEditRL \cite{zhuang2025moleditrl} combines a discrete graph-diffusion model with reinforcement learning to optimize molecular properties while preserving structural similarity. CFP-Gen \cite{yin2025cfp} adopts a diffusion language model for combinatorial functional protein generation.
TransDLM \cite{xiong2024text} proposes a text-guided, multi-property molecular-optimization method that leverages a diffusion language model. 
GenMol \cite{lee2025genmol} presents a single, discrete diffusion model that serves as a versatile generator across diverse pharmaceutical tasks. 
DPLM-2 \cite{wang2024dplm} is a multimodal protein language model capable of understanding and generating both protein sequences and their 3D structures. 
PepTune \cite{tang2025peptune} targets therapeutic-peptide design with a multi-objective, discrete diffusion framework built on a masked language model backbone.
\cite{zhang2025cross} propose CMCM-DLM, which integrates structure-control and property-control modules into a pretrained dLLM for molecules. 
\begin{figure}[t]
\centering
    \includegraphics[width=0.8\linewidth]{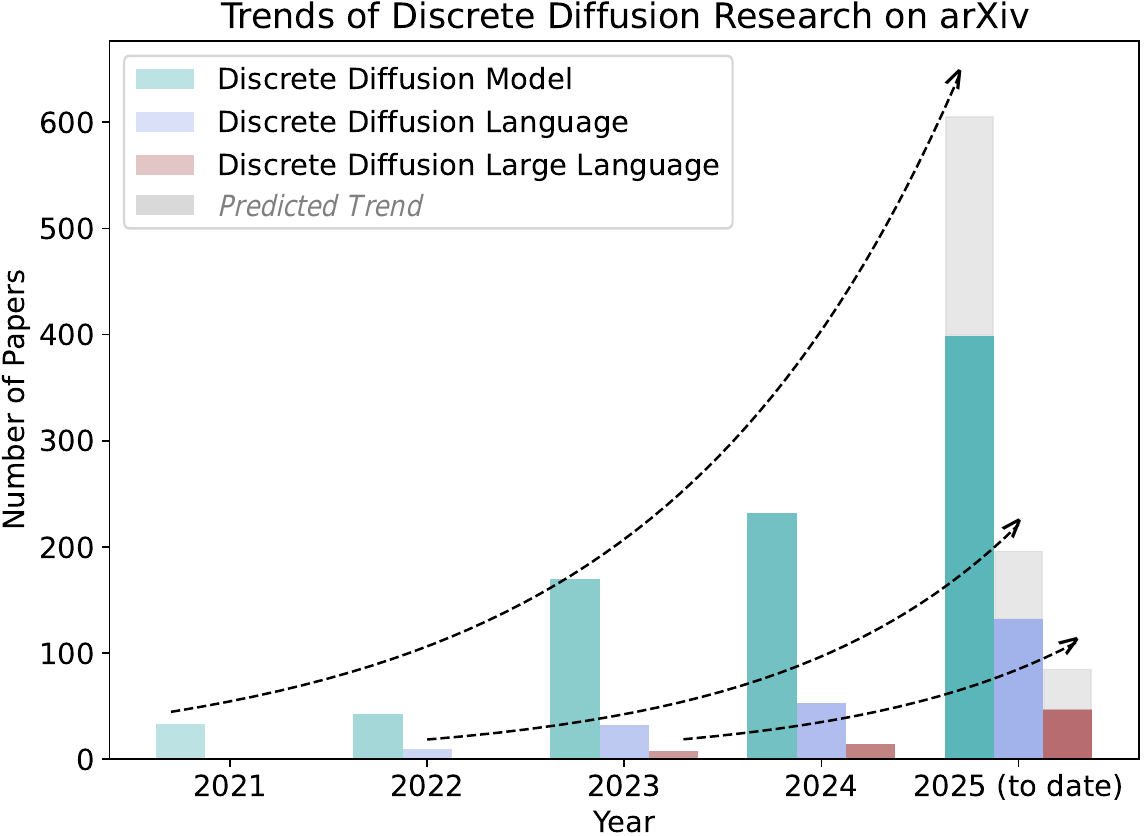}
    \caption{Number of arXiv publications retrieved via keyword-based search (\textit{Discrete Diffusion Model}, \textit{Discrete Diffusion Language}, and \textit{Discrete Diffusion Large Language}) under the \textit{Computer Science (cs)} category using the \textit{All fields} search option, which scans across all metadata including titles, abstracts, and author information. The results show a consistent year-over-year increase, reflecting the growing research interest in this area.}
\label{fig_year}
\vspace{-1em}
\end{figure}

\section{Conclusion}
In summary, this survey provides a comprehensive overview of Discrete Diffusion Large Language Models (dLLMs) and Discrete Diffusion Large Multimodal Models (dMLLMs). We present a detailed exposition of their mathematical foundations and landmark developments. We further detail the training and inference strategies behind them, and summarize the current application domains and potential future directions of them. As a promising alternative to autoregressive LLMs, dLLMs have attracted growing attention (see Figure \ref{fig_year}) and show great potential in a variety of real-world scenarios. We hope this survey will serve as a valuable foundation for future research and development in this fast-evolving and important field. At the end of the Appendix, we also discuss some future directions of dLLM and dMLLM.

\clearpage
\begingroup
\footnotesize
\bibliographystyle{IEEEtran}
\bibliography{main}

\begin{thebibliography}{100}
\providecommand{\url}[1]{#1}
\csname url@samestyle\endcsname
\providecommand{\newblock}{\relax}
\providecommand{\bibinfo}[2]{#2}
\providecommand{\BIBentrySTDinterwordspacing}{\spaceskip=0pt\relax}
\providecommand{\BIBentryALTinterwordstretchfactor}{4}
\providecommand{\BIBentryALTinterwordspacing}{\spaceskip=\fontdimen2\font plus
\BIBentryALTinterwordstretchfactor\fontdimen3\font minus \fontdimen4\font\relax}
\providecommand{\BIBforeignlanguage}[2]{{%
\expandafter\ifx\csname l@#1\endcsname\relax
\typeout{** WARNING: IEEEtran.bst: No hyphenation pattern has been}%
\typeout{** loaded for the language `#1'. Using the pattern for}%
\typeout{** the default language instead.}%
\else
\language=\csname l@#1\endcsname
\fi
#2}}
\providecommand{\BIBdecl}{\relax}
\BIBdecl

\bibitem{openai2024gpt4ocard}
{OpenAI\ }, ``Gpt-4o system card,'' 2024.

\bibitem{openai2024gpt4technicalreport}
{OpenAI}, ``Gpt-4 technical report,'' 2024.

\bibitem{deepseekai2025deepseekr1incentivizingreasoningcapability}
DeepSeek-AI, ``Deepseek-r1: Incentivizing reasoning capability in llms via reinforcement learning,'' 2025.

\bibitem{geminiteam2024gemini15unlockingmultimodal}
{Gemini Team\ }, ``Gemini 1.5: Unlocking multimodal understanding across millions of tokens of context,'' 2024.

\bibitem{geminiteam2025geminifamilyhighlycapable}
{Gemini Team}, ``Gemini: A family of highly capable multimodal models,'' 2025.

\bibitem{nie2025large}
S.~Nie, F.~Zhu, Z.~You, X.~Zhang, J.~Ou, J.~Hu, J.~Zhou, Y.~Lin, J.-R. Wen, and C.~Li, ``Large language diffusion models,'' \emph{arXiv preprint arXiv:2502.09992}, 2025.

\bibitem{dream2025}
J.~Ye, Z.~Xie, L.~Zheng, J.~Gao, Z.~Wu, X.~Jiang, Z.~Li, and L.~Kong, ``Dream 7b: Diffusion large language models,'' 2025.

\bibitem{yu2025dimple}
R.~Yu, X.~Ma, and X.~Wang, ``Dimple: Discrete diffusion multimodal large language model with parallel decoding,'' \emph{arXiv preprint arXiv:2505.16990}, 2025.

\bibitem{you2025llada}
Z.~You, S.~Nie, X.~Zhang, J.~Hu, J.~Zhou, Z.~Lu, J.-R. Wen, and C.~Li, ``Llada-v: Large language diffusion models with visual instruction tuning,'' \emph{arXiv preprint arXiv:2505.16933}, 2025.

\bibitem{li2025lavida}
S.~Li, K.~Kallidromitis, H.~Bansal, A.~Gokul, Y.~Kato, K.~Kozuka, J.~Kuen, Z.~Lin, K.-W. Chang, and A.~Grover, ``Lavida: A large diffusion language model for multimodal understanding,'' \emph{arXiv preprint arXiv:2505.16839}, 2025.

\bibitem{yang2025mmada}
L.~Yang, Y.~Tian, B.~Li, X.~Zhang, K.~Shen, Y.~Tong, and M.~Wang, ``Mmada: Multimodal large diffusion language models,'' \emph{arXiv preprint arXiv:2505.15809}, 2025.

\bibitem{deepmind_gemini_diffusion}
{DeepMind}, ``Gemini diffusion,'' \url{https://deepmind.google/models/gemini-diffusion/}, 2025, accessed: 2025-06-16.

\bibitem{inceptionlabs_mercury}
{Inception Labs}, ``Mercury,'' \url{https://www.inceptionlabs.ai/introducing-mercury}, 2025, accessed: 2025‑06‑16.

\bibitem{prabhudesai2025diffusionbeatsautoregressivedataconstrained}
M.~Prabhudesai, M.~Wu, A.~Zadeh, K.~Fragkiadaki, and D.~Pathak, ``Diffusion beats autoregressive in data-constrained settings,'' 2025.

\bibitem{austin2021structured}
J.~Austin, D.~D. Johnson, J.~Ho, D.~Tarlow, and R.~Van Den~Berg, ``Structured denoising diffusion models in discrete state-spaces,'' \emph{Advances in neural information processing systems}, vol.~34, pp. 17\,981--17\,993, 2021.

\bibitem{hoogeboom2021argmax}
E.~Hoogeboom, D.~Nielsen, P.~Jaini, P.~Forr{\'e}, and M.~Welling, ``Argmax flows and multinomial diffusion: Learning categorical distributions,'' \emph{Advances in neural information processing systems}, vol.~34, pp. 12\,454--12\,465, 2021.

\bibitem{zheng2023reparameterized}
L.~Zheng, J.~Yuan, L.~Yu, and L.~Kong, ``A reparameterized discrete diffusion model for text generation,'' \emph{arXiv preprint arXiv:2302.05737}, 2023.

\bibitem{sahoo2024simple}
S.~Sahoo, M.~Arriola, Y.~Schiff, A.~Gokaslan, E.~Marroquin, J.~Chiu, A.~Rush, and V.~Kuleshov, ``Simple and effective masked diffusion language models,'' \emph{Advances in Neural Information Processing Systems}, vol.~37, pp. 130\,136--130\,184, 2024.

\bibitem{shi2024simplified}
J.~Shi, K.~Han, Z.~Wang, A.~Doucet, and M.~Titsias, ``Simplified and generalized masked diffusion for discrete data,'' \emph{Advances in neural information processing systems}, vol.~37, pp. 103\,131--103\,167, 2024.

\bibitem{gong2024scaling}
S.~Gong, S.~Agarwal, Y.~Zhang, J.~Ye, L.~Zheng, M.~Li, C.~An, P.~Zhao, W.~Bi, J.~Han \emph{et~al.}, ``Scaling diffusion language models via adaptation from autoregressive models,'' \emph{arXiv preprint arXiv:2410.17891}, 2024.

\bibitem{nie2025scalingmaskeddiffusionmodels}
S.~Nie, F.~Zhu, C.~Du, T.~Pang, Q.~Liu, G.~Zeng, M.~Lin, and C.~Li, ``Scaling up masked diffusion models on text,'' 2025.

\bibitem{sohl2015deep}
J.~Sohl-Dickstein, E.~Weiss, N.~Maheswaranathan, and S.~Ganguli, ``Deep unsupervised learning using nonequilibrium thermodynamics,'' in \emph{International conference on machine learning}.\hskip 1em plus 0.5em minus 0.4em\relax pmlr, 2015, pp. 2256--2265.

\bibitem{haxholli2025efficient}
E.~Haxholli, Y.~Z. Gurbuz, O.~Can, and E.~Waxman, ``Efficient perplexity bound and ratio matching in discrete diffusion language models,'' in \emph{The Thirteenth International Conference on Learning Representations}, 2025.

\bibitem{rutte2025generalized}
D.~von R{\"u}tte, J.~Fluri, Y.~Ding, A.~Orvieto, B.~Sch{\"o}lkopf, and T.~Hofmann, ``Generalized interpolating discrete diffusion,'' in \emph{Forty-second International Conference on Machine Learning}, 2025.

\bibitem{continuoustimedlm}
A.~Campbell, J.~Benton, V.~De~Bortoli, T.~Rainforth, G.~Deligiannidis, and A.~Doucet, ``A continuous time framework for discrete denoising models,'' in \emph{Advances in Neural Information Processing Systems}, S.~Koyejo, S.~Mohamed, A.~Agarwal, D.~Belgrave, K.~Cho, and A.~Oh, Eds., vol.~35.\hskip 1em plus 0.5em minus 0.4em\relax Curran Associates, Inc., 2022, pp. 28\,266--28\,279.

\bibitem{concrete_score}
C.~Meng, K.~Choi, J.~Song, and S.~Ermon, ``Concrete score matching: Generalized score matching for discrete data,'' in \emph{Advances in Neural Information Processing Systems}, S.~Koyejo, S.~Mohamed, A.~Agarwal, D.~Belgrave, K.~Cho, and A.~Oh, Eds., vol.~35.\hskip 1em plus 0.5em minus 0.4em\relax Curran Associates, Inc., 2022, pp. 34\,532--34\,545.

\bibitem{sun2023scorebased}
H.~Sun, L.~Yu, B.~Dai, D.~Schuurmans, and H.~Dai, ``Score-based continuous-time discrete diffusion models,'' in \emph{Proceedings of the International Conference on Learning Representations (ICLR)}, 2023.

\bibitem{ou2025absorbingdiscretediffusionsecretly}
J.~Ou, S.~Nie, K.~Xue, F.~Zhu, J.~Sun, Z.~Li, and C.~Li, ``Your absorbing discrete diffusion secretly models the conditional distributions of clean data,'' 2025.

\bibitem{zhang2025target}
R.~ZHANG, S.~Zhai, Y.~Zhang, J.~Thornton, Z.~Ou, J.~M. Susskind, and N.~Jaitly, ``Target concrete score matching: A holistic framework for discrete diffusion,'' in \emph{Forty-second International Conference on Machine Learning}, 2025.

\bibitem{dfm}
I.~Gat, T.~Remez, N.~Shaul, F.~Kreuk, R.~T.~Q. Chen, G.~Synnaeve, Y.~Adi, and Y.~Lipman, ``Discrete flow matching,'' in \emph{Advances in Neural Information Processing Systems}, A.~Globerson, L.~Mackey, D.~Belgrave, A.~Fan, U.~Paquet, J.~Tomczak, and C.~Zhang, Eds., vol.~37.\hskip 1em plus 0.5em minus 0.4em\relax Curran Associates, Inc., 2024, pp. 133\,345--133\,385.

\bibitem{arriola2025block}
M.~Arriola, A.~Gokaslan, J.~T. Chiu, Z.~Yang, Z.~Qi, J.~Han, S.~S. Sahoo, and V.~Kuleshov, ``Block diffusion: Interpolating between autoregressive and diffusion language models,'' in \emph{The Thirteenth International Conference on Learning Representations}, 2025.

\bibitem{kim2025anyorderflexiblelengthmasked}
J.~Kim, L.~Cheuk-Kit, C.~Domingo-Enrich, Y.~Du, S.~Kakade, T.~Ngotiaoco, S.~Chen, and M.~Albergo, ``Any-order flexible length masked diffusion,'' 2025.

\bibitem{chao2025maskedunmaskeddiscretediffusion}
C.-H. Chao, W.-F. Sun, H.~Liang, C.-Y. Lee, and R.~G. Krishnan, ``Beyond masked and unmasked: Discrete diffusion models via partial masking,'' 2025.

\bibitem{zhang2025flexiblelengthtextinfillingdiscrete}
A.~Zhang, A.~Sivakumar, C.~Tang, and C.~Thomas, ``Flexible-length text infilling for discrete diffusion models,'' 2025.

\bibitem{he2022diffusionbert}
Z.~He, T.~Sun, K.~Wang, X.~Huang, and X.~Qiu, ``Diffusionbert: Improving generative masked language models with diffusion models,'' \emph{arXiv preprint arXiv:2211.15029}, 2022.

\bibitem{chen2023cheaper}
J.~Chen, A.~Zhang, M.~Li, A.~Smola, and D.~Yang, ``A cheaper and better diffusion language model with soft-masked noise,'' \emph{arXiv preprint arXiv:2304.04746}, 2023.

\bibitem{zhou2023diffusion}
K.~Zhou, Y.~Li, W.~X. Zhao, and J.-R. Wen, ``Diffusion-nat: Self-prompting discrete diffusion for non-autoregressive text generation,'' \emph{arXiv preprint arXiv:2305.04044}, 2023.

\bibitem{mahabadi2023tess}
R.~K. Mahabadi, H.~Ivison, J.~Tae, J.~Henderson, I.~Beltagy, M.~E. Peters, and A.~Cohan, ``Tess: Text-to-text self-conditioned simplex diffusion,'' \emph{arXiv preprint arXiv:2305.08379}, 2023.

\bibitem{gulrajani2023likelihood}
I.~Gulrajani and T.~B. Hashimoto, ``Likelihood-based diffusion language models,'' \emph{Advances in Neural Information Processing Systems}, vol.~36, pp. 16\,693--16\,715, 2023.

\bibitem{lou2023discrete}
A.~Lou, C.~Meng, and S.~Ermon, ``Discrete diffusion modeling by estimating the ratios of the data distribution,'' \emph{arXiv preprint arXiv:2310.16834}, 2023.

\bibitem{swerdlow2025unified}
A.~Swerdlow, M.~Prabhudesai, S.~Gandhi, D.~Pathak, and K.~Fragkiadaki, ``Unified multimodal discrete diffusion,'' \emph{arXiv preprint arXiv:2503.20853}, 2025.

\bibitem{ye2023diffusion}
J.~Ye, Z.~Zheng, Y.~Bao, L.~Qian, and Q.~Gu, ``Diffusion language models can perform many tasks with scaling and instruction-finetuning,'' \emph{arXiv preprint arXiv:2308.12219}, 2023.

\bibitem{zhu2025llada}
F.~Zhu, R.~Wang, S.~Nie, X.~Zhang, C.~Wu, J.~Hu, J.~Zhou, J.~Chen, Y.~Lin, J.-R. Wen \emph{et~al.}, ``Llada 1.5: Variance-reduced preference optimization for large language diffusion models,'' \emph{arXiv preprint arXiv:2505.19223}, 2025.

\bibitem{tae2025tess}
J.~Tae, H.~Ivison, S.~Kumar, and A.~Cohan, ``Tess 2: A large-scale generalist diffusion language model,'' \emph{arXiv preprint arXiv:2502.13917}, 2025.

\bibitem{Dreamon2025}
\BIBentryALTinterwordspacing
Z.~Wu, L.~Zheng, Z.~Xie, J.~Ye, J.~Gao, Y.~Feng, Z.~Li, V.~W., G.~Zhou, and L.~Kong, ``Dreamon: Diffusion language models for code infilling beyond fixed-size canvas,'' 2025. [Online]. Available: \url{https://hkunlp.github.io/blog/2025/dreamon}
\BIBentrySTDinterwordspacing

\bibitem{dreamcoder2025}
\BIBentryALTinterwordspacing
Z.~Xie, J.~Ye, L.~Zheng, J.~Gao, J.~Dong, Z.~Wu, X.~Zhao, S.~Gong, X.~Jiang, Z.~Li, and L.~Kong, ``Dream-coder 7b,'' 2025. [Online]. Available: \url{https://hkunlp.github.io/blog/2025/dream-coder}
\BIBentrySTDinterwordspacing

\bibitem{gong2025diffucoder}
S.~Gong, R.~Zhang, H.~Zheng, J.~Gu, N.~Jaitly, L.~Kong, and Y.~Zhang, ``Diffucoder: Understanding and improving masked diffusion models for code generation,'' \emph{arXiv preprint arXiv:2506.20639}, 2025.

\bibitem{song2025seed}
Y.~Song, Z.~Zhang, C.~Luo, P.~Gao, F.~Xia, H.~Luo, Z.~Li, Y.~Yang, H.~Yu, X.~Qu \emph{et~al.}, ``Seed diffusion: A large-scale diffusion language model with high-speed inference,'' \emph{arXiv preprint arXiv:2508.02193}, 2025.

\bibitem{wang2025fudoki}
J.~Wang, Y.~Lai, A.~Li, S.~Zhang, J.~Sun, N.~Kang, C.~Wu, Z.~Li, and P.~Luo, ``Fudoki: Discrete flow-based unified understanding and generation via kinetic-optimal velocities,'' \emph{arXiv preprint arXiv:2505.20147}, 2025.

\bibitem{shi2025muddit}
Q.~Shi, J.~Bai, Z.~Zhao, W.~Chai, K.~Yu, J.~Wu, S.~Song, Y.~Tong, X.~Li, X.~Li \emph{et~al.}, ``Muddit: Liberating generation beyond text-to-image with a unified discrete diffusion model,'' \emph{arXiv preprint arXiv:2505.23606}, 2025.

\bibitem{chang2022maskgit}
H.~Chang, H.~Zhang, L.~Jiang, C.~Liu, and W.~T. Freeman, ``Maskgit: Masked generative image transformer,'' in \emph{Proceedings of the IEEE/CVF conference on computer vision and pattern recognition}, 2022, pp. 11\,315--11\,325.

\bibitem{sun2025blockwisesftdiffusionlanguage}
B.~Sun, Y.~Cai, M.-H. Yang, and Y.~Wang, ``Blockwise sft for diffusion language models: Reconciling bidirectional attention and autoregressive decoding,'' 2025.

\bibitem{ye2025beyond}
J.~Ye, J.~Gao, S.~Gong, L.~Zheng, X.~Jiang, Z.~Li, and L.~Kong, ``Beyond autoregression: Discrete diffusion for complex reasoning and planning,'' in \emph{The Thirteenth International Conference on Learning Representations}, 2025.

\bibitem{hayakawa2024distillation}
S.~Hayakawa, Y.~Takida, M.~Imaizumi, H.~Wakaki, and Y.~Mitsufuji, ``Distillation of discrete diffusion through dimensional correlations,'' \emph{arXiv preprint arXiv:2410.08709}, 2024.

\bibitem{han2025discretediffusiontrajectoryalignment}
J.~Han, A.~Wang, M.~Xu, W.~Chu, M.~Dang, Y.~Yue, and S.~Ermon, ``Discrete diffusion trajectory alignment via stepwise decomposition,'' 2025.

\bibitem{tang2025wd1weightedpolicyoptimization}
X.~Tang, R.~Dolga, S.~Yoon, and I.~Bogunovic, ``wd1: Weighted policy optimization for reasoning in diffusion language models,'' 2025.

\bibitem{huang2025reinforcingdiffusionchainlateral}
Z.~Huang, Z.~Chen, Z.~Wang, T.~Li, and G.-J. Qi, ``Reinforcing the diffusion chain of lateral thought with diffusion language models,'' 2025.

\bibitem{asada-miwa-2025-addressing}
M.~Asada and M.~Miwa, ``Addressing the training-inference discrepancy in discrete diffusion for text generation,'' in \emph{Proceedings of the 31st International Conference on Computational Linguistics}.\hskip 1em plus 0.5em minus 0.4em\relax Association for Computational Linguistics, Jan. 2025, pp. 7156--7164.

\bibitem{kim2025train}
J.~Kim, K.~Shah, V.~Kontonis, S.~M. Kakade, and S.~Chen, ``Train for the worst, plan for the best: Understanding token ordering in masked diffusions,'' in \emph{Forty-second International Conference on Machine Learning}, 2025.

\bibitem{benhamu2025acceleratedsamplingmaskeddiffusion}
H.~Ben-Hamu, I.~Gat, D.~Severo, N.~Nolte, and B.~Karrer, ``Accelerated sampling from masked diffusion models via entropy bounded unmasking,'' 2025.

\bibitem{huang2025pcsamplerpositionawarecalibrationdecoding}
P.~Huang, S.~Liu, Z.~Liu, Y.~Yan, S.~Wang, Z.~Chen, and T.~Xiao, ``Pc-sampler: Position-aware calibration of decoding bias in masked diffusion models,'' 2025.

\bibitem{luxembourg2025planspeeddilatedscheduling}
O.~Luxembourg, H.~Permuter, and E.~Nachmani, ``Plan for speed: Dilated scheduling for masked diffusion language models,'' 2025.

\bibitem{wang2025remasking}
G.~Wang, Y.~Schiff, S.~Sahoo, and V.~Kuleshov, ``Remasking discrete diffusion models with inference-time scaling,'' \emph{arXiv preprint arXiv:2503.00307}, 2025.

\bibitem{hong2025wideinnarrowoutrevokabledecoding}
F.~Hong, G.~Yu, Y.~Ye, H.~Huang, H.~Zheng, Y.~Zhang, Y.~Wang, and J.~Yao, ``Wide-in, narrow-out: Revokable decoding for efficient and effective dllms,'' 2025.

\bibitem{ma2025dkv}
X.~Ma, R.~Yu, G.~Fang, and X.~Wang, ``dkv-cache: The cache for diffusion language models,'' \emph{arXiv preprint arXiv:2505.15781}, 2025.

\bibitem{liu2025dllm}
Z.~Liu, Y.~Yang, Y.~Zhang, J.~Chen, C.~Zou, Q.~Wei, S.~Wang, and L.~Zhang, ``dllm-cache: Accelerating diffusion large language models with adaptive caching,'' \emph{arXiv preprint arXiv:2506.06295}, 2025.

\bibitem{wu2025fastdllm}
C.~Wu, H.~Zhang, S.~Xue, Z.~Liu, S.~Diao, L.~Zhu, P.~Luo, S.~Han, and E.~Xie, ``Fast-dllm: Training-free acceleration of diffusion llm by enabling kv cache and parallel decoding,'' 2025.

\bibitem{huang2025ctrldiffboostinglargediffusion}
C.~Huang and H.~Tang, ``Ctrldiff: Boosting large diffusion language models with dynamic block prediction and controllable generation,'' 2025.

\bibitem{xu2025energybased}
M.~Xu, T.~Geffner, K.~Kreis, W.~Nie, Y.~Xu, J.~Leskovec, S.~Ermon, and A.~Vahdat, ``Energy-based diffusion language models for text generation,'' in \emph{The Thirteenth International Conference on Learning Representations}, 2025.

\bibitem{li2025diffusionlanguagemodelsknow}
P.~Li, Y.~Zhou, D.~Muhtar, L.~Yin, S.~Yan, L.~Shen, Y.~Liang, S.~Vosoughi, and S.~Liu, ``Diffusion language models know the answer before decoding,'' 2025.

\bibitem{jin2025thinkinginsidemaskinplace}
X.~Jin, Y.~Wang, Y.~Gao, Z.~Wen, B.~Qi, D.~Liu, and L.~Zhang, ``Thinking inside the mask: In-place prompting in diffusion llms,'' 2025.

\bibitem{dang2025inferencetimescalingdiffusionlanguage}
M.~Dang, J.~Han, M.~Xu, K.~Xu, A.~Srivastava, and S.~Ermon, ``Inference-time scaling of diffusion language models with particle gibbs sampling,'' 2025.

\bibitem{wang2025timefeatureexploitingtemporal}
W.~Wang, B.~Fang, C.~Jing, Y.~Shen, Y.~Shen, Q.~Wang, H.~Ouyang, H.~Chen, and C.~Shen, ``Time is a feature: Exploiting temporal dynamics in diffusion language models,'' 2025.

\bibitem{liu2025longlladaunlockinglongcontext}
X.~Liu, Z.~Liu, Z.~Huang, Q.~Guo, Z.~He, and X.~Qiu, ``Longllada: Unlocking long context capabilities in diffusion llms,'' 2025.

\bibitem{song2025sparsedllmacceleratingdiffusionllms}
Y.~Song, X.~Liu, R.~Li, Z.~Liu, Z.~Huang, Q.~Guo, Z.~He, and X.~Qiu, ``Sparse-dllm: Accelerating diffusion llms with dynamic cache eviction,'' 2025.

\bibitem{chen2025dpadefficientdiffusionlanguage}
X.~Chen, S.~Huang, C.~Guo, C.~Wei, Y.~He, J.~Zhang, H.~H. Li, and Y.~Chen, ``Dpad: Efficient diffusion language models with suffix dropout,'' 2025.

\bibitem{wang2025diffusion}
X.~Wang, C.~Xu, Y.~Jin, J.~Jin, H.~Zhang, and Z.~Deng, ``Diffusion llms can do faster-than-ar inference via discrete diffusion forcing,'' \emph{arXiv preprint arXiv:2508.09192}, 2025.

\bibitem{li2025fixedtrainingfreevariablelengthdenoising}
J.~Li, X.~Dong, Y.~Zang, Y.~Cao, J.~Wang, and D.~Lin, ``Beyond fixed: Training-free variable-length denoising for diffusion large language models,'' 2025.

\bibitem{lin2025quantization}
H.~Lin, H.~Xu, Y.~Wu, Z.~Guo, R.~Zhang, Z.~Lu, Y.~Wei, Q.~Zhang, and Z.~Sun, ``Quantization meets dllms: A systematic study of post-training quantization for diffusion llms,'' \emph{arXiv preprint arXiv:2508.14896}, 2025.

\bibitem{xu2025dllmquant}
C.~Xu and D.~Yang, ``Dllmquant: Quantizing diffusion-based large language models,'' \emph{arXiv preprint arXiv:2508.14090}, 2025.

\bibitem{wen2025devilmaskemergentsafety}
Z.~Wen, J.~Qu, D.~Liu, Z.~Liu, R.~Wu, Y.~Yang, X.~Jin, H.~Xu, X.~Liu, W.~Li, C.~Lu, J.~Shao, C.~He, and L.~Zhang, ``The devil behind the mask: An emergent safety vulnerability of diffusion llms,'' 2025.

\bibitem{zhang2025jailbreaking}
Y.~Zhang, F.~Xie, Z.~Zhou, Z.~Li, H.~Chen, K.~Wang, and Y.~Guo, ``Jailbreaking large language diffusion models: Revealing hidden safety flaws in diffusion-based text generation,'' \emph{arXiv preprint arXiv:2507.19227}, 2025.

\bibitem{xie2025start}
Z.~Xie, X.~Song, and J.~Luo, ``Where to start alignment? diffusion large language model may demand a distinct position,'' \emph{arXiv preprint arXiv:2508.12398}, 2025.

\bibitem{lyu2023fine}
Y.~Lyu, T.~Luo, J.~Shi, T.~C. Hollon, and H.~Lee, ``Fine-grained text style transfer with diffusion-based language models,'' \emph{arXiv preprint arXiv:2305.19512}, 2023.

\bibitem{zhang2025diffusion}
S.~Zhang, Y.~Zhao, L.~Geng, A.~Cohan, A.~T. Luu, and C.~Zhao, ``Diffusion vs. autoregressive language models: A text embedding perspective,'' \emph{arXiv preprint arXiv:2505.15045}, 2025.

\bibitem{zhu2024segment}
X.~Zhu, G.~Karadzhov, C.~Whitehouse, and A.~Vlachos, ``Segment-level diffusion: A framework for controllable long-form generation with diffusion language models,'' \emph{arXiv preprint arXiv:2412.11333}, 2024.

\bibitem{cheng2024diffuspoll}
L.~Cheng and S.~Li, ``Diffuspoll: Conditional text diffusion model for poll generation,'' in \emph{Findings of the Association for Computational Linguistics ACL 2024}, 2024, pp. 925--935.

\bibitem{hu2024poetrydiffusion}
Z.~Hu, C.~Liu, Y.~Feng, A.~T. Luu, and B.~Hooi, ``Poetrydiffusion: Towards joint semantic and metrical manipulation in poetry generation,'' in \emph{Proceedings of the AAAI Conference on Artificial Intelligence}, vol.~38, no.~16, 2024, pp. 18\,279--18\,288.

\bibitem{padole2025improving}
T.~K. Padole, S.~P. Awate, and P.~Bhattacharyya, ``Improving text style transfer using masked diffusion language models with inference-time scaling,'' \emph{arXiv preprint arXiv:2508.10995}, 2025.

\bibitem{lee2025editext}
C.~H. Lee, H.~Kim, J.~Yeom, and S.~Yoon, ``Editext: Controllable coarse-to-fine text editing with diffusion language models,'' \emph{arXiv preprint arXiv:2502.19765}, 2025.

\bibitem{do2025discrete}
D.~A. Do, L.~A. Tuan, W.~Buntine \emph{et~al.}, ``Discrete diffusion language model for efficient text summarization,'' in \emph{Findings of the Association for Computational Linguistics: NAACL 2025}, 2025, pp. 6278--6290.

\bibitem{shao2025diffetm}
W.~Shao, M.~Liu, and L.~Song, ``Diffetm: Diffusion process enhanced embedded topic model,'' \emph{arXiv preprint arXiv:2501.00862}, 2025.

\bibitem{dong2025termdiffusum}
X.~Dong, W.~Li, Y.~Le, Z.~Jiang, J.~Zhong, and Z.~Wang, ``Termdiffusum: A term-guided diffusion model for extractive summarization of legal documents,'' in \emph{Proceedings of the 31st International Conference on Computational Linguistics}, 2025, pp. 3222--3235.

\bibitem{xin2025cdaˆ2}
D.~Xin, K.~Zhao, J.~Sun, and Y.~Li, ``Cdaˆ2: Counterfactual diffusion augmentation for cross-domain adaptation in low-resource sentiment analysis,'' in \emph{Proceedings of the 31st International Conference on Computational Linguistics}, 2025, pp. 61--72.

\bibitem{chen2024effective}
Z.~Chen, L.~Wang, Y.~Wu, X.~Liao, Y.~Tian, and J.~Zhong, ``An effective deployment of diffusion lm for data augmentation in low-resource sentiment classification,'' \emph{arXiv preprint arXiv:2409.03203}, 2024.

\bibitem{zhu2024pinpointing}
L.~Zhu, X.~Chen, X.~Guo, C.~Zhang, Z.~Zhu, Z.~Zhou, and X.~Kong, ``Pinpointing diffusion grid noise to enhance aspect sentiment quad prediction,'' in \emph{Findings of the Association for Computational Linguistics ACL 2024}, 2024, pp. 3717--3726.

\bibitem{iwai2024layout}
S.~Iwai, A.~Osanai, S.~Kitada, and S.~Omachi, ``Layout-corrector: Alleviating layout sticking phenomenon in discrete diffusion model,'' in \emph{European Conference on Computer Vision}.\hskip 1em plus 0.5em minus 0.4em\relax Springer, 2024, pp. 92--110.

\bibitem{ye2024diffusion}
J.~Ye, S.~Gong, L.~Chen, L.~Zheng, J.~Gao, H.~Shi, C.~Wu, X.~Jiang, Z.~Li, W.~Bi \emph{et~al.}, ``Diffusion of thoughts: Chain-of-thought reasoning in diffusion language models,'' \emph{arXiv preprint arXiv:2402.07754}, 2024.

\bibitem{gao2024diffucomet}
S.~Gao, M.~Ismayilzada, M.~Zhao, H.~Wakaki, Y.~Mitsufuji, and A.~Bosselut, ``Diffucomet: Contextual commonsense knowledge diffusion,'' \emph{arXiv preprint arXiv:2402.17011}, 2024.

\bibitem{cao2025dpcl}
Y.~Cao, L.~Wang, and L.~Huang, ``Dpcl-diff: Temporal knowledge graph reasoning based on graph node diffusion model with dual-domain periodic contrastive learning,'' in \emph{Proceedings of the AAAI Conference on Artificial Intelligence}, vol.~39, no.~14, 2025, pp. 14\,806--14\,814.

\bibitem{zhao2025d1}
S.~Zhao, D.~Gupta, Q.~Zheng, and A.~Grover, ``d1: Scaling reasoning in diffusion large language models via reinforcement learning,'' \emph{arXiv preprint arXiv:2504.12216}, 2025.

\bibitem{van2025neurosymbolic}
E.~van Krieken, P.~Minervini, E.~Ponti, and A.~Vergari, ``Neurosymbolic diffusion models,'' \emph{arXiv preprint arXiv:2505.13138}, 2025.

\bibitem{shaahid2025underwater}
A.~Shaahid and M.~Behzad, ``Underwater diffusion attention network with contrastive language-image joint learning for underwater image enhancement,'' \emph{arXiv preprint arXiv:2505.19895}, 2025.

\bibitem{chi2024m2d2m}
S.~Chi, H.-g. Chi, H.~Ma, N.~Agarwal, F.~Siddiqui, K.~Ramani, and K.~Lee, ``M2d2m: Multi-motion generation from text with discrete diffusion models,'' in \emph{European Conference on Computer Vision}.\hskip 1em plus 0.5em minus 0.4em\relax Springer, 2024, pp. 18--36.

\bibitem{sun2025ar}
M.~Sun, W.~Wang, G.~Li, J.~Liu, J.~Sun, W.~Feng, S.~Lao, S.~Zhou, Q.~He, and J.~Liu, ``Ar-diffusion: Asynchronous video generation with auto-regressive diffusion,'' in \emph{Proceedings of the Computer Vision and Pattern Recognition Conference}, 2025, pp. 7364--7373.

\bibitem{jiang2025diffvla}
A.~Jiang, Y.~Gao, Z.~Sun, Y.~Wang, J.~Wang, J.~Chai, Q.~Cao, Y.~Heng, H.~Jiang, Z.~Zhang \emph{et~al.}, ``Diffvla: Vision-language guided diffusion planning for autonomous driving,'' \emph{arXiv preprint arXiv:2505.19381}, 2025.

\bibitem{liang2025discrete}
Z.~Liang, Y.~Li, T.~Yang, C.~Wu, S.~Mao, L.~Pei, X.~Yang, J.~Pang, Y.~Mu, and P.~Luo, ``Discrete diffusion vla: Bringing discrete diffusion to action decoding in vision-language-action policies,'' \emph{arXiv preprint arXiv:2508.20072}, 2025.

\bibitem{cui2025vilad}
C.~Cui, Y.~Zhou, J.~Peng, S.-Y. Park, Z.~Yang, P.~Sankaranarayanan, J.~Zhang, R.~Zhang, and Z.~Wang, ``Vilad: A large vision language diffusion framework for end-to-end autonomous driving,'' \emph{arXiv preprint arXiv:2508.12603}, 2025.

\bibitem{he2024learning}
H.~He, C.~Bai, L.~Pan, W.~Zhang, B.~Zhao, and X.~Li, ``Learning an actionable discrete diffusion policy via large-scale actionless video pre-training,'' \emph{arXiv preprint arXiv:2402.14407}, 2024.

\bibitem{liang2025discreteGuided}
J.~Liang, S.~Koenig, and F.~Fioretto, ``Discrete-guided diffusion for scalable and safe multi-robot motion planning,'' \emph{arXiv preprint arXiv:2508.20095}, 2025.

\bibitem{wang2025learning}
Z.~Wang, J.~Shi, N.~Heess, A.~Gretton, and M.~K. Titsias, ``Learning-order autoregressive models with application to molecular graph generation,'' \emph{arXiv preprint arXiv:2503.05979}, 2025.

\bibitem{li2025redisc}
Y.~Li, Y.~Lu, Z.~Wang, Z.~Wei, Y.~Li, and B.~Ding, ``Redisc: A reparameterized masked diffusion model for scalable node classification with structured predictions,'' \emph{arXiv preprint arXiv:2507.14484}, 2025.

\bibitem{jung2025scaffolddiffusionsparsemulticategory}
J.~Jung, ``Scaffold diffusion: Sparse multi-category voxel structure generation with discrete diffusion,'' 2025.

\bibitem{zhuang2025moleditrl}
Y.~Zhuang, D.~Shen, and Y.~Sun, ``Moleditrl: Structure-preserving molecular editing via discrete diffusion and reinforcement learning,'' \emph{arXiv preprint arXiv:2505.20131}, 2025.

\bibitem{yin2025cfp}
J.~Yin, C.~Zha, W.~He, C.~Xu, and X.~Gao, ``Cfp-gen: Combinatorial functional protein generation via diffusion language models,'' \emph{arXiv preprint arXiv:2505.22869}, 2025.

\bibitem{xiong2024text}
Y.~Xiong, K.~Li, W.~Liu, J.~Wu, B.~Du, S.~Pan, and W.~Hu, ``Text-guided multi-property molecular optimization with a diffusion language model,'' \emph{arXiv preprint arXiv:2410.13597}, 2024.

\bibitem{lee2025genmol}
S.~Lee, K.~Kreis, S.~P. Veccham, M.~Liu, D.~Reidenbach, Y.~Peng, S.~Paliwal, W.~Nie, and A.~Vahdat, ``Genmol: A drug discovery generalist with discrete diffusion,'' \emph{arXiv preprint arXiv:2501.06158}, 2025.

\bibitem{wang2024dplm}
X.~Wang, Z.~Zheng, F.~Ye, D.~Xue, S.~Huang, and Q.~Gu, ``Dplm-2: A multimodal diffusion protein language model,'' \emph{arXiv preprint arXiv:2410.13782}, 2024.

\bibitem{tang2025peptune}
S.~Tang, Y.~Zhang, and P.~Chatterjee, ``Peptune: De novo generation of therapeutic peptides with multi-objective-guided discrete diffusion,'' \emph{ArXiv}, pp. arXiv--2412, 2025.

\bibitem{zhang2025cross}
Y.~Zhang, Y.~Wang, K.~V. Nguyen, and P.~Hong, ``Cross-modality controlled molecule generation with diffusion language model,'' \emph{arXiv preprint arXiv:2508.14748}, 2025.

\bibitem{devlin2019bert}
J.~Devlin, M.-W. Chang, K.~Lee, and K.~Toutanova, ``Bert: Pre-training of deep bidirectional transformers for language understanding,'' in \emph{Proceedings of the 2019 conference of the North American chapter of the association for computational linguistics: human language technologies, volume 1 (long and short papers)}, 2019, pp. 4171--4186.

\bibitem{lewis2019bart}
M.~Lewis, Y.~Liu, N.~Goyal, M.~Ghazvininejad, A.~Mohamed, O.~Levy, V.~Stoyanov, and L.~Zettlemoyer, ``Bart: Denoising sequence-to-sequence pre-training for natural language generation, translation, and comprehension,'' \emph{arXiv preprint arXiv:1910.13461}, 2019.

\bibitem{pmlr-v235-bachmann24a}
G.~Bachmann and V.~Nagarajan, ``The pitfalls of next-token prediction,'' in \emph{Proceedings of the 41st International Conference on Machine Learning}, ser. Proceedings of Machine Learning Research, R.~Salakhutdinov, Z.~Kolter, K.~Heller, A.~Weller, N.~Oliver, J.~Scarlett, and F.~Berkenkamp, Eds., vol. 235.\hskip 1em plus 0.5em minus 0.4em\relax PMLR, 21--27 Jul 2024, pp. 2296--2318.

\bibitem{song2020score}
Y.~Song, J.~Sohl-Dickstein, D.~P. Kingma, A.~Kumar, S.~Ermon, and B.~Poole, ``Score-based generative modeling through stochastic differential equations,'' \emph{arXiv preprint arXiv:2011.13456}, 2020.

\bibitem{kaufmann2023survey}
T.~Kaufmann, P.~Weng, V.~Bengs, and E.~H{\"u}llermeier, ``A survey of reinforcement learning from human feedback,'' \emph{arXiv preprint arXiv:2312.14925}, vol.~10, 2023.

\bibitem{radford2019language}
A.~Radford, J.~Wu, R.~Child, D.~Luan, D.~Amodei, I.~Sutskever \emph{et~al.}, ``Language models are unsupervised multitask learners,'' \emph{OpenAI blog}, vol.~1, no.~8, p.~9, 2019.

\bibitem{touvron2023llama}
H.~Touvron, T.~Lavril, G.~Izacard, X.~Martinet, M.-A. Lachaux, T.~Lacroix, B.~Rozi{\`e}re, N.~Goyal, E.~Hambro, F.~Azhar \emph{et~al.}, ``Llama: Open and efficient foundation language models,'' \emph{arXiv preprint arXiv:2302.13971}, 2023.

\bibitem{grattafiori2024llama}
A.~Grattafiori, A.~Dubey, A.~Jauhri, A.~Pandey, A.~Kadian, A.~Al-Dahle, A.~Letman, A.~Mathur, A.~Schelten, A.~Vaughan \emph{et~al.}, ``The llama 3 herd of models,'' \emph{arXiv preprint arXiv:2407.21783}, 2024.

\bibitem{qwen2025qwen25technicalreport}
Qwen, :, A.~Yang, B.~Yang, B.~Zhang, B.~Hui, B.~Zheng, B.~Yu, C.~Li, D.~Liu, F.~Huang, H.~Wei, H.~Lin, J.~Yang, J.~Tu, J.~Zhang, J.~Yang, J.~Yang, J.~Zhou, J.~Lin, K.~Dang, K.~Lu, K.~Bao, K.~Yang, L.~Yu, M.~Li, M.~Xue, P.~Zhang, Q.~Zhu, R.~Men, R.~Lin, T.~Li, T.~Tang, T.~Xia, X.~Ren, X.~Ren, Y.~Fan, Y.~Su, Y.~Zhang, Y.~Wan, Y.~Liu, Z.~Cui, Z.~Zhang, and Z.~Qiu, ``Qwen2.5 technical report,'' 2025.

\bibitem{hui2024qwen25codertechnicalreport}
B.~Hui, J.~Yang, Z.~Cui, J.~Yang, D.~Liu, L.~Zhang, T.~Liu, J.~Zhang, B.~Yu, K.~Lu, K.~Dang, Y.~Fan, Y.~Zhang, A.~Yang, R.~Men, F.~Huang, B.~Zheng, Y.~Miao, S.~Quan, Y.~Feng, X.~Ren, X.~Ren, J.~Zhou, and J.~Lin, ``Qwen2.5-coder technical report,'' 2024.

\bibitem{Qwen-VL}
J.~Bai, S.~Bai, S.~Yang, S.~Wang, S.~Tan, P.~Wang, J.~Lin, C.~Zhou, and J.~Zhou, ``Qwen-vl: A versatile vision-language model for understanding, localization, text reading, and beyond,'' \emph{arXiv preprint arXiv:2308.12966}, 2023.

\bibitem{liu2023llava}
H.~Liu, C.~Li, Q.~Wu, and Y.~J. Lee, ``Visual instruction tuning,'' 2023.

\bibitem{liu2023improvedllava}
H.~Liu, C.~Li, Y.~Li, and Y.~J. Lee, ``Improved baselines with visual instruction tuning,'' 2023.

\bibitem{liu2024llavanext}
\BIBentryALTinterwordspacing
H.~Liu, C.~Li, Y.~Li, B.~Li, Y.~Zhang, S.~Shen, and Y.~J. Lee, ``Llava-next: Improved reasoning, ocr, and world knowledge,'' January 2024. [Online]. Available: \url{https://llava-vl.github.io/blog/2024-01-30-llava-next/}
\BIBentrySTDinterwordspacing

\bibitem{zhai2023sigmoid}
X.~Zhai, B.~Mustafa, A.~Kolesnikov, and L.~Beyer, ``Sigmoid loss for language image pre-training,'' in \emph{Proceedings of the IEEE/CVF international conference on computer vision}, 2023, pp. 11\,975--11\,986.

\bibitem{shaul2024flow}
N.~Shaul, I.~Gat, M.~Havasi, D.~Severo, A.~Sriram, P.~Holderrieth, B.~Karrer, Y.~Lipman, and R.~T. Chen, ``Flow matching with general discrete paths: A kinetic-optimal perspective,'' \emph{arXiv preprint arXiv:2412.03487}, 2024.

\bibitem{wu2025janus}
C.~Wu, X.~Chen, Z.~Wu, Y.~Ma, X.~Liu, Z.~Pan, W.~Liu, Z.~Xie, X.~Yu, C.~Ruan \emph{et~al.}, ``Janus: Decoupling visual encoding for unified multimodal understanding and generation,'' in \emph{Proceedings of the Computer Vision and Pattern Recognition Conference}, 2025, pp. 12\,966--12\,977.

\bibitem{sun2024autoregressive}
P.~Sun, Y.~Jiang, S.~Chen, S.~Zhang, B.~Peng, P.~Luo, and Z.~Yuan, ``Autoregressive model beats diffusion: Llama for scalable image generation,'' \emph{arXiv preprint arXiv:2406.06525}, 2024.

\bibitem{flux2024}
B.~F. Labs, ``Flux,'' \url{https://github.com/black-forest-labs/flux}, 2024.

\bibitem{bai2024meissonic}
J.~Bai, T.~Ye, W.~Chow, E.~Song, Q.-G. Chen, X.~Li, Z.~Dong, L.~Zhu, and S.~Yan, ``Meissonic: Revitalizing masked generative transformers for efficient high-resolution text-to-image synthesis,'' in \emph{The Thirteenth International Conference on Learning Representations}, 2024.

\bibitem{van2017neural}
A.~Van Den~Oord, O.~Vinyals \emph{et~al.}, ``Neural discrete representation learning,'' \emph{Advances in neural information processing systems}, vol.~30, 2017.

\bibitem{radford2021learning}
A.~Radford, J.~W. Kim, C.~Hallacy, A.~Ramesh, G.~Goh, S.~Agarwal, G.~Sastry, A.~Askell, P.~Mishkin, J.~Clark \emph{et~al.}, ``Learning transferable visual models from natural language supervision,'' in \emph{International conference on machine learning}.\hskip 1em plus 0.5em minus 0.4em\relax PmLR, 2021, pp. 8748--8763.

\bibitem{zhang2025cosineschedulefisherraooptimalmasked}
L.~Zhang, ``The cosine schedule is fisher-rao-optimal for masked discrete diffusion models,'' 2025.

\bibitem{he2025mdpoovercomingtraininginferencedivide}
H.~He, K.~Renz, Y.~Cao, and A.~Geiger, ``Mdpo: Overcoming the training-inference divide of masked diffusion language models,'' 2025.

\bibitem{schiff2025simple}
Y.~Schiff, S.~S. Sahoo, H.~Phung, G.~Wang, S.~Boshar, H.~Dalla-torre, B.~P. de~Almeida, A.~M. Rush, T.~PIERROT, and V.~Kuleshov, ``Simple guidance mechanisms for discrete diffusion models,'' in \emph{The Thirteenth International Conference on Learning Representations}, 2025.

\bibitem{nisonoff2025unlocking}
H.~Nisonoff, J.~Xiong, S.~Allenspach, and J.~Listgarten, ``Unlocking guidance for discrete state-space diffusion and flow models,'' in \emph{The Thirteenth International Conference on Learning Representations}, 2025.

\bibitem{rojas2025theoryinformedimprovementsclassifierfreeguidance}
K.~Rojas, Y.~He, C.-H. Lai, Y.~Takida, Y.~Mitsufuji, and M.~Tao, ``Theory-informed improvements to classifier-free guidance for discrete diffusion models,'' 2025.

\bibitem{LocalLLaMA_dynamically_scaled_RoPE_2023}
{u/emozilla (Reddit user)}, ``{Dynamically Scaled RoPE further increases performance of long context LLaMA with zero fine-tuning},'' 2023, post on r/LocalLLaMA, Reddit. Available at: \url{https://www.reddit.com/r/LocalLLaMA/comments/14mrgpr/dynamically_scaled_rope_further_increases/}.

\bibitem{salimans2022progressive}
T.~Salimans and J.~Ho, ``Progressive distillation for fast sampling of diffusion models,'' \emph{arXiv preprint arXiv:2202.00512}, 2022.

\bibitem{watson2021learning}
D.~Watson, J.~Ho, M.~Norouzi, and W.~Chan, ``Learning to efficiently sample from diffusion probabilistic models,'' \emph{arXiv preprint arXiv:2106.03802}, 2021.

\bibitem{dettmers2022gpt3}
T.~Dettmers, M.~Lewis, Y.~Belkada, and L.~Zettlemoyer, ``Gpt3. int8 (): 8-bit matrix multiplication for transformers at scale,'' \emph{Advances in neural information processing systems}, vol.~35, pp. 30\,318--30\,332, 2022.

\bibitem{zhang2024extracting}
C.~Zhang, J.~X. Morris, and V.~Shmatikov, ``Extracting prompts by inverting llm outputs,'' \emph{arXiv preprint arXiv:2405.15012}, 2024.

\bibitem{chen2024text}
Y.~Chen, H.~Lent, and J.~Bjerva, ``Text embedding inversion security for multilingual language models,'' \emph{arXiv preprint arXiv:2401.12192}, 2024.

\bibitem{morris2023language}
J.~X. Morris, W.~Zhao, J.~T. Chiu, V.~Shmatikov, and A.~M. Rush, ``Language model inversion,'' \emph{arXiv preprint arXiv:2311.13647}, 2023.

\bibitem{li2025vid}
Q.~Li, R.~Yu, and X.~Wang, ``Vid-sme: Membership inference attacks against large video understanding models,'' \emph{arXiv preprint arXiv:2506.03179}, 2025.

\bibitem{li2024membership}
Z.~Li, Y.~Wu, Y.~Chen, F.~Tonin, E.~Abad~Rocamora, and V.~Cevher, ``Membership inference attacks against large vision-language models,'' \emph{Advances in Neural Information Processing Systems}, vol.~37, pp. 98\,645--98\,674, 2024.

\bibitem{he2025towards}
Y.~He, B.~Li, L.~Liu, Z.~Ba, W.~Dong, Y.~Li, Z.~Qin, K.~Ren, and C.~Chen, ``Towards label-only membership inference attack against pre-trained large language models,'' in \emph{USENIX Security}, 2025.

\bibitem{zhang2024min}
J.~Zhang, J.~Sun, E.~Yeats, Y.~Ouyang, M.~Kuo, J.~Zhang, H.~F. Yang, and H.~Li, ``Min-k\%++: Improved baseline for detecting pre-training data from large language models,'' \emph{arXiv preprint arXiv:2404.02936}, 2024.

\bibitem{wang2025towards}
C.-L. Wang, Q.~Li, Z.~Xiang, Y.~Cao, and D.~Wang, ``Towards lifecycle unlearning commitment management: Measuring sample-level unlearning completeness,'' \emph{arXiv preprint arXiv:2506.06112}, 2025.

\bibitem{carlini2023extracting}
N.~Carlini, J.~Hayes, M.~Nasr, M.~Jagielski, V.~Sehwag, F.~Tramer, B.~Balle, D.~Ippolito, and E.~Wallace, ``Extracting training data from diffusion models,'' in \emph{32nd USENIX Security Symposium (USENIX Security 23)}, 2023, pp. 5253--5270.

\bibitem{qu2023unsafe}
Y.~Qu, X.~Shen, X.~He, M.~Backes, S.~Zannettou, and Y.~Zhang, ``Unsafe diffusion: On the generation of unsafe images and hateful memes from text-to-image models,'' in \emph{Proceedings of the 2023 ACM SIGSAC conference on computer and communications security}, 2023, pp. 3403--3417.

\bibitem{zhang2024generate}
Y.~Zhang, J.~Jia, X.~Chen, A.~Chen, Y.~Zhang, J.~Liu, K.~Ding, and S.~Liu, ``To generate or not? safety-driven unlearned diffusion models are still easy to generate unsafe images... for now,'' in \emph{European Conference on Computer Vision}.\hskip 1em plus 0.5em minus 0.4em\relax Springer, 2024, pp. 385--403.

\bibitem{shen2024anything}
X.~Shen, Z.~Chen, M.~Backes, Y.~Shen, and Y.~Zhang, ``"do anything now": Characterizing and evaluating in-the-wild jailbreak prompts on large language models,'' in \emph{Proceedings of the 2024 on ACM SIGSAC Conference on Computer and Communications Security}, 2024, pp. 1671--1685.

\bibitem{kaplan2020scaling}
J.~Kaplan, S.~McCandlish, T.~Henighan, T.~B. Brown, B.~Chess, R.~Child, S.~Gray, A.~Radford, J.~Wu, and D.~Amodei, ``Scaling laws for neural language models,'' \emph{arXiv preprint arXiv:2001.08361}, 2020.

\bibitem{bahri2024explaining}
Y.~Bahri, E.~Dyer, J.~Kaplan, J.~Lee, and U.~Sharma, ``Explaining neural scaling laws,'' \emph{Proceedings of the National Academy of Sciences}, vol. 121, no.~27, p. e2311878121, 2024.

\bibitem{floridi2020gpt}
L.~Floridi and M.~Chiriatti, ``Gpt-3: Its nature, scope, limits, and consequences,'' \emph{Minds and Machines}, vol.~30, pp. 681--694, 2020.

\bibitem{ouyang2022training}
L.~Ouyang, J.~Wu, X.~Jiang, D.~Almeida, C.~Wainwright, P.~Mishkin, C.~Zhang, S.~Agarwal, K.~Slama, A.~Ray \emph{et~al.}, ``Training language models to follow instructions with human feedback,'' \emph{Advances in neural information processing systems}, vol.~35, pp. 27\,730--27\,744, 2022.

\bibitem{bai2023qwen}
J.~Bai, S.~Bai, Y.~Chu, Z.~Cui, K.~Dang, X.~Deng, Y.~Fan, W.~Ge, Y.~Han, F.~Huang \emph{et~al.}, ``Qwen technical report,'' \emph{arXiv preprint arXiv:2309.16609}, 2023.

\bibitem{yang2025qwen3}
A.~Yang, A.~Li, B.~Yang, B.~Zhang, B.~Hui, B.~Zheng, B.~Yu, C.~Gao, C.~Huang, C.~Lv \emph{et~al.}, ``Qwen3 technical report,'' \emph{arXiv preprint arXiv:2505.09388}, 2025.

\bibitem{dao2022flashattention}
T.~Dao, D.~Fu, S.~Ermon, A.~Rudra, and C.~R{\'e}, ``Flashattention: Fast and memory-efficient exact attention with io-awareness,'' \emph{Advances in neural information processing systems}, vol.~35, pp. 16\,344--16\,359, 2022.

\bibitem{abadi2016deep}
M.~Abadi, A.~Chu, I.~Goodfellow, H.~B. McMahan, I.~Mironov, K.~Talwar, and L.~Zhang, ``Deep learning with differential privacy,'' in \emph{Proceedings of the 2016 ACM SIGSAC conference on computer and communications security}, 2016, pp. 308--318.

\bibitem{santos2022avoiding}
C.~F. G.~D. Santos and J.~P. Papa, ``Avoiding overfitting: A survey on regularization methods for convolutional neural networks,'' \emph{ACM Computing Surveys (Csur)}, vol.~54, no. 10s, pp. 1--25, 2022.

\bibitem{ying2019overview}
X.~Ying, ``An overview of overfitting and its solutions,'' in \emph{Journal of physics: Conference series}, vol. 1168.\hskip 1em plus 0.5em minus 0.4em\relax IOP Publishing, 2019, p. 022022.

\bibitem{chen2024data}
D.~Chen, Y.~Huang, Z.~Ma, H.~Chen, X.~Pan, C.~Ge, D.~Gao, Y.~Xie, Z.~Liu, J.~Gao \emph{et~al.}, ``Data-juicer: A one-stop data processing system for large language models,'' in \emph{Companion of the 2024 International Conference on Management of Data}, 2024, pp. 120--134.

\bibitem{li2024superfiltering}
M.~Li, Y.~Zhang, S.~He, Z.~Li, H.~Zhao, J.~Wang, N.~Cheng, and T.~Zhou, ``Superfiltering: Weak-to-strong data filtering for fast instruction-tuning,'' \emph{arXiv preprint arXiv:2402.00530}, 2024.

\end{thebibliography}
\endgroup

\clearpage

\begin{center}
    \huge \textbf{- \textit{Appendix} -}
\end{center}

\appendices
\renewcommand\thesection{Apdx.\Alph{section}}
\renewcommand\thesubsection{\thesection.\Roman{subsection}}
\renewcommand\thesubsubsection{\thesection.\Roman{subsection}.\arabic{subsubsection}}
\renewcommand\thesubsubsubsection{\thesection.\Roman{subsection}.\arabic{subsubsection}.(\alph{subsubsubsection})}

In this appendix, we provide additional details on the works and techniques introduced in the main text. Specifically, we present more comprehensive mathematical formulations for certain methods and include several related techniques not covered earlier. The organization of the appendix largely follows the structure of the main-text. Moreover, at the end of the appendix, we also discuss potential future research directions in dLLM and dMLLM.

\section{Mathematical Formulations}
\subsection{More Transition Matrices}
\begin{itemize}
    \item \textbf{Hybrid}: 
    Hybrid transition was initially discussed in~\cite{austin2021structured}, where multiple types of transition are combined to create more expressive diffusion processes. For example, the Routlette Diffusion in \cite{haxholli2025efficient} and the Generalized Interpolating Discrete Diffusion (GIDD)~\cite{rutte2025generalized}
    both study the linear combination of the absorbing transition and the uniform transition. A key motivation for such ``Linear + Absorbing'' transitions is to overcome a fundamental limitation of absorbing diffusion: once a token is denoised, it cannot be revised. Such ``Linear + Absorbing'' transitions address this by allowing previously denoised tokens to re-enter the diffusion process via the uniform component. This grants the model the ability to correct earlier generation errors during inference.
    \item \textbf{Discretized Gaussian}:
    \[
    [Q_t]_{ij} = 
    \begin{cases}
    \frac{\exp\left(-\frac{4|i-j|^2}{(K-1)^2\beta_t}\right)}{\sum\limits_{n=-(K-1)}^{K-1} \exp\left(-\frac{4n^2}{(K-1)^2\beta_t}\right)} & \text{if } i \ne j \\
    1 - \sum\limits_{l \ne i} [Q_t]_{il} & \text{if } i = j
    \end{cases}
    \]
    This matrix simulates Gaussian diffusion, suitable for ordinal data. Each state \( i \) is most likely to transition to nearby states \( j \) with probabilities resembling a Gaussian kernel. Closer states receive higher probabilities, while distant states receive lower ones. Diagonal values are chosen to ensure that each row sums to 1, yielding a uniform stationary distribution.
    \item \textbf{Band-diagonal}:
    \[
    [Q_t]_{ij} = 
    \begin{cases}
    \frac{1}{K \beta_t} & 0 < |i - j| \le v \\
    0 & |i - j| > v \\
    1 - \sum\limits_{l \ne i} [Q_t]_{il} & i = j
    \end{cases}
    \]
    Band-diagonal imposes local transitions only: state \( i \) can only transition to its \( v \)-nearest neighbors on the ordinal scale. It biases the forward process toward small, local perturbations.
    \item \textbf{Embedding-based}: 
    Let $A$ be an adjacent matrix build from based on the similarity between tokens in the embedding space, in \cite{austin2021structured} the Embedding-based transition matrix is defined as
    \begin{align}
    Q_t &= \exp(\alpha_t R) = \sum_{n=0}^{\infty} \frac{\alpha_t^n}{n!} R^n,\\
    R_{ij} &= 
    \begin{cases}
    - \sum_{l \ne i} A_{il} & i = j \\
    A_{ij} & i \ne j
    \end{cases}.
    \end{align}
    This transition matrix promotes a diffusion process where tokens are more likely to transition to others that are semantically or syntactically similar in the embedding space.
\end{itemize}

\subsection{Reparameterized Discrete Diffusion Model}
\label{math_2}
Reparameterized Discrete Diffusion Models (RDMs)~\cite{zheng2023reparameterized} reformulate the backward process of discrete diffusion in D3PM into a two-stage sampling procedure. The core idea is to introduce a latent routing variable $v_t$ that determines decoding behavior for each token at every step.

Given a forward transition of the form:
\begin{align}
q(x_t \mid x_{t-1}) &= \beta_t x_{t-1} + (1 - \beta_t) q_{\text{noise}},\\
q(x_t \mid x_0) &= \alpha_t x_{t-1} + (1 - \alpha_t) q_{\text{noise}},
\end{align}
where $q_{\text{noise}}$ is the noise distribution, and $\alpha_t := \prod_{i=1}^t \beta_i$.
The backward posterior $q(x_{t-1} \mid x_t, x_0)$ can be expressed as a mixture of two cases, depending on whether $x_t = x_0$:
\begin{equation}
q(x_{t-1} \mid x_t, x_0) =
\begin{cases}
\lambda^{(1)}_{t-1} x_t + (1 - \lambda^{(1)}_{t-1}) q_{\text{noise}}, & x_t = x_0, \\
\lambda^{(2)}_{t-1} x_0 + (1 - \lambda^{(2)}_{t-1}) q_{\text{noise}}(x_t), & x_t \neq x_0,
\end{cases}\label{eq:rdm:back}
\end{equation}
where $q_{\text{noise}}(x_t)$ is the interpolation between $x_t$ and $q_{\text{noise}}$, and $\lambda^{(1)}_{t-1}$, $\lambda^{(2)}_{t-1}$ are scalar coefficients derived from $\beta_t$, $\alpha_t$ and $q_{\text{noise}}(x_t)$.

To reparameterize the backward transition, RDM introduces Bernoulli latent variables:
\begin{align}
v_{t-1}^{(1)} &\sim \text{Bernoulli}(\lambda^{(1)}_{t-1}), \quad u^{(1)}_t \sim q_{\text{noise}}, \\
v_{t-1}^{(2)} &\sim \text{Bernoulli}(\lambda^{(2)}_{t-1}), \quad u^{(2)}_t \sim q_{\text{noise}}(x_t).
\end{align}
Let $b_t = \mathbf{1}[x_t = x_0]$, the reparameterized sample is computed as:
\begin{align}
x_{t-1} &= b_t \left[ v_{t-1}^{(1)} x_t + (1 - v_{t-1}^{(1)}) u^{(1)}_t \right] \nonumber \\
& \quad+ (1 - b_t) \left[ v_{t-1}^{(2)} x_0 + (1 - v_{t-1}^{(2)}) u^{(2)}_t \right].\label{rdm:sampling}
\end{align}
This formulation allows the model to explicitly route tokens through different behaviors: either retaining the current token, resetting to noise, or denoising back to the ground truth.

Based on \cref{rdm:sampling}, RDMs define a joint generative model:
\begin{equation}
p_\theta(x_0, x_{1:T}, v_{1:T}) = p_\theta(x_T) \prod_{t=1}^T p_\theta(x_{t-1}, v_{t-1} \mid x_t),
\end{equation}
and the evidence lower bound (ELBO) becomes:
\begin{equation}
\log p_\theta(x_0) \geq \mathcal{L}_{1} - \underbrace{\sum_{t=2}^T \mathcal{L}_{t}}_{\mathcal{L}_{T}} + C,
\end{equation}
where $C$ is a constant.

For $t > 1$, the loss term decomposes as:
\begin{align}
\mathcal{L}_t &= \mathbb{E}_{x_{1:T},v_{1:T}\mid x_0} \Big[ \mathrm{KL}(q(v_{t-1}) \parallel p_\theta(v_{t-1} \mid x_t))\nonumber +\nonumber\\&\mathrm{KL}(q(x_{t-1} \mid v_{t-1}, x_t, x_0) \parallel p_\theta(x_{t-1} \mid v_{t-1}, x_t)) \Big].
\end{align}

By aligning $p_\theta(v_{t-1} \mid x_t)$ with $q(v_{t-1})$ and using the $x_0$-parameterization, the loss can be simplified into
\begin{align}
\mathcal{L}_t &= \mathbb{E}_{x_0, x_{1:T}} \Big[ - \lambda_{t-1}^{(2)} \sum_{n=1}^N (1 - b_{t,n}) x_{0,n} \log [f_{\theta}(x_{t})]_n
\Big],\label{eq:rpm}
\end{align}
where $\lambda_{t-1}^{(2)} = \frac{\alpha_{t-1} - \alpha_t}{1 - \alpha_t}$.

A central issue of RDM lies in its dependence on the ground truth $x_0$ to compute the backward transition probabilities \cref{eq:rdm:back}. However, in the inference stage, $x_0$ is unknown, making it infeasible to directly evaluate the indicator function $b_t$ required for determining the appropriate transition path. 
To overcome this limitation, the authors propose a recursive approximation for computing $b_t$ by utilizing the Bernoulli routing variables $v$. Beginning with $b_T = 0$, which assumes the initial sequence is fully noisy, the clean token set is recursively updated via:
\begin{equation}
b_{t-1,n} = \left( b_{t,n} \land v_{t-1,n}^{(1)} \right) \lor v_{t-1,n}^{(2)},
\end{equation}
where $\land$ and $\lor$ denote logical conjunction and disjunction, respectively. 

\subsection{Concrete Score}

\subsubsection{Training Loss}
Training is typically done by minimizing divergence-based losses that compare the predicted ratio to the true ratio.
\subsubsubsection{Concrete Score Matching (CSM)~\cite{concrete_score}} 
The most direct approach is Concrete Score Matching (CSM)~\cite{concrete_score}, which uses a squared-error loss to match the predicted and true ratios:
\begin{equation}
\mathcal{L}_{\text{CSM}}(t) = \frac{1}{2} \, \mathbb{E}_{x \sim p_t} \left[ \sum_{x' \ne x} \left( s_\theta(x, t)_{x'} - \frac{p_t(x')}{p_t(x)} \right)^2 \right]
\end{equation}
While theoretically sound, this $\ell_2$ loss does not penalize invalid (e.g., negative or zero) predictions sufficiently, potentially leading to instability.

\subsubsubsection{Diffusion-Weighted Denoising Score Entropy~\cite{lou2023discrete}} 
Leveraging Bregman divergence, score entropy is formulated as another score matching loss. Score entropy is non-negative, symmetric, and convex. It is also an extension of the conventional cross-entropy to general positive-valued functions beyond the probability simplex. Score entropy enables the construction of an ELBO for discrete diffusion models, resulting in the Diffusion-Weighted Denoising Score Entropy (DWDSE) loss~\cite{lou2023discrete}
\begin{align}
\mathcal{L}_{\text{DWDSE}}(x_0) &= \int_0^T \mathbb{E}_{x_t \sim p_{t|0}(\cdot|x_0)} \sum_{x' \ne x_t} R_t(x_t, x') \Bigg(
\nonumber\\ &\qquad s_\theta(x_t, t)_{x'} - \frac{p_{t|0}(x'|x_0)}{p_{t|0}(x_t|x_0)} \log s_\theta(x_t, t)_{x'} 
\nonumber\\ &\qquad + K\left( \frac{p_{t|0}(x'|x_0)}{p_{t|0}(x_t|x_0)} \right) \Bigg) \, dt,
\end{align}
where $K$ is a normalizing constant function.

\subsubsubsection{Target Concrete Score Matching~\cite{zhang2025target}} 
Target Concrete Score Matching (TCSM) introduces two score matching losses: the \textit{score-based} and \textit{distribution-based} losses. 
The score-based objective operates directly on the concrete score vectors:
\begin{align}
    &\mathcal{L}_{\mathrm{score}}(\theta) = \mathbb{E}_{\omega(t)p(x_t)h(x_1|x_t)} \sum_{i=1}^{L} \ell^i_{\text{score}},\\
    &\ell^i_{\text{score}} = \mathcal{D}\left(
    \left[ \frac{p_{1|t}([y^i, x_1^{\backslash i}]|x_t)}{p_{1|t}(x_1|x_t)} \right]_{y^i=1}^{V},
    \left[ \frac{p^\theta_{1|t}([y^{i}, x_1^{\backslash i}]|x_t)}{p_\theta^{1|t}(x_1|x_t)} \right]_{y^i=1}^{V}
    \right),
\end{align}
where $V$ is the vocabulary size, $\mathcal{D}(\cdot, \cdot)$ denotes a divergence measure, $\omega(t)$ is the distribution used to sample time index $t$, and $h(x_1|x_t)$ is a proposal distribution such as the ground-truth conditional distribution $p_{1|t}(x_1|x_t)$. In the equation, $[y^i, x^{\backslash i}] := [x^1, \ldots, x^{i-1}, y^i, x^{i+1}, \ldots, x^L]$ is used to define a new sequence of the same length as $x_1$, where the token at position $i$ is replaced by $y^i$ and all other tokens remain identical to those in $x_1$. $x^{\backslash i}_1$ is used to indicate all tokens in $x_1$ except for the token at position $i$.

The distribution-based objective aligns the model and true conditional distributions:
\begin{align}
    &\mathcal{L}_{\mathrm{distrib}}(\theta) = \mathbb{E}_{\omega(t)p(x_t)} \sum_{i=1}^{L} \mathbb{E}_{h(x_1^{\backslash i}|x_t)} \ell^i_{\text{distrib}} ,\\
    &\ell^i_{\text{distrib}} = \mathbb{D}\left(
    p_{1|t}(x_1^{i}|x_1^{\backslash i}, x_t) \, \| \, p^\theta_{1|t}(x_1^{i}|x_1^{\backslash i}, x_t)
    \right),
\end{align}
where $\mathbb{D}(\cdot, \cdot)$ is a statistical divergence measures the differences between probability distributions.
Shown in Proposition 3 of \cite{zhang2025target}, with $h(x_1|x_t) = p^{1|t}(x_1|x_t)$, the two objectives are equivalent:
\begin{equation}
    \mathcal{L}_{\mathrm{score}}(\theta; \mathcal{D} = \mathcal{D}_{\mathrm{GKL}}) \equiv \mathcal{L}_{\mathrm{distrib}}(\theta; \mathbb{D} = V\mathbb{D}_\mathrm{KL} + \mathbb{D}_\mathrm{IS}),
\end{equation}
where $\mathcal{D}_{\mathrm{GKL}}$, $\mathbb{D}_\mathrm{KL}$ and $\mathbb{D}_\mathrm{IS}$ refer to the generalized Kullback–Leibler divergence, the Kullback–Leibler divergence and Itakura–Saito divergence, respectively.

By selecting different discrepancy measures and proposal distributions, there are different instantiations of the TCSM loss. For instance, choosing the generalized KL divergence as the discrepancy $\mathcal{D}$ and the true conditional distribution $p_{1|t}(x_1|x_t)$ as the proposal $h(x_1|x_t)$, the score-based TCSM becomes
\begin{align}
    \ell_{\mathrm{score}}^i &= 
    \left(
        -\log p^{\theta}_{1|t}(x_1^i | x_t)
        + \frac{1}{V p^{\theta}_{1|t}(x_1^i | x_t)}
    \right)\nonumber\\
    &\qquad+ \frac{1}{V} \sum_y \log p^{\theta}_{1|t}(y | x_t);
\end{align}
choosing the KL divergence as the discrepancy $\mathbb{D}$ and the true conditional distribution $p_{1|t}(x_1|x_t)$ as the proposal $h(x_1|x_t)$, the distribution-based TCSM becomes
\begin{equation}
    \ell_{\mathrm{distrib}}^i = - \mathbb{E}_{p_{1|t}} \log p^{\theta}_{1|t}(x_1^i | x_t) + C,
\end{equation}
where $C$ is a constant. The distribution-based objective reduces to a cross-entropy loss, maximizing the pseudo-likelihood of $p^{\theta}_{1|t}$ under the true denoising distribution.

\subsubsection{Connection with CE Loss}
\cite{sun2023scorebased} leverages the theorem stating that two probability distributions are identical if and only if their conditional distributions are equal for all values of the condition. Based on this, the original marginal probability distributions in the concrete score are transformed into conditional probability distributions.
Since both the numerator and denominator of the concrete score share the same functional form, they can be represented by a single neural network. That is, the concrete score can be expressed as:
\begin{align}
    &p_t(X^d \mid x^{\setminus d}; \theta) \approx q_t(X^d \mid x^{\setminus d}) 
\Rightarrow \nonumber\\
&\frac{q_t(y^d, x^{\setminus d})}{q_t(x^d, x^{\setminus d})} 
= 
\frac{q_t(X_t^d = y^d \mid x^{\setminus d})}{q_t(X_t^d = x^d \mid x^{\setminus d})} 
\approx 
\frac{p_t(X_t^d = y^d \mid x^{\setminus d}; \theta)}{p_t(X_t^d = x^d \mid x^{\setminus d}; \theta)}.
\end{align}
From this perspective, \cite{sun2023scorebased} propose the categorical ratio matching loss, which is a cross-entropy loss to fit conditional probabilities. Thus, \cite{sun2023scorebased} shows that training a neural network with a cross-entropy loss is also able to fit the concrete score. 

\cite{haxholli2025efficient} also connects the concrete score matching with the CE loss. Consider two discrete sequences \( x \) and \( y \) that only differ at position \( i \).
By Bayesian Theorem, the probability ratio $\frac{p_t(y)}{p_t(x)}$ can be rewritten as
\begin{equation}
\frac{p_t(y)}{p_t(x)} = \sum_{h \in \mathcal{V}} \frac{p_{t|0}(y^i \mid h)}{p_{t|0}(x^i \mid h)} \cdot p^i_{0|t}(h \mid x_t),
\end{equation}
where \( p_{t|0}(\cdot \mid h) \) is the known conditional transition probability in the forward process, and \( p^i_{0|t}(h \mid x_t) \) is the posterior of the clean token \( h \) at position \( i \) given the noised sequence \( x_t \). Thus, the concrete score can be parameterized as 
\begin{equation}
s^i_\theta(x_t, t) = \sum_{h \in \mathcal{V}} \frac{p_{t|0}(y^i \mid h)}{p_{t|0}(x^i \mid h)} \cdot f^i_\theta(x_t, t)[h],
\end{equation}
where \( f^i_\theta(x_t, t)[h] \) models the likelihood that token \( h \) was the original clean token at position \( i \). This allows training to proceed via a simple cross-entropy loss over the posterior \( p_{0|t} \), rather than requiring explicit score estimation:
\begin{equation}
\mathcal{L}_{\text{CEDD}} = \mathbb{E}_{t} \mathbb{E}_{x_0,x_t} \sum_{i=1}^{L} w(t) \cdot \log f^i_\theta(x_t, t)[x^i_0],
\end{equation}
where \( w(t) \) is a timestep-dependent weighting function. 

\subsubsection{Time Independency}
An issue with the Concrete Score is its dependency on time, which prevents the use of caching techniques for inference and results in lower efficiency.
\cite{ou2025absorbingdiscretediffusionsecretly} shows that the concrete score in absorbing discrete diffusion can be reparameterized as a product of a time-dependent scalar and a conditional distribution over clean data. Specifically, in practice, $R_t$ can be parameterized as the multiplication between a scalar function and a constant rate, \textit{i.e.}, $R_t = \sigma(t)R$. 
let $x_i = [\mathrm{M}]$ denote a masked token at position $i$. Then, the concrete score for replacing $x_i$ with a token $x'_i \ne [\mathrm{M}]$ is defined as:
\begin{equation}
\frac{p_t(x')}{p_t(x)} = \frac{p_t(x_1, \dots, x'_i, \dots)}{p_t(x_1, \dots, x_i, \dots)} = \frac{e^{-\bar{\sigma}(t)}}{1 - e^{-\bar{\sigma}(t)}} \cdot p_0(x'_i \mid x^{\mathrm{UM}}),
\end{equation}
where:
\begin{itemize}
    \item $\bar{\sigma}(t) = \int_0^t \sigma(s) ds$ is the cumulative noise schedule,
    \item $x^{\mathrm{UM}}$ consists of all unmasked tokens of $x$,
    \item $p_0(x'_i \mid x^{\mathrm{UM}})$ is the conditional distribution of the clean data.
\end{itemize}
This reparameterization removes the dependence on $t$ from the model output, enabling the training of a time-independent network $c_\theta$:
\begin{equation}
c_\theta(x)[i, \hat{x}^{(i)}] \approx p_0(\hat{x}^{(i)} \mid x^{\mathrm{UM}}).
\end{equation}
Such a model, termed RADD (Reparameterized Absorbing Discrete Diffusion), permits caching of the network outputs across steps when tokens remain unchanged, reducing the number of function evaluations (NFEs) during inference.

\section{Modeling Language Diffusion}

\subsection{Block Diffusion Models}
Given a sequence $x = (x_1, \ldots, x_L)$, BD3-LMs partition it into $B$ blocks of length $L'$, denoted $x = (x^{(1)}, \ldots, x^{(B)})$. The joint likelihood under a Block Diffusion model is factorized autoregressively across blocks:
\begin{equation}
\log p_\theta(x) = \sum_{b=1}^{B} \log p_\theta(x^{(b)} \mid x^{(<b)}),
\end{equation}
where each conditional distribution $p_\theta(x^{(b)} \mid x^{(<b)})$ is modeled by a discrete diffusion process applied within block $b$:
\begin{align}
& p_\theta(x^{(b)}_s \mid x^{(b)}_t, x^{(<b)}) \nonumber\\ &\quad= \sum_{x^{(b)}} q(x^{(b)}_s \mid x^{(b)}_t, x^{(b)}) \, p_\theta(x^{(b)} \mid x^{(b)}_t, x^{(<b)}),
\end{align}
where $q(\cdot \mid \cdot)$ is the forward noise process, and $p_\theta(x^{(b)} \mid x^{(b)}_t, x^{(<b)})$ is the learned denoising model.

The model is parameterized by a transformer $f_\theta$ with a block-causal attention mask. For each block $x^{(b)}$, the model predicts:
\begin{equation}
f_\theta\left(x^{(b)}_t, x^{(<b)}\right) \rightarrow \hat{x}^{(b)}_0.
\end{equation}
During inference, block sampling proceeds sequentially over blocks, but parallel sampling is used within blocks. Block-wise KV caching can be used to avoid recomputing transformer states for previously generated blocks.

The training loss for the full sequence is obtained by summing the variational lower bound over all blocks:
\begin{equation}
- \log p_\theta(x) \leq \mathcal{L}_{\mathrm{BD}}(x; \theta) := \sum_{b=1}^{B} \mathcal{L}(x^{(b)}, x^{(<b)}; \theta),
\end{equation}
where each $\mathcal{L}(x^{(b)}, x^{(<b)}; \theta)$ follows the discrete diffusion loss, optionally adapted to continuous time or masking processes.

\subsection{Flexible-Length Masked Diffusion}
In a standard Masked Diffusion Models, each token can only exist in one of two states: \emph{masked} or \emph{unmasked}. In FlexMDM, each token may reside in one of three states: \textit{empty}, \textit{masked}, or \textit{ground truth (gt)}. In the forward process, a token first transitions from the \textit{ground truth} state to the \textit{masked} state, and finally to the \textit{empty} state. Let $x_1 = (x^1_1, \ldots, x^n_1) \sim p_1$ be a target sequence of length $n$. FlexMDM defines two smooth monotone schedules $\alpha, \beta : [0,1] \to [0,1]$ with boundary conditions $(\alpha_0,\beta_0)=(0,0)$ and $(\alpha_1,\beta_1)=(1,1)$. For each token $i$, FlexMDM independently sample an \emph{insertion time} $T^i_1$ and an \emph{unmasking time} $T^i_2$:
\[
T^i_1 \sim \dot{\alpha}_t \, dt, 
\quad
T^i_2 \sim \mathbf{1}\{t \geq T^i_1\}\,\frac{\dot{\beta}_t}{1 - \beta_{T^i_1}} \, dt,
\]
where $\dot{\alpha}_t$ and $\dot{\beta}_t$ are the derivatives of $\alpha_t$ and $\beta_t$.
The token state $x^i_t$ at time $t \in [0,1]$ evolves as:
\[
x^i_t = 
\begin{cases}
\emptyset, & 0 < t < T^i_1, \\
m, & T^i_1 \leq t < T^i_2, \\
x^i_1, & T^i_2 \leq t \leq 1,
\end{cases}
\]
where $m$ denotes the special mask symbol and $\emptyset$ represents token removal. The sequence $x_t$ is obtained by concatenating all non-empty $x^i_t$.

Thus, in FlexMDM, there are two complementary tasks:  
\begin{enumerate}
    \item \textbf{Unmasking Task.}  
    Given a partially masked sequence $x_t$, the model predicts the ground-truth token for each masked position. This is identical to the original MDM formulation. Specifically, a diffusion language model $f_\theta$ is trained to approximate the posterior distribution over clean tokens.
    \item \textbf{Insertion Task.}  
    Beyond unmasking, FlexMDM introduces the additional task of inserting tokens into the sequence. For this purpose, an auxiliary network $g_\theta$ is trained to predict the expected number of mask tokens that should be inserted before each existing token.
    During inference, this prediction is combined with a Poisson distribution to determine the actual number of insertions:
    \[
    k_i \sim \text{Poisson}\!\left(r\cdot g_\theta(x_t, t)[i]\right),
    \]
    where $r$ is a scaling factor.
    The sampled $k$ tokens are inserted as mask symbols $m$, thereby extending the sequence length.  
\end{enumerate}

At inference time, sequence generation alternates between the two tasks:
\begin{enumerate}
    \item \textbf{Insertion:} For each position $i$, sample $k_i \sim \text{Poisson}(r\cdot g_\theta(x_t, t)[i])$ and insert $k_i$ mask tokens into the sequence.
    \item \textbf{Unmasking:} Apply $f_\theta$ to predict clean tokens at masked positions and replace them accordingly.
\end{enumerate}
This insertion–unmasking cycle is repeated iteratively until the sequence converges to a fully unmasked form, thereby producing variable-length outputs while preserving the any-order property of diffusion-based inference.

\subsection{Diffusion with Optimal Transport Position Coupling}
At each timestep $t$, DDOT can output both (i) the predicted token value distribution and (ii) the velocity of token positions:
\begin{equation}
s_\theta(x_t, t) \quad \text{and} \quad v_\theta(z_t, t),
\end{equation}
where $x_t$ are token values and $z_t \in [-1,1]^L$ are continuous position variables. The extra velocity of token position is predicted by and additional linear head. The token positions are diffused in continuous space from an initial distribution $z_T \sim U(-1,1)^L$ to the ground-truth positions $z_0$. The token denoising objective follows the score entropy loss from SEDD~\cite{lou2023discrete}, while the position denoising is learned via a weighted mean-squared error loss:
\begin{equation}
\mathcal{L}_{\text{pos}}(\theta) = \mathbb{E}_{(z,t)} \Big[ Q_t(x_t,y)\, \| v_\theta(z_t, t) - (z_0 - z_T) \|^2 \Big].
\end{equation}

\section{Training Techniques}
\subsection{Masking Scheduling Technique}
\subsubsection{Additional Uniform  Masking Scheduling strategies}
\begin{itemize}
    \item \textbf{Geometric Schedule~\cite{lou2023discrete,shi2024simplified}:}
    \begin{align}
        \alpha_t &= \exp\left(-\bar{\beta}_{\min}^{1 - t} \bar{\beta}_{\max}^t\right), \\
        w_t &= \left( \frac{\exp\left(-\bar{\beta}_{\min}^{1 - t} \bar{\beta}_{\max}^t\right)}{1 - \exp\left(-\bar{\beta}_{\min}^{1 - t} \bar{\beta}_{\max}^t\right)} \right) \bar{\beta}_{\min}^{1 - t} \bar{\beta}_{\max}^t \log\left(\frac{\bar{\beta}_{\min}}{\bar{\beta}_{\max}}\right).
    \end{align}

    \item \textbf{Polynomial Schedule~\cite{shi2024simplified}:}
    \begin{align}
        \alpha_t &= 1 - t^r, \\
        w_t &= -\frac{r}{t}.
    \end{align}
\end{itemize}
\subsubsection{Token-wise Masking Scheduling}
The Spindle-shaped Noise Schedule defines a custom corruption probability $\alpha_t^i$ for each token position $i$ at timestep $t$, determined by:
\begin{align}
\alpha_t^i &= 1 - \frac{t}{T} - S(t) \cdot \tilde{H}(x_0^i), \label{eq:spindle-alpha} \\
S(t) &= \lambda \sin\left( \frac{t\pi}{T} \right), \label{eq:spindle-S} \\
\tilde{H}(x_0^i) &= 1 - \frac{\sum_{j=1}^n H(x_0^j)}{n \cdot H(x_0^i)}, \label{eq:spindle-H}
\end{align}
where $H(x_0^i)$ denotes the entropy of the $i$-th token, measuring its information content, $S(t)$ is a sinusoidal scaling function that ensures zero informativeness contribution at $t=0$ and $t=T$, $\lambda$ is a hyperparameter controlling the strength of entropy influence.

\subsection{Distillation through Dimensional Correlations}

The distillation loss conveys the probabilistic knowledge of a teacher model that performs multi-step denoising to a student model designed for fewer-step generation.  This is achieved by aligning the student’s predicted posterior with that of the teacher at an intermediate noise level $\delta$. Formally, the loss is defined as:
\begin{align}
\mathcal{L}_{\text{distil}}(\theta) &
= \mathbb{E}_{x_\delta \sim r_\delta} \Big[ 
    \mathrm{D_{KL}}\big( 
    p^{\psi}_{0|\delta}(\cdot \mid x_\delta) p^{\theta}_{0|\delta}(\cdot \mid x_\delta)
    \big) \Big]
\end{align}
where $p^{\psi}_{0|\delta}(\cdot | x_\delta)$ is the posterior distribution over clean data $x_0$ given intermediate noisy input $x_\delta$ under the teacher model; $p^{\theta}_{0|\delta}(\cdot | x_\delta)$ is the student model's predicted posterior; $r_\delta$ is a reference distribution over noisy states (typically chosen to match the forward diffusion at timestep $\delta$). This loss encourages to transfer the full-step generative knowledge from teacher to student by matching posteriors.

The consistency loss enforces that the student model produces stable predictions across varying intermediate noise levels. Specifically, if $x_t$ is a noisy sample at step $t$, there should be agreement between: 1). First denoising $x_t \to x_u$ via the teacher, then $x_u \to x_s$ via the student; 2). Directly predicting $x_t \to x_s$ via the student. This intuition is captured by the following KL divergence:
\begin{align}
\mathcal{L}_{\text{consis}}(\theta) ={} 
\mathbb{E}_{x_t \sim r_t} \Big[
\mathrm{D_{KL}}\bigl(
(p^{\theta}_{s|u} \circ p^{\psi}_{u|t})(\cdot \mid x_t) \|\, p^{\theta}_{s|t}(\cdot \mid x_t)
\bigr) \Big]
\end{align}
where $p^{\psi}_{u|t}(x_u | x_t)$ is the teacher’s distribution from timestep $t$ to $u$; $p^{\theta}_{s|u}(x_s | x_u)$ is the student’s distribution from $u$ to $s$; $p^{\theta}_{s|t}(x_s | x_t)$ is the direct prediction by the student; $(p^{\theta}_{s|u} \circ p^{\psi}_{u|t})(\cdot | x_t)$ denotes the composite distribution over $x_s$ via intermediate $x_u$. This loss enforces functional coherence across generation paths, capturing multi-dimensional correlations without assuming independence.

\subsection{Input Discrepancy Between the Training and Testing}
\label{training-test-dis}
\cite{asada-miwa-2025-addressing} mentioned  a discrepancy between training and inference of dLLM. During training, the model receives ground-truth noisy tokens as input, while at inference time, the inputs of the model are the previously predicted tokens. To address this, \cite{asada-miwa-2025-addressing} propose the two-step loss.

Let $x_0$ denote the ground truth sequence. During training, a time step $t$ is randomly selected, and a noised version $x_t$ is generated from $x_0$ via a diffusion process. The model then predicts the original sequence $\hat{x}_0 = f_{\theta}(x_t)$.
Subsequently, a second input $\hat{x}_{t-1}$ is generated by applying the forward diffusion transition matrixto the predicted sequence $\hat{x}_0$
and the model again attempts to recover the ground truth $\hat{\hat{x}}_0 = f_{\theta}(\hat{x}_{t-1})$.
The two-step loss is calculated between the ground truth $x_0$ and the twice-denoised output $\hat{\hat{x}}_0$. 

To ease training in early stages, the model does not always use this two-step strategy. Instead, a mixed strategy is adopted. With probability $1 - p_k$, the two-step strategy is used and the loss is evaluated between $\hat{\hat{x}}_0$ and $x_0$. With probability $p_k$, the conventional one-step strategy is used and the loss is evaluated between $\hat{x}_0$ and $x_0$. $p_k$ is set to be linearly increasing along the training step $k$.

\subsection{Reinforcement Learning}

\subsubsection{Diffusion-based GRPO}  
UniGRPO in \cite{yang2025mmada} adapts policy optimization to the diffusion setting by combining structured noising, likelihood approximation, and clipped policy gradients.  
Let $q$ and $\{o_i\}_{i=1}^G$ denote a query and a batch of responses, respectively. For each response $o_i$, UniGRPO samples a masking ratio $p_i \sim U[0,1]$ and create a perturbed version $\tilde{o}_{i,p}$ by masking tokens. The token-level likelihood and the sequence-level likelihood are approximated as 
\begin{align}
    \pi'_\theta(o_{i,t} \mid q, \tilde{o}_{i,p}, p_i) &= \mathbb{E}_{p_i}\big[\mathbf{1}[\tilde{o}_{i,t,p}=\texttt{[M]}]\log p_\theta(o_{i,t,p} \mid q)\big],\\
    \pi'_i &= \frac{1}{M}\sum_{o_{i,t}\in M}\log p_\theta(o_{i,t}\mid q).
\end{align}
Following GRPO, UniGRPO loss is an integration of the clipped surrogate rewards $R_{i,t}$ with the KL regularization:
\begin{align}
    &\mathcal{J}_{\text{UniGRPO}}(\theta) =\mathbb{E}\!\left[\frac{1}{G}\sum_{i=1}^G \frac{1}{|o_i|}\sum_{t=1}^{|o_i|} R_{i,t} - \beta D_{\mathrm{KL}}(\pi'_\theta \| \pi'_{\mathrm{ref}})\right].
\end{align}

\subsubsection{Variance-Reduced Preference Optimization}  
LLaDA 1.5~\cite{zhu2025llada} introduces the Variance-Reduced Preference Optimization (VRPO) which replaces the intractable log-likelihoods in DPO with Evidence Lower Bounds (ELBOs)
\begin{align}
\mathcal{B}_{\pi}(y \mid x) &\triangleq \mathbb{E}_{t \sim \mathcal{U}[0,1]} \mathbb{E}_{y_t \sim q(y_t \mid t, y, x)} \ell_{\pi}(y_t, t, y \mid x)\\ &\leq \log \pi(y \mid x),
\end{align}
where $\ell_{\pi}$ is the mask prediction loss.
To reduce estimator variance, VRPO introduces (i) increased sampling budgets, (ii) full sampling budget over timesteps, and (iii) antithetic sampling that shares noise between $\pi_\theta$ and $\pi_{\text{ref}}$. 

\subsubsection{Stepwise Decomposition Preference Optimization}

A central challenge in reinforcement learning alignment for discrete diffusion models is how to propagate reward information through the entire denoising trajectory. Stepwise Decomposition Preference Optimization (SDPO)~\cite{han2025discretediffusiontrajectoryalignment} reformulates trajectory alignment into a collection of tractable per-step alignment objectives.
 
The standard KL-regularized reward optimization objective for diffusion trajectories is:
\begin{equation}
\max_{p_\theta} \; \mathbb{E}_{p_\theta(x_{0:T}\mid c)}\big[r(x_0,c)\big] - \beta D_{\mathrm{KL}}\!\big[p_\theta(x_{0:T}\mid c)\;\|\; p_{\mathrm{ref}}(x_{0:T}\mid c)\big],
\end{equation}
where $c$ is the conditioning context, $r(x_0,c)$ is a reward function on the clean sequence, and $p_{\mathrm{ref}}$ is the reference model. However, this requires sampling and scoring the entire trajectory, which is computationally prohibitive. Instead of optimizing the whole trajectory, SDPO aligns the \emph{per-step posterior}:
\begin{equation}
p_\theta(x_0 \mid x_t, c),
\end{equation}
which admits exact likelihood evaluation under masked diffusion models. The stepwise alignment objective is:
\begin{equation}
\max_{p_\theta}\mathbb{E}_{p_\theta(x_0 \mid x_t, c)}[r(x_0, c)]-\beta_t D_{\mathrm{KL}}\big[p_\theta(x_0 \mid x_t, c) ||p_{\mathrm{ref}}(x_0 \mid x_t, c)\big],
\end{equation}
with step-dependent regularization $\beta_t = \beta / w(t)$. The work shows that this step-wise optimization is equivalent to a distribution matching problem, which can be used to further simplify the loss function.

\subsubsection{Weighted Policy Optimization}
Weighted Policy Optimization (wd1)~\cite{tang2025wd1weightedpolicyoptimization} reformulates the reinforcement learning objective as a \emph{weighted likelihood} objective.

Under the reverse-KL–regularized policy optimization, the target solution $\pi^\ast$ has the closed form. wd1 minimizes the KL divergence $D_{\mathrm{KL}}(\pi^\ast \,\|\, \pi_\theta)$, which reduces to a weighted negative log-likelihood:
\begin{equation}
L_{\mathrm{wd1}}(\theta) = - \mathbb{E}_{q\sim D,\{o_i\}\sim \pi^{\text{ref-old}}} 
\left[ \sum_{i=1}^G w(q,o_i) \cdot \log \pi_\theta(o_i \mid q) \right].
\end{equation}
To address the issue where samples with small advantages receive disproportionately low weights, the weights are defined as:
\begin{align}
w(q, o_i) &= -w^+(q,o_i) + w^-(q,o_i) , \\
w^+(q,o_i) &= \frac{\exp(\psi A_i)}{\sum_{j=1}^G \exp(\psi A_j)}, \\
w^-(q,o_i) &= \frac{\exp(-\psi A_i)}{\sum_{j=1}^G \exp(-\psi A_j)},
\end{align}
where $A_i$ is the centered reward and $\psi$ is a scaling factor.

\subsubsection{Diffusion Chain of Lateral Thought}
Diffusion Chain of Lateral Thought (DCoLT)~\cite{huang2025reinforcingdiffusionchainlateral} introduces a Unmask Policy Module (UPM), which is trained via reinforcement learning to control the decoding order. The UPM learns a ranking-based policy over masked tokens:
\begin{equation}
h^i_{\theta,t} = \text{UPM}(x_{t-1}, t, m_i),
\end{equation}
where $h^i_{\theta,t}$ is the predicted score for token $i$ at step $t$, and $m_i$ is its mask indicator. A Plackett–Luce distribution is then used to sample a top-$K$ unmasking set $U_t$. Once $U_t$ is selected, the model predicts token values for those positions using the standard diffusion langauge model.
The UPM is implemented as a lightweight transformer block attached to diffusion langauge model. 

\section{Decoding Techniques}

\subsection{Unmasking and Remasking for Continuous-Time Models}
\subsubsection{Continuous-Time Unmasking (Flow Matching)}
In continuous-time inference under the discrete flow matching framework (e.g., FUDOKI~\cite{wang2025fudoki}), unmasking is modeled as a stochastic jump process along a probability path.

Let $\mathbf{x}_t$ denote the current sequence state at time $t \in [0,1]$, and let $\mathbf{x}_1$ be the target sequence. For each token position $i$, the update from $\mathbf{x}_t$ to $\mathbf{x}_{t+h}$ is governed by:

\begin{enumerate}
    \item Sample a predicted target $\hat{x}_1^{(i)} \sim p_\theta(x_1^{(i)} \mid \mathbf{x}_t)$;
    \item Compute a total transition rate 
    \[
    \lambda^{(i)} = \sum_{x \ne x_t^{(i)}} u_t^{(i)}(x, x_t^{(i)} \mid \hat{x}_1^{(i)});
    \]
    \item Draw a random variable $Z \sim \mathcal{U}(0, 1)$;
    \item If $Z \leq 1 - e^{-h \lambda^{(i)}}$, update $x_t^{(i)}$ by sampling from:
    \[
    x_{t+h}^{(i)} \sim \frac{u_t^{(i)}(x, x_t^{(i)} \mid \hat{x}_1^{(i)})}{\lambda^{(i)}}.
    \]
\end{enumerate}

This process dynamically determines which tokens to update (i.e., unmask) based on local transition rates. The higher the rate $\lambda^{(i)}$, the more likely token $i$ will jump to a new value, allowing the model to continuously refine its predictions in a semantically meaningful way.

\subsubsection{Remasking under Discrete Flow Matching.}
In the discrete flow matching framework~\cite{dfm}, remasking is incorporated via a velocity field that interpolates between forward and backward update directions:
\begin{equation}
\bar{u}_t^i(x^i, z) = \alpha_t \hat{u}_t^i(x^i, z) - \beta_t \check{u}_t^i(x^i, z),
\end{equation}
where $\hat{u}_t^i$ and $\check{u}_t^i$ denote the forward-time and backward-time velocity fields, respectively. This combined velocity $\bar{u}_t^i$ is valid as long as $\alpha_t, \beta_t > 0$ and satisfies the probability flow condition for $t \in (0, 1)$. When $\alpha_t - \beta_t = 1$, each step progresses forward in time with remasking (corrector sampling), enabling iterative refinement. When $\alpha_t - \beta_t = 0$, the model operates in a stationary regime (corrector iteration), reintroducing noise and adjusting tokens within a fixed diffusion step.

\subsection{Guidance Techniques}
\subsubsection{Classifier-Free Guidance}
\cite{schiff2025simple,nie2025scalingmaskeddiffusionmodels} introduce the unsupervised \emph{classifier-free guidance} strategy for discrete diffusion generation. The method performs two forward predictions at each diffusion timestep $t$:
\begin{itemize}
    \item A \textbf{conditional prediction}, conditioned on both the prompt $p_0$ and the noisy response $r_t$;
    \item An \textbf{unconditional prediction}, conditioned on a sequence of mask token $M$ and the same response $r_t$.
\end{itemize}
The unconditional prediction captures the model’s inherent bias in the absence of instruction signals. The final guided prediction is adjusted as:
\begin{equation}
\tilde{p}_\theta(r_0 \mid p_0, r_t) \propto \frac{p_\theta(r_0 \mid p_0, r_t)^{1 + w}}{p_\theta(r_0 \mid m, r_t)^w},
\end{equation}
where $w$ is a tunable hyperparameter controlling the strength of the guidance. This guidance promotes \emph{text diversity} by reducing the dominance of generic, encouraging prompt-independent responses.

In the continuous-time setting, \cite{nisonoff2025unlocking} provides the formulation of classifier-free guidance applied to the reverse transition matrix
\begin{align}
\hat{R}_t(x,x')= R_t(x', x) \left( \frac{p_t(x)}{p_t(x')} \right)^{w} 
   \left( \frac{q_t(x)}{q_t(x')} \right)^{1-w},
\end{align}
where distribution $q_t$ is the reference distribution.
\cite{rojas2025theoryinformedimprovementsclassifierfreeguidance} further demonstrate that applying strong guidance during early decoding can severely degrade sample quality. This degradation arises because high guidance accelerates the unmasking rate, leading to overly confident and premature token predictions. To mitigate this issue, the authors suggests applying a column-wise normalization of the guided transition matrix. 

\subsubsection{Classifier Guidance}
To improve controllability, for block diffusion model, \cite{schiff2025simple,huang2025ctrldiffboostinglargediffusion} introduce the \emph{classifier guidance} framework that integrates class-conditional preferences into the sampling process. 

At each diffusion step $t$ for block $b$, the guided reverse process modifies the original sampling distribution $p_\theta$ by incorporating the signal from a classifier $p_\xi$, yielding:
\begin{align}
&p_\gamma(x_b^s \mid x_b^t, x_{<b}, y) \nonumber\\ &\qquad\propto p_\theta(x_b^s \mid x_b^t, x_{<b}) \cdot p_\xi(y \mid x_b^s, x_b^t, x_{<b})^\gamma,
\end{align}
where $y$ is the desired class label and $\gamma$ controls the influence of the classifier.
To reduce computational complexity, the method assumes intra-block independence and approximates the classifier as:
\begin{equation}
p_\xi(y \mid x_b^s, x_b^t, x_{<b}) \approx \prod_{\ell=1}^{L} p_\xi(y \mid \hat{x}_{b,t|s}^\ell, x_{<b}),
\end{equation}
where $\hat{x}_{b,t|s}^\ell$ denotes the sequence with the $\ell$-th token in $x_b^t$ replaced by the candidate token $x_{b,\ell}^s$.
This allows the guided probability to be reformulated as:
\begin{align}
&p_\gamma(x_b^s \mid x_b^t, x_{<b}, y) = \nonumber\\&\quad \prod_{\ell=1}^{L} \frac{
p_\theta(x_{b,\ell}^s \mid x_b^t, x_{<b}) \cdot p_\xi(y \mid \hat{x}_{b,t|s}^\ell, x_{<b})^\gamma
}{
\sum_{x'} p_\theta(x' \mid x_b^t, x_{<b}) \cdot p_\xi(y \mid \hat{x}_{b,t|s}'^\ell, x_{<b})^\gamma
}.
\end{align}
By integrating classifier predictions with the model's native probabilities, this approach enables fine-grained, attribute-conditioned generation across blocks while maintaining computational feasibility.

\subsubsection{Reward Guidance}
TESS 2 \cite{tae2025tess} represents a unique approach under the extra-model guidance category by leveraging an external reward model to guide token prediction. The main purpose of this method is to improve the quality of the generated response. Specifically, at each diffusion step, the model output $\hat{S}_\theta$ is first transformed into a token probability distribution:
\begin{align}
    \mathbf{p}_t &= \text{softmax}(\hat{S}_\theta), \\
    \mathbf{c}_w &= \mathbf{E} \mathbf{p}_t,
\end{align}
where $\mathbf{E}$ is the embedding matrix. The resulting continuous representation $\mathbf{c}_w$ is fed into the reward model, which outputs a scalar reward $R \in \mathbb{R}$.

To maximize this reward, TESS 2 performs gradient ascent on the logits by computing $\nabla_\theta R$ and updates the model output:
\begin{equation}
    \hat{S}_\theta := \hat{S}_\theta + \eta \cdot \nabla_\theta R,
\end{equation}
where $\eta$ is a tunable guidance coefficient. This process is performed during inference only, and no guidance is applied during training. By incorporating gradient signals from the reward model, TESS 2 can steer the generation towards more desirable outputs without modifying the base diffusion model.

\subsubsection{Energy-Based Diffusion}
\cite{xu2025energybased} proposes the Energy-based Diffusion Language Model (EDLM), which augments a pretrained diffusion model \( p_{\theta}(x_0 \mid x_t) \) with an unnormalized energy model \( E_{\phi}(x_0, x_t, t) \), yielding a joint denoising distribution:
\begin{equation}
    p_{\theta,\phi}(x_0 \mid x_t) = p_{\theta}(x_0 \mid x_t) \cdot \frac{\exp\left(-E_\phi(x_0, x_t,t)\right)}{Z_{\phi}(x_t)},
\end{equation}
where \( Z_{\phi}(x_t) \) is the intractable partition function required for normalization. This residual formulation corrects the denoising distribution by reweighting samples from the diffusion model using the energy function. The energy function can be derived from either a pretrained autoregressive (AR) model or a finetuned diffusion  model. To implement this energy function-based guidance, \cite{xu2025energybased} adopts an importance sampling strategy. The decoding process at each time step can be summarized as follows.
\begin{enumerate}
    \item Generate \( k \) candidate samples \( \{x_0^{(i)}\}_{i=1}^k \sim p_{\theta}(x_0 \mid x_t) \) using the diffusion model.
    \item Compute the unnormalized energy scores \( e^{(i)} = E_{\phi}(x_0^{(i)}, x_t,t) \) for all samples in $\{x_0^{(i)}\}_{i=1}^k$.
    \item Sample one \( x_0 \) from the candidate pool according to importance weights:
    \begin{equation}
        w^{(i)} = \frac{\exp(-e^{(i)})}{\sum_{j=1}^k \exp(-e^{(j)})}.
    \end{equation}
    \item Use the sampled \( x_0 \) to perform one denoising step via the backward posterior:
    \begin{equation}
        x_{t-1} \sim q(x_{t-1} \mid x_t, x_0).
    \end{equation}
\end{enumerate}

\section{Quantization}
\subsection{DLLMQuant}
\cite{xu2025dllmquant} introduces \textbf{DLLMQuant}, a framework includes Temporal-Mask Adaptive Sampling (TMAS), Interaction-Aware Activation Quantization (IA-AQ), and Certainty-Guided Quantization (CGQ). 

TMAS addresses the calibration challenge in dLLMs by accounting for variations across time steps and masking ratios. Specifically, it divides the iterative generation into blocks and selects calibration inputs at specific time intervals, ensuring coverage of diverse mask ratios across all timesteps. 

Quantization errors \(L(x_t)\) at time step \(t\) in dLLMs accumulate geometrically across denoising steps. Formally, the error propagation can be expressed as:
\begin{equation}
\begin{aligned}
L(x_t) &= x_t - \text{Deq}\!\left(Q\!\left(x_t + L(x_{t+1})\right)\right) \\
       &= Q_{\text{Model}}(x_{t+1}) - Q_{\text{Model}}\!\left(\text{Deq}\!\left(Q(x_{t+1})\right)\right)
\end{aligned}
\end{equation}
where $Q(\cdot)$ is the quantization process, Deq is the dequantization operation, and \(Q_{\text{Model}}\) denotes the quantized model.

A key source of error is the matrix multiplication between softmax outputs and the value matrix in attention. IA-AQ redefines the quantization loss for value matrix $V$ as:
\begin{equation}
L(s) = \left\| \Big( \lfloor \tfrac{V - z}{s} \rfloor - V \Big) \cdot \text{Deq}(O_{\text{softmax}}) \right\|^2_F ,
\end{equation}
where $O_{\text{softmax}}$ is the softmax output, $z$ is zero-point, and $s$ is the scale factor.  
The optimal scaling factor is chosen by:
\begin{equation}
s = \arg \min_{\alpha \in \{1.0, 0.8\}} L(\alpha \odot \hat{s}), \quad \hat{s} = \frac{V_{\max} - V_{\min}}{Q_{\max} - Q_{\min}} .
\end{equation}

In addition, not all tokens equally affect subsequent iterations. Errors on unmasked or low-confidence tokens are small, while masked tokens with high confidence dominate the next step. To account for this, CGQ incorporates token certainty into the Hessian used for weight quantization:
\begin{equation}
H = \big( X \odot (1[X_t = M] + \sqrt{sc_t}) \big) \cdot 
    \big( X \odot (1[X_t = M] + \sqrt{sc_t}) \big)^\top ,
\end{equation}
where $1[X_t = M]$ indicates a custom weighted indicator function, and $sc_t$ is the final confidence score to each token in model output.  
This certainty-weighted Hessian prioritizes minimizing quantization error on critical masked tokens.

\section{Future Directions}
\label{future}
\subsection{Infrastructure}
The infrastructure for dLLMs remains relatively underdeveloped compared to their autoregressive counterparts. In the autoregressive paradigm, the community has benefited from mature open-source models, standardized training frameworks, and reproducible pipelines that facilitate rapid iteration and deployment at scale. Therefore, establishing standardized modular and scalable training frameworks, and open-sourced pretrained models will be critical directions for the community. Building a robust infrastructure will not only promote fair comparisons and accelerate innovation, but also enable practical deployment across a wide range of real-world applications.

\subsection{Inference Efficiency}
Despite their recent successes, dLLMs still face substantial limitations in inference efficiency and system scalability \cite{nie2025large,li2025lavida,ma2025dkv,liu2025dllm}. Future work can explore several key directions to improve their deployability and performance. At the architectural level, incorporating efficient attention mechanisms and multi-scale token representations may help reduce the compute burden during inference. In terms of the denoising process, advancing fast sampling techniques—such as progressive distillation \cite{salimans2022progressive} and adaptive timestep scheduling \cite{watson2021learning,chang2022maskgit}—could accelerate generation without compromising quality. 
In addition, integration with quantized inference (e.g., INT8 or INT4) \cite{dettmers2022gpt3} may yield high-throughput, low-latency generation pipelines. 

\subsection{Security and Privacy}
The security and privacy implications of dLLMs are an emerging concern as these models become more widely used. 
Diffusion models share similar risks with other large generative models \cite{zhang2024extracting,chen2024text,morris2023language,li2025vid,li2024membership,he2025towards,zhang2024min,wang2025towards}: they can inadvertently memorize and regurgitate sensitive training data, raising the possibility of privacy leakage or copyright violations. Recent studies have demonstrated that diffusion models trained on vast internet data can reproduce portions of their training examples~\cite{carlini2023extracting}, much like LLMs. 
In addition, security in terms of model misuse and alignment is another crucial aspect. Like any powerful language model, dLLMs could be misused to generate harmful, false, or biased content \cite{qu2023unsafe,zhang2024generate}. One challenge is that controlling a diffusion model’s output may require new methods: unlike AR models that can be guided token-by-token or halted upon generating disallowed tokens \cite{shen2024anything}, diffusion models generate content in a more holistic way. This makes real-time content moderation non-trivial---dLLMs might only reveal problematic content once the final denoised text is produced. These areas remain critical future directions to address before dLLMs can be responsibly deployed at scale.

\nocite{zheng2023reparameterized,ou2025absorbingdiscretediffusionsecretly,ye2023diffusion,zheng2023reparameterized,ou2025absorbingdiscretediffusionsecretly,ye2023diffusion,devlin2019bert,kaplan2020scaling,bahri2024explaining,floridi2020gpt,ye2023diffusion,ye2023diffusion,gong2024scaling,radford2019language,touvron2023llama,gong2024scaling,touvron2023llama,tae2025tess,gong2024scaling,ouyang2022training,gong2025diffucoder,hayakawa2024distillation,wang2025fudoki,dfm,schiff2025simple,nie2025scalingmaskeddiffusionmodels,nisonoff2025unlocking,rojas2025theoryinformedimprovementsclassifierfreeguidance,schiff2025simple,huang2025ctrldiffboostinglargediffusion,tae2025tess,xu2025energybased,xu2025energybased,liu2025longlladaunlockinglongcontext,LocalLLaMA_dynamically_scaled_RoPE_2023,lyu2023fine,zhang2025diffusion,zhu2024segment,cheng2024diffuspoll,hu2024poetrydiffusion,padole2025improving,lee2025editext,do2025discrete,shao2025diffetm,dong2025termdiffusum,xin2025cdaˆ2,chen2024effective,zhu2024pinpointing,iwai2024layout,shaahid2025underwater,chi2024m2d2m,sun2025ar,chang2022maskgit,bai2024meissonic,bai2023qwen,qwen2025qwen25technicalreport,yang2025qwen3,liu2023llava,Qwen-VL,liu2024llavanext,liu2023improvedllava,liu2024llavanext,nie2025large,li2025lavida,ma2025dkv,liu2025dllm,dao2022flashattention,salimans2022progressive,watson2021learning,chang2022maskgit,dettmers2022gpt3,zhang2024extracting,chen2024text,morris2023language,li2025vid,li2024membership,he2025towards,zhang2024min,wang2025towards,carlini2023extracting,abadi2016deep,santos2022avoiding,ying2019overview,chen2024data,li2024superfiltering,qu2023unsafe,zhang2024generate,shen2024anything}

\end{document}